\documentclass[11pt]{article}
\usepackage{gbm}
\newcommand{\geo}[4]{\bG_{#1,#2}(#3,#4)}
\newcommand{\num}{\mathrm{num}}
\newcommand{\Unif}{\mathrm{Unif}}
\newcommand{\op}{\mathrm{op}}

\usepackage{setspace}
\usepackage{cprotect}
\usepackage{makeidx}
\usepackage[mathscr]{euscript}
\usepackage{accents}

\title{Spectral clustering in the Gaussian mixture block model}

\author{Shuangping Li\thanks{Stanford University. \texttt{fifalsp@stanford.edu}.} \and Tselil Schramm\thanks{Stanford University. \texttt{tselil@stanford.edu}.}}

\date{}

\begin{document}
\maketitle
\begin{abstract}
\emph{Gaussian mixture block models} are distributions over graphs that strive to model modern networks: to generate a graph from such a model, we associate each vertex $i$ with a latent ``feature'' vector $u_i \in \R^d$ sampled from a mixture of Gaussians, and we add edge $(i,j)$ if and only if the feature vectors are sufficiently similar, in that $\iprod{u_i,u_j} \ge \tau$ for a pre-specified threshold $\tau$.
The different components of the Gaussian mixture represent the fact that there may be different types of nodes with different distributions over features---for example, in a social network each component represents the different attributes of a distinct community.
Natural algorithmic tasks associated with these networks are embedding (recovering the latent feature vectors) and clustering (grouping nodes by their mixture component).

In this paper we initiate the study of clustering and embedding graphs sampled from high-dimensional Gaussian mixture block models, where the dimension of the latent feature vectors $d\to \infty$ as the size of the network $n \to \infty$.
This high-dimensional setting is most appropriate in the context of modern networks, in which we think of the latent feature space as being high-dimensional.
We analyze the performance of canonical spectral clustering and embedding algorithms for such graphs in the case of 2-component spherical Gaussian mixtures, and begin to sketch out the information-computation landscape for clustering and embedding in these models.

\end{abstract}

\thispagestyle{empty}
\clearpage

{\hypersetup{hidelinks}\tableofcontents}
\thispagestyle{empty}
\clearpage
\setcounter{page}{1}
\section{Introduction}

For algorithmic problems arising in data science, it is useful to study ``model organisms:'' families of synthetic datasets which aspire to faithfully represent the input data, while being simple enough to admit provable guarantees for the algorithms in question.

Consider, for example, the task of clustering in social networks: we observe an unlabeled graph in which each node belongs to one of several communities, and we would like to partition the graph into communities.
This problem is hard to study in the traditional worst-case analysis framework for several reasons.
Firstly, many of the objectives associated with finding community structure (e.g. sparsest cut, correlation clustering) are computationally intractable \cite{MS90,BBC04} (even to approximate \cite{CKKRS06}) in the worst case.
Secondly, it is not even clear that sparsest cut and other proxy clustering objectives faithfully capture community structure in real-life networks; in fact, they are sometimes poorly correlated with network structure even in non-adversarial synthetic models (see Section 3.1 of \cite{Abbe17}).
Model organisms (such as the stochastic block model) allow researchers to ``theoretically benchmark'' the performance of their clustering algorithms.
Even if the algorithm has no provable guarantees for worst-case inputs, at least we can rest assured that it works well in a simple generative data model.

Perhaps the most popular model for this task is the \textit{Stochastic Block Model} (SBM) \cite{HLL83}, a generalization of an \erdos-\renyi graph in which each node belongs to an unknown community, and each edge is present independently with probability $p$ between nodes in the same community and probability $q$ between nodes in different communities (see \cite{Abbe17} for a definition of the model in its full generality).
When $p > q$, this model reflects the empirical observation that social networks tend to have denser connections within communities.
The SBM has been a successful model organism in that it has allowed for a nuanced study of spectral methods, motif (small subgraph) counting, and other algorithms for community detection (e.g. \cite{MNS15,Mas14,BMR21,DONS21}), and a rich mathematical theory has developed to describe its behavior (e.g. \cite{MPW16,MNS16,AS15}, see also the survey \cite{Abbe17}).
However, this simple model leaves much to be desired because it fails to capture much of the structure of real-life social networks (see e.g. the discussion in \cite{GMSS18}).

We study the following \textit{Gaussian mixture block model} (GMBM) as a model organism for community recovery.
The model is meant to reflect the conception of networks wherein each node is associated with a latent \textit{feature vector} which describes its characteristics, and pairs of nodes with similar feature vectors are more likely to be connected.
To generate a sample from the spherical 2-community Gaussian mixture block model, $G \sim \geo{n}{d}{p}{\mu}$ with $n,d \in \Z_+, \mu \in \R_+, p \in [0,1]$, (1) independently sample $n$ latent vectors from a mixture of Gaussians in $\R^d$,
\begin{align*}
u_1,\ldots,u_n \sim \frac{1}{2} \calN(-\mu \cdot e_1, \tfrac{1}{d}I_d) + \frac{1}{2} \calN(\mu \cdot e_1, \tfrac{1}{d}I_d),
\end{align*}
then (2) for each $i\neq j \in [n]$, add edge $(i,j)$ to $G$ if and only if the corresponding vectors are $\tau$-correlated, $\iprod{u_i,u_j} \ge \tau$, where $\tau$ is chosen in advance as a function of $n,d,\mu$, and $p$ so that the edge probability $\Pr[(i,j) \in E(G)]$ is $p$.
Note that we ultimately observe only $G$ and not the latent embedding $u_1,\ldots,u_n$.

Each Gaussian component of the mixture represents the characteristics of a community as a distribution over feature space, and the distance between the means, $2\mu$, is a measure of the communities' separation.
The edge criterion $\iprod{u_i,u_j} \ge \tau$ reflects the intuition that nodes with similar feature vectors are more likely to be connected; the larger we set $\tau$, the more stringent the connection criterion is, and therefore the sparser the resulting network becomes.

Variants of this model have been studied in the past, albeit with minor variations (for example, $u_i$ sampled from the uniform distribution over the sphere rather than a Gaussian mixture) \cite{GMSS18, EMP22}.
But to date, the focus has been on the (more mathematically tractable) low-dimensional regime, where $d$ remains fixed as $n \to \infty$.

In this work, we will study the performance of spectral algorithms in the GMBM in the \textit{high-dimensional regime}, where $d \to \infty$ as $n \to \infty$.
This high-dimensional setting is more compatible with our conception of modern networks, in which we think of the feature space as being large, on a scale comparable to the networks' size.
We will show that so long as $p$ is chosen to ensure that the network is not too sparse and so long as the dimension of the feature space $d$ is not too large relative to the number of nodes $n$, the canonical spectral embedding algorithm provides a good estimate of the latent embedding $u_1,\ldots,u_n$ (up to rotation).
Further, if the separation between the communities $\mu$ is large enough, the spectral embedding allows us to test for the presence of and/or recover the communities.

Spectral embedding methods are used widely throughout network science, and our analysis is the first that provides provable guarantees for their performance in the relatively realistic high-dimensional Gaussian mixture block model.
However, our work is merely an initial step; we will formulate several open questions for future research, both towards the goal of increasing the realism of the model, and better understanding the information-computation landscape of this simplest model.

\subsection{Our results}

We begin by defining our model and formulating our algorithmic objectives.

\begin{definition}[Gaussian mixture block model]
For $n,d \in \Z_+$ and $\mu \ge 0, p \in[0,1]$, the \textit{$n$-vertex, $d$-dimensional, $\mu$-separated, 2-community Gaussian mixture block model with edge probability $p$} is the distribution over $n$-vertex graphs $G \sim \geo{n}{d}{p}{\mu}$ defined by the following sampling procedure:
\begin{enumerate}[itemsep=-2pt, topsep=2pt]
\item Independently sample $n$ $d$-dimensional vectors $u_1,\ldots,u_n\sim \frac{1}{2}\calN(-\mu \cdot e_1, \frac{1}{d}I_d) + \frac{1}{2}\calN(\mu \cdot e_1, \frac{1}{d}I_d)$.
\item Let $V(G) = [n]$ and add $(i,j)$ to $E(G)$ if and only if $\iprod{u_i,u_j} \ge\tau$,
\end{enumerate}
where $\tau = \tau(n,d,\mu,p)$ is a threshold chosen in advance so that $\Pr[\iprod{u_i,u_j} \ge \tau] = p$.

We say that vertex $i \in [n]$ comes from community $+1$ if $u_i$ comes from the component in the mixture with mean $\mu \cdot e_1$; otherwise we say that vertex $i$ comes from community $-1$.
\end{definition}

\begin{remark}
Because of the rotational invariance of $\calN(0,\frac{1}{d}I_d)$, the distribution over graphs produced by the mixture with means $\pm \mu \cdot e_1$ is equivalent to that produced by the means $\pm \theta$ for any $\theta \in \R^d$ with $\|\theta\| = \mu$.
We suspect that our techniques will generalize in a straightforward manner to the case in which the points are sampled from a mixture with any pair of means $\frac{1}{2}\cN(\theta_1,\frac{1}{d}I_d) + \frac{1}{2}\cN(\theta_2,\frac{1}{d}I_d)$, or to the case when edges are based on the distance criterion $\|u_i-u_j\| < \sigma$ (so that $\theta_1 = -\theta_2$ is equivalent to the general case) rather than correlation criterion.
We opted to analyze only this special case in order to keep the proofs simpler.
\end{remark}

\begin{problem}
There are three algorithmic problems we associate with the GMBM:
\begin{enumerate}[itemsep=-2pt, topsep=2pt]
\item \textbf{Latent vector recovery:} Given $G \sim \geo{n}{d}{p}{\mu}$, can we estimate the latent vectors $u_1,\ldots,u_n$ up to rotation?
\item \textbf{Hypothesis testing:} Can we tell if the data that generated $G$ comes from two distinct clusters?
Formally, we would like to hypothesis test between the ``one community'' null hypothesis that $G \sim \geo{n}{d}{p}{0}$ (so that $\mu = 0$ and the two components of the mixture are merged) and the ``two community'' alternative hypothesis $G \sim \geo{n}{d}{p}{\mu}$ with $\mu > 0$.
\item \textbf{Clustering:} When $\mu > 0$, can we partition the vertices $V(G)$ into sets $S,V\setminus S$ so that for all (or most) $i \in S$, $i$ comes from community $+1$ and for all (or most) $j \in V \setminus S$, $j$ is in community $-1$?
\end{enumerate}
\end{problem}

\paragraph{Spectral algorithm.}
Spectral methods are broadly employed for clustering and embedding tasks in network science.
Though many variations on the basic concept exist, in this work we are mostly concerned with providing rigorous guarantees for this model organism, and we analyze a very canonical and ``vanilla'' spectral algorithm.
Roughly, the algorithm is as follows: given input graph $G$ on $n$ vertices,
\begin{enumerate}[itemsep=-2pt, topsep=4pt]
\item[(a)] Assemble the $n \times n$ adjacency matrix $A$ with $A_{i,j} = \Ind[(i,j) \in E(G)]$, then
\item[(b)] Compute the top $d+1$ eigenvalues and unit eigenvectors $\{(\eta_i,w_i)\}_{i=0}^{d}$ of $A$, where $\eta_0 \ge \cdots \ge \eta_{n-1}$.
\end{enumerate}
\noindent
If $d$ is unknown, in step (b) we instead find the largest gap $\eta_i-\eta_{i+1}$ over $2\leq i\leq n-2$ and use the maximizing index as our estimate of $d$. (See Proposition~\ref{p:unknownd} in Section~\ref{sec:unknownd} for a formal statement.)
We then use this spectral information to solve vector recovery, hypothesis testing, or clustering, employing a slightly different final step in each case:
\begin{itemize}[itemsep=-2pt,topsep=4pt]
\item For the task of recovery, we define the vectors $\hat u_1,\ldots,\hat u_n$ with $\hat u_j(i) \propto w_{i}(j)$, and let $\hat u_j$ be our estimate for $u_j$ (up to rotation).\footnote{Note vector entries $u_j(\cdot)$ and $w_i(\cdot)$ are indexed starting at $1$, whereas the eigenvectors $w_{(\cdot)}$ are indexed starting at $0$. The precise normalization constant depends on the Gegenbauer polynomial expansion and is defined in Sections~2 and~4.}
\item For the task of hypothesis testing, we check if $\eta_1 > \theta$ for a threshold $\theta \in \R$ chosen as a function of $n,d,\mu,\tau$ (larger $\eta_1$ corresponds to the case when $\mu > 0$).\footnote{The precise threshold is defined in Section~4.}
\item Finally, for the task of clustering, we embed the points on the line according to the vector $w_1$ and we let community membership be defined by a threshold cut along the line.
\end{itemize}
\noindent For each task, we are able to prove that the above algorithm works, provided the number of vertices $n$ is large enough as a function of the dimension $d$, $p$ is large enough so that the graph is not too sparse, and the separation $\mu$ between the cluster centers is not too small (except for some gaps of logarithmic width).
In what follows, we use $\lllog$ to denote $\ll$ with a polylog factor.
The following diagram summarizes our results; we give the theorem statements below.

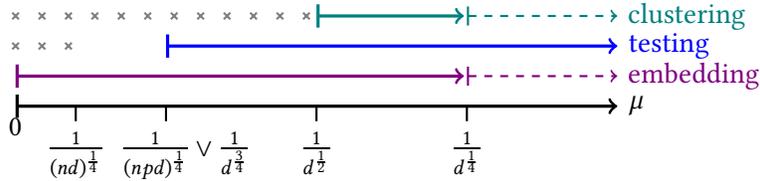
\begin{figure}[h!]
    \centering
\usetikzlibrary {decorations.shapes}
\begin{tikzpicture}[scale=0.8]
\draw[|->, very thick] node[below]{$0$} (0,0) -- (10,0) node[right]{$\mu$};
\draw[thick]  (7.5,-.25) node[below]{$\frac{1}{d^{\frac{1}{4}}}$} -- (7.5,0);
\draw[thick]  (5,-.25) node[below] {$\frac{1}{d^{\frac{1}{2}}}$} -- (5,0);
\draw[thick]  (2.5,-.25) node[below]{$\quad\,\,\frac{1}{(npd)^{\frac{1}{4}}} \vee \frac{1}{d^{\frac{3}{4}}}$} -- (2.5,0);
\draw[thick]  (1,-.25) node[below]{$\frac{1}{(nd)^{\frac{1}{4}}}$} -- (1,0);
\draw[violet, very thick, |->] (0,.5) -- (7.45,.5);
\draw[violet, thick, dashed, |->] (7.5,.5) -- (10,.5) node[right,violet]{embedding};
\draw[blue, very thick, |->] (2.5,1) -- (10,1) node[right,blue]{testing};
\draw[teal, very thick, |->] (5,1.5) -- (7.45,1.5);
\draw[teal, thick, dashed, |->] (7.5,1.5) -- (10,1.5) node[right,teal]{clustering};
\draw[gray, thick, -] decorate [decoration={name=crosses}] {(0,1.5) -- (5,1.5)};
\draw[gray, thick, -] decorate [decoration={name=crosses}] {(0,1) -- (1,1)};
\end{tikzpicture}
    \caption{Diagram illustrating the range of $\mu$ for which we show that the spectral algorithm completes each task successfully (up to logarithmic factors), all under the condition that $1\llog d\lllog pn$.
    The solid lines correspond to our theorems.
    The dashed teal line indicates that beyond $d^{-1/4}$, each community corresponds to a distinct connected component in the graph and thus spectral clustering trivially succeeds.
    Similarly, the dashed violet line indicates that beyond $d^{-1/4}$, the community labels suffice to recover an approximate embedding.
The gray $x$'s mark a range in which clustering/testing is impossible even when the latent embedding is known (lower bounds for clustering in \cite{Ndaoud22}, for testing in \pref{app:kl}).}
    \label{f:diagram}
\end{figure}
\begin{theorem}[Latent vector recovery/embedding]\torestate{
\label{thm:latent1}
Suppose that $n,d\in \Z_+$ and $\mu \in \R_+$, and  $p \in [0,1/2-\varepsilon]$ for any constant $\eps>0$, satisfy the conditions $\log^{16} n\ll d < n$, $\mu^2 \leq 1/(\sqrt{d}\log n)$, and $pn\gg 1$.
Then given $G \sim \geo{n}{d}{p}{\mu}$ generated by latent vectors $u_1,\ldots,u_n \in \R^d$, the spectral algorithm described above produces vectors $\hat{u}_1,\ldots,\hat{u}_n$ which satisfy
\begin{align*}
    \E_{i,j \sim [n]} |\inp{\hat u_i,\hat u_j}- \inp{u_i,u_j}| \ll \max \left\{ \sqrt{\tfrac{\log 1/p}{d}},\, \mu^2\cdot \sqrt{\tfrac{d}{\log \frac{1}{p}}}, \sqrt{\tfrac{d}{np\log \frac{1}{p}}} \right\} \log^9 n
\E_{i,j \sim [n]} |\inp{u_i,u_j}|,
\end{align*}
with high probability as $n$ goes to infinity.
}
\end{theorem}

    As long as $1\lllog d \lllog pn$ and $\mu \lllog d^{-1/4}$, the relative error in \pref{thm:latent1} is $o(1)$ and the $\hat u_i$ recover the $u_i$ approximately, up to rotation.
See also \pref{thm:latent2} which states an approximation result in terms of the spectral distance between the matrices whose columns are given by $\hat{u}_i$ and $u_i$ respectively.

The condition $d \ll pn$ asks that the dimension not exceed the average vertex degree.
It is not difficult to see that if $d \to \infty$ too fast relative to $n$, the geometry disappears (because the quantities $\{\iprod{u_i,u_j}\}_{i,j \in [n]}$ become increasingly independent).
Similarly, when $p$ is small more information is ``lost.''
So it is unsurprising that we see an upper bound on $d$ as a function of $n$ and $p$.
However, it is not clear to us whether $d \ll np$ is sharp (even up to logarithmic factors); we discuss more below in the ``lower bounds'' paragraph.

The condition $\mu \lllog d^{-1/4}$ ensures that the separation between the communities does not ``drown out'' other geometric information present in the graph.
Indeed, whenever $\mu\gg (1/\sqrt{d\tau})=\Theta((d\log\frac{1}{p})^{-1/4})$, the number of edges between communities is of a smaller order than the number of edges within each community, and whenever $\mu \gg d^{-1/4}\log^{1/4}n$, the components of the graph corresponding to the communities become disconnected with high probability.\footnote{
For reference, so long as $d = \omega(1)$ and $\mu$ is small enough that it does not dominate the edge probability, $\tau = \Theta(\ssqrt{\frac{1}{d}\log 1/p})$.
    }
In this $\mu \gglog d^{-1/4}$ range, embedding reduces to clustering, as $\hat{u}_i = x_i \mu e_1$ for $x_i$ the $\pm 1$ label of node $i$ is a good approximate embedding for $u_i$.

\begin{theorem}[Hypothesis Testing]\label{thm:testingmain}
 Define the one-community model to be the null hypothesis $H_0 = \geo{n}{d}{p}{0}$ and the two-community mixture model to be the alternative hypothesis $H_1 = \geo{n}{d}{p}{\mu}$.
If $d,n,\mu,p$ satisfy
\begin{align*}
    \mu^2 \geq \max \left\{ \sqrt{\tfrac{\log 1/p}{d^{3}}}, \sqrt{\tfrac{1}{np d \log\frac{1}{p}}} \right\} \log^9 n, \quad \log^{16} n\ll d < n, \quad pn \gg 1, \quad p \in [0,1/2-\varepsilon],
\end{align*}
for any constant $\eps > 0$, then if we run the spectral algorithm described above on input graph $G$ we have that
\begin{align*}
    &\min\left\{\Pr(\text{accept } H_0 \mid G \sim H_0),\, \Pr(\text{reject } H_0 \mid G \sim H_1)\right\}\geq 1-o_n(1),
\end{align*}
with high probability as $n$ goes to infinity. In other words, both type 1 and type 2 error go to zero as $n\to\infty$.
\end{theorem}

   \pref{thm:testingmain} requires that $\mu \gglog \max( d^{-3/4}, (npd)^{-1/4} )$ for the errors to vanish.
This is much smaller than $1/\sqrt{d}$ whenever $d \lllog pn$. In other words, whenever $d \lllog pn$, we can tell apart the two models even if the mean separation is $o(1/\sqrt{d})$ and almost-exact clustering is impossible.
A second moment computation shows that $\mu = \Omega((nd)^{-1/4})$ is necessary for testing (\pref{app:kl}).

\begin{theorem}[Spectral clustering]\torestate{
\label{thm:spectralc}
Suppose $d,n \in \Z_+$, $p \in [0,1/2-\varepsilon]$ for any constant $\eps > 0$, and $\mu > 0$ satisfy the conditions $d^{-1/2} \ll \mu \leq d^{-1/4}\log^{-1/2}n$, $\log^{16} n\ll d < n$ and $pn\gg 1$.
If $G \sim \geo{n}{d}{p}{\mu}$, then with high probability the spectral algorithm described above correctly labels (up to a global sign flip) a
\begin{align*}
1- O\left(\frac{1}{\mu \sqrt{d}}+ \sqrt{\max\left\{ \mu^2 \cdot \sqrt{\tfrac{\log 1/p}{d}}, \tfrac{1}{\mu^2 \sqrt{npd \log \frac{1}{p}}}\right\} \log^9 n }\right)\text{-fraction of the vertices.}
\end{align*}
}
\end{theorem}
    \pref{thm:spectralc} implies that when $1\lllog d\lllog pn$ and $d^{-1/2}\lllog \mu \lllog d^{-1/4}$, the spectral clustering algorithm clusters a $(1-o(1))$-fraction of vertices correctly.
As discussed briefly above, we expect that when $\mu$ exceeds the range covered by \pref{thm:spectralc}, the spectral algorithm also works because the cut between the $+1$ and $-1$ labeled vertices becomes sparse (and eventually has no crossing edges with high probability when $\mu \gg (\frac{1}{d}\log n)^{1/4}$); however this regime requires a different analysis than our theorem (and can probably avoid the trace method), and we defer it to future work.
The bound $d^{-1/2} \lllog \mu$ matches the lower bound in the Gaussian mixture model (when the $u_i$ are not latent but observed) up to logarithmic factors \cite{Ndaoud22}.

\paragraph{Lower bounds.}
It is interesting to know whether our results are tight. A priori, in some regimes we can deduce information-theoretic barriers to solving the embedding and/or hypothesis testing problem by appealing to a known barrier for another model:
\begin{itemize}[itemsep=0pt]
\item When $\mu$ is too small (precisely, when $\mu \lesssim \ssqrt{\frac{1}{d} + \frac{1}{\sqrt{nd}}}$), clustering in the underlying Gaussian mixture model is information-theoretically impossible \cite{Ndaoud22}.
Given only $G \sim \geo{n}{d}{p}{\mu}$, we have access to strictly less information than if we were handed the latent embedding, so clustering in $G$ in this regime is impossible.
\item Similarly, we show in \pref{app:kl} that when $\mu \lesssim \left(\frac{1}{nd}\right)^{1/4}$, the underlying Gaussian mixture model cannot be hypothesis tested against $\calN(0,\frac{1}{d}\Id)$ from $n$ samples.
This implies that testing is impossible in this regime.
\item When $d = \Omega(\log n)$ and $\mu = 0$, the graph we observe is a random geometric graph with points sampled from $\calN(0,\frac{1}{d}I_d)$, which is not too different from a random geometric graph over $\calS^{d-1}$.
A random geometric graph over $\calS^{d-1}$ with average degree $np$ is known to be indistinguishable from \erdos-\renyi $\bG(n,p)$ when $d \gtrsim n^3 p^2$, and is conjectured indistinguishable when $d \gtrsim (n H(p))^3$ for $H(\cdot)$ the binary entropy function \cite{LMSY22test}.
In this regime, the embedding that generated the GMBM graph $G$ is likely not identifiable, and at the very least it is not meaningful because the geometry has effectively ``disappeared'' in the observed graph.
\end{itemize}
These lower bounds are inherited by $\geo{n}{d}{p}{\mu}$ from these simpler models, and they are consistent with our results which do need $\mu$ not too small to cluster and $d$ not too large relative to $np$.
However, these lower bounds do not give the full story---the first two bounds do not account for the information lost when the embedding is thrown out, and the latter lower bound is very far from our result,\footnote{We require $d \lesssim np$, rather than $d \lesssim (n H(p))^3$, though we are not sure if either of these bounds is sharp.} and does not explain what happens when $\mu$ is large.

The condition $d = O(n H(p))$ for $H(\cdot)$ the binary entropy function seems plausibly tight for spectral algorithms; see \pref{sec:future} for more discussion.
We also mention that since the span of the embedding vectors $u_1,\ldots,u_n$ has rank at most $n$, when the embedding dimension $d > n$, there is always some set of vectors in $n$ dimensions, $v_1,\ldots,v_n \in \R^n$, so that $\iprod{u_i,u_j} = \iprod{v_i,v_j}$ for all $i,j$; hence the embedding which produced the graph is certainly no longer unique up to rotation.
However, it is possible that the embedding may still be identifiable (up to rotation) in the $d > n$ regime, because the $u_i$ also behave like Gaussian vectors; it is not clear whether the same is true for alternate embeddings such as the one given by $\{v_i\}_{i \in [n]}$.

We leave proving lower bounds as an interesting open question.
In the case of embedding, one might imagine that in the large-$d$ regime, it might be possible to demonstrate (via direct calculation or otherwise) that the posterior distribution over embeddings has high entropy.
An alternate approach in the sparse regime might be to try to adapt the arguments for lower bounds in the stochastic block model (using information flow on trees); see \cite{EMP22} for a discussion of this possibility.

\subsection{Related work}

\paragraph{Gaussian mixture block model and variations.}
Most of the prior work on geometric block models in the literature focuses on the low-dimensional regime, where the dimension $d$ is held fixed as $n \to \infty$.
For example, in \cite{ABRS20}, the authors study the performance of a (somewhat different) spectral algorithm for the approximate clustering problem in the special case of $d = 2$.
Another previously studied slight variation on our model was studied in \cite{GMSS18,galhotra2019connectivity}.
There, the authors introduce a variant of the model in which the $i$th node's latent vector $z_i$ is sampled uniformly at random from the unit sphere $\calS^{d-1}$, and each node also has a latent community label $x_i \in [k]$.
Edge $(i,j)$ is present if and only if $\iprod{z_i,z_j} \geq \tau_{x_i,x_j}$, so the connectivity threshold depends on the communities that the nodes belong to.
Though not exactly the same, this is not too dissimilar from our model.\footnote{In our case $u_i = Z_i + x_i \mu e_1$ for $Z_i \sim \calN(0,\frac{1}{d}I_d)$ (almost like vectors on the sphere when $d$ is large) and $x_i \in \{\pm 1\}$ the community label of node $i$, so that $\iprod{u_i,u_j} = \iprod{Z_i,Z_j} + \mu \iprod{x_j Z_i  + x_i Z_j, e_1} + x_i x_j \mu^2$.
Now the condition $\iprod{u_i,u_j} \ge \tau$ is equivalent to the condition $\iprod{Z_i,Z_j} \ge \tau - x_i x_j \mu^2$ up to the random fluctuation $\mu \iprod{x_j Z_i  + x_i Z_j, e_1}$.}
The authors study the relatively low-dimensional case $d=O(\log n)$ for this model, and give a clustering algorithm based on counting motifs (small subgraphs) that works in some parameter regimes.
See also \cite{ABD21} for a result on the success of spectral clustering when the feature vectors are drawn from the uniform measure over the torus $\mathbb{T}^d$ with $d = O(1)$.

Other works have considered variants of the classical stochastic block model which incorporate higher-dimensional geometry.
In \cite{SB18,ABS21} the authors consider a version of the SBM in which one observes a {known} embedding of the nodes in $\R^d$, but the community labels of the nodes are latent and the edges are a function of both the embedding and the labels.
Their setting differs significantly from ours because the node embedding is not latent.
Another instance of a geometric variant of the block model is the mixed-membership $k$-community stochastic block model \cite{ABFX08}.
There each node $i$ has a latent $k$-dimensional ``community membership'' distribution $u_i \in \Delta_k$ where $\Delta_k$ denotes the $k$-dimensional simplex; $u_i$ is supposed to represent node $i$'s fractional belonging to each of the $k$ different communities.
The $u_i$ are sampled independently from a Dirichlet distribution, and then the presence of the edge $(i,j)$ is a randomized function of $u_i$ and $u_j$.

In this context (more sophisticated) spectral algorithms are also known to recover the latent embedding, even in the sparse and high-dimensional regime where $k$ grows with $n$ \cite{HS17}.
The underlying geometry in the mixed-membership block model is quite different from the Gaussian mixture block model (as the embedding vectors $u_i$ are supposed to represent something else), so the result of \cite{HS17} does not imply anything for our setting.
Still, the fact that the techniques in these cases are similar further points to a potential universality of methods for recovering embeddings in random geometric graphs.
We also think it likely that the spectral algorithms from \cite{HS17} could give sharper (up to logarithmic factors) algorithmic thresholds, for instance removing polylogs from the inequality $d \lesssim np$ and potentially allowing us to handle sparse graphs with $np = \Theta(1)$; however in this work our goal was to analyze the more ``canonical'' and efficient basic spectral method.

Another closely related class of latent position models is the \emph{Random Dot Product Graph} (RDPG), which has received significant attention in the statistics literature (see e.g.~\cite{young2007random,athreya2018statistical}). In the RDPG model, each vertex $i$ is associated with a latent vector $x_i \in \mathbb{R}^d$, and edges are generated independently with probability $x_i^\top x_j$. Our model is closely related in that both frameworks posit latent vectors whose inner products determine edge formation. However, there are several important differences. First, most work on RDPGs focuses on the regime where the latent dimension $d$ is fixed as $n \to \infty$, whereas our focus is on the high-dimensional setting where $d$ may grow with $n$. Second, edges in the Gaussian mixture block model are generated via a threshold rule $\Ind[\langle u_i, u_j \rangle \ge \tau]$ rather than through a probabilistic link function. These differences lead to distinct technical challenges and require a different analysis of the spectral embedding in the high-dimensional regime.

\paragraph{Recovering embeddings of random geometric graphs.}
A \textit{random geometric graph} is any graph which is generated by sampling points according to a measure on a metric space, associating each point to a vertex, and connecting vertices according to a probability which depends on their distance in the metric.
The GMBM is a geometric random graph where the metric space is Euclidean space and the measure is a mixture of two spherical Gaussians.
Recovering the embedding is a natural algorithmic task on random geometric graphs; we mention a couple of relevant works here.

Motivated by social networks, the work \cite{EMP22} considers the task of recovering the embedding of a random geometric graph over a single unit sphere $\calS^{d-1}$, where the edge indicator for $(i,j)$ is distributed as a $\mathrm{Bernoulli}(\frac{1}{n}\phi(\iprod{u_i,u_j}))$ for $u_i,u_j$ the latent embedding vectors (the fact that the edge probabilities are a function of distance is supposed to mimic community structure).
They prove that so long as $\phi$'s spectrum satisfies certain conditions (which restrict their result to the low-dimensional setting), the spectral embedding approximates the latent embedding well.

The work of \cite{EMP22} makes use of techniques used in the study of {kernel random matrices}, which is a class of random matrices that includes the adjacency matrices of random geometric graphs.
These are matrices which are sampled by first sampling $n$ latent vectors $u_1,\ldots,u_n$ uniformly from $\calS^{d-1}$, and then taking the $(i,j)$-th entry to be a (deterministic) function $\phi$ of $\iprod{u_i,u_j}$.
The spectrum (eigenvalues and eigenvectors) of a kernel random matrix can be thought of as a random approximation to the spectrum of the associated integral operator.
A sufficiently strong quantitative bound on the strength of this approximation in the context of the random geometric graph kernel function, $\phi(x) = \Ind[x \ge \tau]$, would imply that spectral algorithms recover the latent embedding.
For example, in the low-dimensional setting where $d = O(1)$, the classical results \cite{KG00} imply that a spectral algorithm recovers the embedding of a kernel random matrix up to error which goes to zero as $n \to \infty$.
See also \cite{AY19}.

In the high dimensional setting less is known.
A series of works \cite{Karoui10,CS13,DV13,Bordenave13,FM19,LY22} characterizes the empirical spectral densities of well-behaved kernel functions, under restrictions on the relationship between $d$ and $n$ (state of the art requires $d = n^{1/k}$ for integer $k$).
Recently, in a push to more finely characterize the non-asymptotic performance of Kernel methods in machine learning, a series of papers has made progress on understanding the spectral edge and eigenvectors of certain kernel random matrices as well.
See \cite{FM19, ghorbani2021linearized,mei2021learning,mei2022generalization}, which study a fairly flexible class of kernel functions (though as far as we understand, their results do not accommodate random geometric graphs).
Our result shows that part of the spectrum of the random geometric graph approximates the associated operator's spectrum well.
We mention as well the work \cite{LMSY22}, in which the authors obtain sharp bounds on the spectral gap of random geometric graphs on the high-dimensional sphere.

\paragraph{Clustering Gaussian mixtures.}
The well-studied problem of clustering mixture distributions, and specifically Gaussian mixtures, is the easier version of our problem in which we are already given access to the latent embedding of our vertices in $\R^d$.
The performance of spectral clustering algorithms in this context has been well studied, with a focus on the many-cluster case (see e.g. \cite{VW04}, and \cite{Ndaoud22} for the case of 2 Gaussians).

Our problem is harder because we do not observe this latent embedding.
Our clustering algorithms work by trying to find an approximate embedding, and then applying spectral clustering.
It is interesting to ask whether we could use a more sophisticated clustering algorithm on our approximate embedding.
Though the robust Gaussian clustering problem is well-studied, the works that we are aware of consider contamination models in which a small fraction of the data points have been corrupted arbitrarily (e.g. \cite{DKKLMS18,HL18,KSS18}).
Here, we are instead interested in the case where all of the points may be corrupted, but the corruption overall is bounded in operator norm.

\paragraph{Comparison to the Stochastic Block Model.}
Of interest is how our results compare to the classic two-community stochastic block model (SBM).
The distribution $\mathrm{SSBM}(n,2,A,B)$ is defined as follows:
every vertex $i \in [n]$ is assigned a community label independently from $\Unif(\{\pm 1\})$, and then each edge $(i,j)$ is added independently with probability $A$ if nodes $i,j$ belong to the same community, and with probability $B$ otherwise.
The signal-to-noise ratio in this model can be expressed as $\lambda(n,A,B) = \sqrt{\frac{n(A-B)^2}{2(A+B)}}$.

The clustering and testing problems have been well-studied in the SBM (embedding has no analogue).
Here we focus on the $\Omega(\log n)$-average-degree results, as they are most directly comparable to our setting.
For clustering, almost exact recovery (a $1-o_n(1)$ fraction of the vertices are labelled correctly) can be achieved if and only if $\lambda(n,A,B) = \omega(1)$ \cite{yun2014community,mossel2014consistency,AS15}.
Some works have studied the performance of spectral algorithms for clustering in SBMs, including \cite{boppana1987eigenvalues,mcsherry2001spectral,choi2012stochastic,vu2018simple,rohe2011spectral}.
The setting in each of these works differs slightly, but the bottom line is that exact recovery (where all vertices are labeled correctly) by spectral algorithm is possible if $\lambda(n,A,B) = \Omega(\sqrt{\log{n}})$.

For the task of testing, the null hypothesis is the \erdos-\renyi graph $\bG(n,\frac{A+B}{2})$ with $n$ vertices and edge probability $\frac{A+B}{2}$.
It is known that if $\lambda(n,A,B) >1$, then consistent testing is possible, in both the bounded degree case \cite{MNS15} ($nA$ and $nB$ are constants as $n \to \infty$) and in the growing degree case \cite{janson1995random,banerjee2018contiguity} ($nA, nB \to \infty$ as $n \to \infty$).
Furthermore, there are tests with polynomial time complexity in both cases \cite{MNS15,banerjee2017optimal,montanari2016semidefinite}.
When $\lambda(n,A,B) <1$, the two distributions are asymptotically mutually contiguous, thus there is no consistent test.

In order to compare with the stochastic block model, we first match the parameters in our setting to the SBM case. In the GMBM setting, when \muall, the within-community edge probability is $A_{\mathrm{GMBM}} = p+\Theta(p\tau \mu^2 d)$, and the across-community edge probability is $B_{\mathrm{GMBM}} = p-\Theta(p\tau \mu^2 d)$.
A calculation shows that if we plug these edge probabilities in the SBM signal-to-noise ratio function, $\lambda(n,A_{\mathrm{GMBM}},B_{\mathrm{GMBM}}) = \Theta(\sqrt{npd \log(1/p)}\mu^2)$.
Requiring $\lambda(n,A_{\mathrm{GMBM}},B_{\mathrm{GMBM}}) \gg 1$ is equivalent (up to polylog factors) to one of our requirements for testing, that $\mu \gglog (npd \log(1/p))^{-1/4}$.
Our second requirement in the GMBM, that $\mu \gglog d^{-3/4}$, is a result of the geometric structure of the model, and is the dominant term in the maximum only when $d^2\ll np$; it is plausible that the signal-to-noise ratio for testing in the GMBM is indeed impacted by the geometry, and differs in this lower-dimensional setting.

For clustering, the requirement that $\mu \gglog d^{-1/2}$ is stricter than $\lambda(n,A_{\mathrm{GMBM}},B_{\mathrm{GMBM}}) = \omega(1)$ when $np \gg d$.
However, this requirement derives from the fact that even the Gaussian mixture itself is not clusterable when $\mu \ll d^{-1/2}$, so even given perfect access to the Gaussian mixture embedding this clustering task is impossible.
For $d \gg np$, we anticipate that the threshold will be the same (or similar) for both the GMBM and the SBM.
Our results primarily focus on the moderate range where $d\ll np$.

\subsection{Directions for future research}
\label{sec:future}
Our work makes an initial study of algorithmic tasks in a basic high-dimensional geometric block model.
It is our hope that this is merely a small early step, and that GMBMs will become a standard model organism for network science.
Here, we highlight a couple of intriguing directions for future research.
\begin{enumerate}
\item \textit{Characterize the information-computation landscape of the Gaussian Mixture block model.}

Here, we have given polynomial-time algorithms for this basic geometric block model that work so long as certain conditions are met: $1\lllog d \lllog pn$ in all cases, and $d^{-1/2} \lllog\mu$ for clustering (save for a logarithmic-scale window around $\mu = d^{-1/4}$, also present for embedding, where we suspect a more careful analysis might succeed), and $d^{-3/4} + (npd)^{-1/4} \llog \mu$ for hypothesis testing.
 But our understanding of the information-theoretic landscape for these problems is incomplete.

It seems that a natural requirement for the success of spectral algorithms is $d = O(nH(p))$ for $H(\cdot)$ the binary entropy function.
This is because the order of random fluctuations in the spectrum of the adjacency matrix is at least $\sqrt{np}$, and the $d$-dimensional embedding has eigenvalues on the order of $np\tau = \Theta(np \sqrt{d^{-1} \log 1/p})$, so one should only expect spectral embedding to succeed when $np \tau > \sqrt{np}$, that is when $d = O(n H(p))$.
In the sparse regime $np = O(1)$, the fluctuations of the adjacency matrix are actually of higher order than $\sqrt{np}$ (which is why our algorithm requires $np = \Omega(\polylog n)$), but it is possible that the precise threshold $d \sim nH(p)$ could be achieved by the non-backtracking matrix or a spectral algorithm in the style of \cite{HS17}.
It is unclear to us whether \textit{any} algorithm, polynomial-time or otherwise, can succeed in the regime $d\gg pn$.

In the stochastic block model, extensive study has been made of the information and computational landscape of the clustering problem, and beautiful conjectures from statistical physics have been confirmed by an elegant mathematical theory to establish the existence of \textit{\em information-theoretic} and \textit{computational} phase transitions.
When the signal-to-noise ratio in the SBM (a function of the inter- and intra-community edge probabilities and the number of communities) is below the Kesten-Stigum threshold, the hypothesis testing problem is believed to be computationally hard; there is also an information-theoretic threshold below which it is impossible to tell a graph generated from the SBM apart from an \erdos-\renyi graph\footnote{The information-theoretic threshold coincides with the Kesten-Stigum transition in the $2$-community case \cite{DKMZ11,Abbe17}, see also the more recent \cite{MSS22}.}.

It seems plausible that geometric block models exhibit a similarly rich computational landscape, both for clustering and separately for embedding; we feel that charting this landscape is an exciting direction for future research.
The excitement is deepened by the fact that the mathematical tools used in the context of the stochastic block model seem ill-suited to this more geometric setting (see also \cite{EMP22} for some discussion).
To our knowledge, the field is wide open both on the algorithmic/mathematical side and also from the perspective of predictions in statistical physics.

\item \textit{Understand the performance of spectral (and other) algorithms for GMBMs generated by a wider class of Gaussian mixtures.}

Though the GMBM that we have studied in this paper is certainly more realistic than, for example, the stochastic block model, it is still far from capturing most real-world settings.
Recall that the underlying Gaussian mixture is supposed to model a distribution over the feature space of network nodes.
Here we have only studied the case of a mixture of at most two perfectly spherical Gaussians.

It would be interesting to understand in which scenarios the spectral embedding algorithm continues to work if the number of communities is larger than two.
More problematically, it seems unlikely that spectral embedding will succeed out-of-the-box when the Gaussian covariances are far from being spherical; this is even more so true for clustering, as spectral algorithms are known to fail even when given access to the true Gaussian mixture model in the non-spherical case \cite{AM05}.
Is it possible to design other algorithms for embedding and clustering for this more general ``model organism,'' and would such algorithms yield insights which would transfer well to practice?

\end{enumerate}

\section{Technical overview}
At a high level, the adjacency matrix can be viewed as a kernel matrix whose entries are given by a threshold function of the latent inner products $\langle u_i,u_j\rangle$. Although the threshold function is discontinuous, it is fixed and can be expanded in a polynomial basis (in our analysis we use Gegenbauer polynomials). The zeroth-order term of this expansion corresponds to a constant matrix, whose eigenvalue is on the order of the average degree of the graph. The first-order term is linear in the inner products and therefore encodes the Gram matrix of the latent vectors $(\langle u_i, u_j\rangle)_{1\le i,j\le n}$, which has rank $d$. Higher-order terms contribute only to the error. Ignoring these higher-order terms, the adjacency matrix behaves like a random rank-$(d+1)$ matrix whose eigenvectors recover the span of the latent vectors. Consequently, the eigenvectors corresponding to the second through $(d+1)$-st eigenvalues reconstruct the latent vectors up to normalization and rotation.

This heuristic suggests estimating the latent vectors from the top eigenvectors of the adjacency
matrix. We now elaborate on this idea in more detail.

Recall that we have let $A$ denote the adjacency matrix of $G$ and that $(\eta_i,w_i)$ are the eigenvalues and corresponding unit eigenvectors of $A$, where $\eta_0 \ge \cdots \ge \eta_{n-1}$. Define the vectors $\hat{u}_1,\ldots,\hat{u}_n \in \R^d$ by setting
\begin{align*}
    \hat u_j(i):= \sqrt{\frac{\max(\eta_{i}, 0)}{\td \lambda_1}}w_{i}(j),
\end{align*}
where $\td$ and $\lambda_1$ are to be specified later. Define the $n\times d$ matrix $\sU$ by putting the vectors $\hat u_1, \ldots, \hat u_n$ in its rows. In other words, $\td \lambda_1 \sU \sU^\top$ is the spectral truncation of $A$ to the span of the eigenvectors corresponding to its second through $(d+1)$-st eigenvalues.
Similarly, define the $n\times d$ matrix $U$ by putting the true latent vectors $u_1, \ldots, u_n$ in its rows.
The key is to show that the $\hat{u}_i$ approximate the latent $u_i$, spectrally.
For technical reasons, we treat the case when $\mu$ is small and large separately.
When $\mu$ is small, we show:
\begin{theorem}\label{thm:testing}
    Suppose $d,n \in \Z_+$, $p \in [0,1/2-\varepsilon]$ for any constant $\eps > 0$, and $\mu \geq 0$ satisfy the conditions \condsmall. Then
    \begin{align*}
        \|\sU \sU^\top-U U^\top\|_{\mathrm{op}} \ll \max \left\{ \frac{n\tau}{d}, \frac{\sqrt{n}}{\sqrt{p} d\tau} \right\} \log^9 (n).
    \end{align*}
\end{theorem}
\noindent For reference (and for comparison with the theorem statements in the introduction), when $\mu \lllog d^{-1/4}$ is not too large and in the high-dimensional regime $d = \Omega(\log n)$, $\tau = \Theta(\ssqrt{\frac{1}{d}\log1/p})$.
This can be seen by noting that the distribution of $\iprod{u_i,u_j}$ is close to $\calN(0,\frac{1}{d})$, and so $\Pr[\iprod{u_i,u_j} \ge \tau] \approx \exp(-d\tau^2/2)$.

Proving \pref{thm:testing} amounts to showing that the top $d+1$-dimensional eigenspace of $A$ is spanned by the columns of $U$ and a non-negative vector $\topev \in \R^n$ (whose entries scale as a function of the length of the corresponding $u_i$). Here $\mathbbm{1}_n$ denotes the all-ones vector, and $\topev$ is a weighted version of it.
Specifically, we show:
\begin{proposition}\label{p:op1}
Suppose $d,n \in \Z_+$, $p \in [0,1/2-\varepsilon]$ for any constant $\eps > 0$, and $\mu \geq 0$ satisfy the conditions \condsmall. Then there exist a length-$n$ vector $\tilde{\mathbbm{1}}_n$ and scalars $p_0$, $\tilde d$, and $\lambda_1$, to be defined later, such that with high probability,
\begin{align*}
    \| A- p_0\tilde{\mathbbm{1}}_n \tilde{\mathbbm{1}}_n^\top - \td \lambda_1 UU^\top\|_{\mathrm{op}} \ll  \log^9(n) \max\left\{ np\tau^2, \sqrt{np} \right \}.
\end{align*}
\end{proposition}

The above is enough to imply that $\topev$ is close to the top eigenvector of $A$ and the columns of $U$ are close to the span of the next $d$ eigenvectors of $A$ so long as
\begin{align*}
p_0\|\topev\|^2 \asymp np \gg \sigma_{\max}(\tilde{d} \lambda_1 UU^\top) \ge \sigma_{\min}^{+}(\tilde{d} \lambda_1 UU^\top) \gg \max\left\{np\tau^2, \sqrt{np}\right\}\log^9 n,
\end{align*}
where $\sigma_{\min}^{+}$ denotes the smallest nonzero singular value and $\sigma_{\max}$ the largest singular value.
Applying spectral concentration for sample covariance matrices, the smallest nonzero singular value of $\tilde{d}\lambda_1 UU^\top$ is of order $np\tau$ with high probability; its largest singular value remains below the degree spike in our parameter regime. Thus $A$'s top eigenspace is well-approximated by $U$ when $\frac{1}{\sqrt{np}} \lllog \tau \lllog 1 \iff 1 \lllog d \lllog np$ since $\tau = \Theta(\ssqrt{\frac{1}{d}\log 1/p})$; this is the source of our upper bound on $d$.

\paragraph{Linear approximation of the adjacency matrix.}
The $i,j$-th entry of $A$ is a function of the inner product of the latent $u_i,u_j$:
\begin{align*}
A_{i,j} = A_{i,j}(\iprod{u_i,u_j}) = \Ind[\iprod{u_i,u_j} \ge \tau].
\end{align*}
We can understand the matrix $p_0\topev \topev^\top + \tilde{d} \lambda_1\cdot UU^\top$ subtracted in \pref{p:op1} as a linear approximation of $A$ in the inner products $\iprod{u_i,u_j}$.
Intuitively, if we were to express $\Ind[\iprod{u_i,u_j} \ge \tau]$ as a polynomial in $\iprod{u_i,u_j}$, we'd see that $p_0\tilde{\mathbbm{1}}_n \tilde{\mathbbm{1}}_n^\top$ is roughly the zeroth order term of the polynomial and $\td \lambda_1 UU^\top$ is the first order term.
The error term $np\tau^2$ in \pref{p:op1} reflects the fact that the second order coefficient in the polynomial expansion of $A_{i,j}$ is quadratic in $\tau$ (the other error term, $\sqrt{np}$, comes from random fluctuations).
So in effect, we want to show that $A$ is well-approximated by its linear term in a polynomial basis.

Our proof will proceed by applying the trace method.
We will expand $A$'s entries in the basis of Gegenbauer polynomials, which is a basis of polynomials that enjoys nice orthogonality properties when evaluated on inner products of random vectors on the unit sphere (see \pref{sec:geg} for details).
In the high-dimensional setting $d\gg 1$, the latent vectors $u_i$ lie roughly on the sphere $\calS^{d-1}$, so we can write $u_i = v_i + \text{error}$, where each $v_i \sim\unif(\calS^{d-1})$.
Therefore, we can write
\begin{align*}
    A_{i,j} = \Ind(\inp{u_i,u_j}\geq \tau) = \Ind(\inp{v_i,v_j}\geq \tau^{i,j})
\end{align*}
for some $\tau^{i,j}$ close to $\tau$.
Ignoring the difference between $\tau^{i,j}$ and $\tau$ for the moment, we then expand the threshold function $\Ind(\cdot \geq \tau)$ in the Gegenbauer polynomial basis $q_0,q_1,\ldots$ (in this proof overview, the $q_k$ are implicitly renormalized to ease notation), so we have that
\begin{align*}
A_{i,j} = \sum_{k=0}^\infty c_k q_k(\iprod{v_i,v_j}),
\end{align*}
    and our goal now reduces to showing that when we subtract the $k=0$ and $k=1$ terms, the operator norm of the resulting matrix $A_{\ge2}$ with $A_{\ge 2}(i,j) = \sum_{k=2}^\infty c_k q_k(\iprod{v_i,v_j})$ is bounded.

\paragraph{The trace method.}
The trace method relates the maximum eigenvalue of a matrix to the expectation of a power's trace (using Markov's inequality): for any integer $\ell$,
\begin{align*}
\|M\| > t\implies \tr(M^{2\ell}) > t^{2\ell},\end{align*}
so
\begin{align*}\Pr[\|M\| > e^{\eps} \E[\tr(M^{2\ell})]^{1/2\ell}] \le \Pr\left[\tr(M^{2\ell}) > e^{2\ell\eps} \E[\tr(M^{2\ell})]\right]\le e^{-2\ell\eps}.
\end{align*}
So choosing, say, $\eps = \frac{1}{\log n}$ and an integer $\ell$ of order $\log^{3/2} n$ gives us that $\|M\| \le (1+o(1)) \E[\tr(M^{2\ell})]^{1/2\ell}$ with high probability.

The trace method thus allows us to relate the operator norm, an analytic quantity, to degree-$2\ell$ moments of entries of a random matrix.
Specifically, we can relate the trace of a power of $M$ to expected value of products over ``walks'' of length $2\ell$ in $K_n$ weighted by the expected product of the edge ``weights'' given by the entries of $M$:
\begin{align*}
\E\tr(M^{2\ell}) = \sum_{i_1,\ldots,i_{2\ell} \in [n]} \E\left[\prod_{s = 1}^{2\ell} M_{i_s,i_{s+1}}\right],
\end{align*}
where the subscript $s+1$ is understood to be taken modulo $2\ell$.

We apply the trace method with $M = A_{\ge 2}$, and our goal becomes to upper bound $\E[\tr(A_{\ge2}^{2\ell})]$ by a quantity scaling like $\tilde{O}(np\tau^2)^{2\ell}$ for $\ell = \polylog n$.
To analyze the expected trace, we make use of the orthogonality properties of Gegenbauer polynomials evaluated on inner products of random vectors on $\calS^{d-1}$.
We have
\begin{align*}
\E \tr (A_{\ge 2}^{2\ell})
= \sum_{i_1,\ldots,i_{2\ell}\in [n]} \E\left[\prod_{s=1}^{2\ell}\sum_{k=2}^\infty c_k q_k(\iprod{v_{i_s},v_{i_{s+1}}})\right]
= \sum_{\substack{i_1,\ldots,i_{2\ell} \in [n] \\k_1,\ldots,k_{2\ell} \ge 2}} \E\left[\prod_{s=1}^{2\ell} c_{k_s} q_{k_s}(\iprod{v_{i_s},v_{i_{s+1}}})\right]
\end{align*}
The orthogonality properties of the Gegenbauer polynomials will (at a high level) allow us to eliminate summands which contain any terms of different orders, $k_s \neq k_{s'}$, unless the indices $i_{s},i_{s'}$ appear with high multiplicity.
This is very helpful in our accounting and allows us to show that the $c_2 \approx p\tau^2$ coefficients more-or-less dominate the summation.
The all-$c_2$ term in which no $i_s$ is repeated has a contribution bounded by $d^2 n^{2\ell} c_2^{2\ell}$, and since this is roughly the dominant\footnote{We lose polylogarithmic factors because we do not show that it is completely dominant; we suspect that a more careful argument would be able to establish full dominance and eliminate the polylogs.} term when $\ell = \polylog(n)$ and $np\tau^2 \gg \sqrt{np}$,\footnote{
When $np\tau^2 \ll \sqrt{np}$, terms where indices $i_s$ appear with high multiplicity dominate, which gives the bound $\|A_{\ge 2}\| \le \tilde{O}(\sqrt{np})$.}
we get that with high probability $\|A_{\ge 2}\| \le \polylog n \cdot \left(d^2 n^{2\ell} c_2^{2\ell}\right)^{1/2\ell} =  n c_2 \polylog n = np\tau^2 \polylog n$, giving us the correct order of magnitude for the error.

\paragraph{Accounting for large separation.}
Recall that (though we have momentarily ignored this detail) the coefficients in the polynomial expansion of $A_{ij}(\iprod{v_i,v_j}) = \Ind[\iprod{v_i,v_j} \ge \tau^{i,j}]$ depend on $\tau^{i,j}$ as well, so really $A_{ij} = \sum_{k=0}^\infty c_k^{i,j} q_{k}(\iprod{v_i,v_j})$.
The coefficients $c_k^{i,j}$ concentrate well when the separation $\mu$ is small, but when $\mu$ is large the analysis is slightly more involved when we relate the Gaussian mixture to the sphere. We prove a similar result with an application of the trace method:
\begin{theorem}\label{thm:testing2}
    Suppose $d,n \in \Z_+$, $p \in [0,1/2-\varepsilon]$ for any constant $\eps > 0$, and $\mu > 0$ satisfy the conditions \condbig. Then, with high probability,
    \begin{align*}
        \|\sU \sU^\top-U U^\top\|_{\mathrm{op}} \ll \max \left\{ \frac{n\mu^4}{\tau}, \frac{\sqrt{n}}{\sqrt{p} d\tau} \right\} \log^9 (n).
    \end{align*}
\end{theorem}
\noindent As in the small $\mu$ case, we prove \pref{thm:testing2} through the following proposition, using the same strategy that applied in the small $\mu$ case.
\begin{proposition}\label{p:largemu}
Suppose $d,n \in \Z_+$, $p \in [0,1/2-\varepsilon]$ for any constant $\eps > 0$, and $\mu > 0$ satisfy the conditions \condbig. Then there exist a length-$n$ vector $\tilde{\mathbbm{1}}_n$ and scalars $p_0$, $\td$, and $\lambda_1$, to be defined later, such that with high probability,
\begin{align*}
    \| A- p_0\tilde{\mathbbm{1}}_n \tilde{\mathbbm{1}}_n^\top - \td \lambda_1 UU^\top\|_{\mathrm{op}} \ll  \max\left\{ npd\mu^4,  \sqrt{np} \right \} \log^9(n).
\end{align*}
\end{proposition}

\paragraph{Hypothesis testing and clustering.}
Once we have a good approximation to the latent embedding vectors, we can use them to hypothesis test and to cluster.
For clustering, we show using standard matrix concentration techniques that the top eigenvector of the positive semidefinite matrix $UU^\top$ in $\R^n$ must have signs which closely match the cluster labeling, and therefore by a classic eigenvector perturbation argument (the Davis--Kahan theorem) the same is true of $\sU\sU^\top$ because $\|UU^\top - \sU\sU^\top\|$ is small.
For hypothesis testing, we show that if $\mu$ is large enough, $U$ and $\sU$ have a spectral gap, and thus $\eta_1$ furnishes a good hypothesis test.

\paragraph{Estimating the latent dimension $d$.}
The spectral structure described above also allows one to recover the latent dimension when it is unknown. The degree direction and the mixture-mean direction may create large early spectral gaps, but under an explicit signal-to-noise condition the largest gap $\eta_i-\eta_{i+1}$ over $2\leq i\leq n-2$ occurs at $i=d$. This gives an estimator of the latent dimension. A formal statement and proof are given later in Section~\ref{sec:unknownd}.

\section{Preliminaries and notation}
\label{sec:prelim}
We use standard asymptotic notation. For deterministic sequences $A_n$ and positive sequences $B_n$, we write $A_n=O(B_n)$ if $|A_n|\leq C B_n$ for all sufficiently large $n$, for some constant $C<\infty$. We write $A_n=o(B_n)$ or $A_n\ll B_n$ if $|A_n|/B_n\to0$, and use $\Omega$, $\omega$, and $\Theta$ analogously. We write $A_n=\tilde O(B_n)$ if $|A_n|\leq B_n\log^C n$ for all sufficiently large $n$ and some constant $C$; the notation $\tilde\Omega$ and $\tilde\Theta$ is defined similarly.

For any $A_n, B_n$, we use $A_n \llwhp B_n$ to denote that for every constant $\varepsilon>0$, we have $|A_n| \leq \varepsilon B_n$ for large enough $n$ with high probability.

For any $n \times n$ matrix $B$, we define $\diag(B)$ to be the $n \times n$ matrix with the same diagonal as $B$ and all zero entries off-diagonal.
The notation $\|B\|$ and $\|B\|_{\text{op}}$ denote the operator norm of $B$.

We use $\Ind(\cdot)$ to denote the indicator function.

If we don't specify in the setting, we always assume that $n,d\in \Z_+$, $\mu \in \R_+$, $p \in [0,1/2-\varepsilon]$ for any constant $\eps>0$, $\log^{16} n\ll d < n $ and $pn\gg 1$. This is the regime we focus on.

\subsection*{Gegenbauer polynomials}\label{ss:gegenbauer}
For $u,w \sim \unif(\cS^{d-1})$, we denote by $\cD_d$ the law of $\sqrt{d}\cdot \inp{u,w}$ (scaling by $\sqrt{d}$ ensures that $\E[(\sqrt d\,\iprod{u,w})^2] = 1$).
The \textit{Gegenbauer polynomials} are an orthonormal basis for functions in $L^2([-\sqrt{d},\sqrt{d}],\cD_d)$.
The Gegenbauer polynomials can be obtained via application of the Gram-Schmidt process to the monomial basis, so that they naturally form a sequence of polynomials $\{q_\ell^{(d)}\}_{\ell \in \N}$, increasing in degree so that $\deg(q_\ell^{(d)}) = \ell$.
For instance, the first three Gegenbauer polynomials are
\begin{align*}
    q_0^{(d)}(x) = 1, \quad q_1^{(d)}(x) = x, \quad \text{and} \quad q_2^{(d)}(x) = \frac{1}{\sqrt{2}} \sqrt{\frac{d+2}{d-1}} (x^2-1).
\end{align*}
The orthogonality of the Gegenbauer polynomials is equivalent to the property that
\begin{align*}
\E_{x \sim \cD_d}[q_\ell^{(d)}(x)q_k^{(d)}(x)] = \Ind[k=\ell].
\end{align*}

The Gegenbauer polynomials are related to the \textit{spherical harmonics}, an orthonormal basis for functions on $\cS^{d-1}$.
We write $\{\phi_{\ell,t}(u)\}_{t\in [N_\ell^{(d)}]}$ as the spherical harmonics of degree $\ell$ associated to $u$, (and we use $N^{(d)}_\ell$ to denote the cardinality of the orthonormal degree-$\ell$ spherical harmonics associated with a fixed vector on $\cS^{d-1}$).
It is known that
\begin{align*}
    N_\ell^{(d)} = \frac{2\ell + d - 2}{\ell} \binom{\ell+d-3}{\ell-1}.
\end{align*}
If $u \sim \unif(\cS^{d-1})$, orthonormality of the spherical harmonics implies that
\begin{align*}
\E_{u}\left[\phi_{\ell_1,t_1}(u) \phi_{\ell_2,t_2}(u) \right] = \Ind[\ell_1=\ell_2, t_1 = t_2],
\end{align*}
for any $\ell_1, \ell_2 \in \mathbb{Z}_{\geq 0}$ and $t_1, t_2 \in [N_\ell^{(d)}]$.

The spherical harmonics are related to the Gegenbauer polynomials through the addition theorem
\begin{align*}
    q_\ell^{(d)}(\sqrt{d}\cdot \iprod{u,v}) = \frac{1}{\sqrt{N_\ell^{(d)}}} \sum_{t\in [N_\ell^{(d)}]} \phi_{\ell,t} (u)  \phi_{\ell,t} (v),
\end{align*}
See \cite{dai2013approximation,efthimiou2014spherical} for a proof of this statement.
The addition theorem and the orthonormality of the spherical harmonics together imply the following remarkable property: if $v \sim \unif(\cS^{d-1})$, then
\begin{align*}
\E_{v}\left[q^{(d)}_\ell(\sqrt{d}\cdot \iprod{u,v})\cdot q^{(d)}_k(\sqrt{d}\cdot \iprod{v,w})\right] = \Ind[k=\ell] \cdot\frac{1}{ \sqrt{N^{(d)}_\ell}} \cdot q^{(d)}_\ell(\sqrt{d}\cdot \iprod{u,w}).
\end{align*}
Further, from the orthonormality of the spherical harmonics we can derive that
\begin{align*}
    q_\ell^{(d)}(\sqrt{d})
= \E_u q_\ell^{(d)}(\sqrt{d}\iprod{u,u}) = \frac{1}{\sqrt{N_\ell^{(d)}}} \sum_{t\in [N_\ell^{(d)}]} \E_u \phi_{\ell,t} (u)^2 = \sqrt{N_\ell^{(d)}}.
\end{align*}
These identities follow from standard properties of spherical harmonics; see e.g. \cite{atkinson2012spherical}.

\section{Proofs of the main results}\label{sec:proofs}
\subsection{Polynomial expansion of the indicator function}\label{sec:geg}
We will expand the threshold function $\Ind(x\geq \sqrt{d}\tau)$ in the basis of $d$-dimensional Gegenbauer polynomials.
As our vectors $u_1,\ldots,u_n$ are sampled from a Gaussian mixture distribution, the Gegenbauer polynomials are no longer an orthogonal basis for $\iprod{u_i,u_j}$.
To correct for this, we begin by shifting and rescaling our vectors.

For each $u_i$, we define the first entry of it to be $a_i$ and let the remaining $d-1$ coordinates form the $(d-1)$-vector $w_i$.
Furthermore, let $\ell_i = \|w_i\|$ and write $w_i = \ell_i v_i$.
We then have that $u_i = (a_i, \ell_i v_i)$ where each $v_i$ is a unit vector.
For each $i\neq j \in [n]$, the $(i,j)$-th entry of the adjacency matrix is now given by
\begin{equation}\label{eq:v}
    A_{i,j} = \Ind(\inp{u_i,u_j}\geq \tau) = \Ind(a_ia_j+\ell_i\ell_j\inp{v_i,v_j}\geq \tau) = \Ind\bigp{\inp{v_i,v_j}\geq \frac{\tau-a_ia_j}{\ell_i \ell_j}}.
\end{equation}
For notational convenience, call
\begin{equation}\label{eq:deftauij}
    \tau^{i,j} = \frac{\tau-a_ia_j}{\ell_i \ell_j}.
\end{equation}
We will later show that the $\tau^{i,j}$ are well-concentrated around $\tau$, with $\tau^{i,j} \approx (\tau \pm \mu^2)(1\pm d^{-1/2})$.
The dimension of the $v_i$ is now $d-1$, and in what follows we write $\td = d-1$ for simplicity.

We now expand the threshold function $\Ind(x \geq \sqrt{\td}\tau)$ as well as the threshold function corresponding to each $(i,j)$ entry, $\Ind(x\geq \sqrt{\td}\tau^{i,j})$, in the $\td$-dimensional sphere $\cS^{\td-1}$ and define
\begin{align*}
   \Ind(x\geq \sqrt{\td} \tau) = \sum_{k=0}^\infty c_k q^{(\td)}_k(x), \quad \text{and} & \qquad \Ind(x\geq \sqrt{\td} \tau^{i,j}) = \sum_{k=0}^\infty c_k^{i,j} q^{(\td)}_k(x).
\end{align*}
Importantly, we further define
\begin{equation}\label{eq:deflambda}
    \lambda_k = \frac{c_k}{\sqrt{N_k}}\quad \text{and}  \quad \lambda_k^{i,j} = \frac{c_k^{i,j} }{\sqrt{N_k}},
\end{equation}
where $N_k = N_k^{(\td)}$ is the cardinality of the orthonormal degree-$k$ spherical harmonics associated with any fixed vector on the sphere $\cS^{\td-1}$, as introduced in \pref{ss:gegenbauer}.
As a convention, we write $p_0:=\lambda_0$ and $p_0^{i,j}:=\lambda_0^{i,j}$.
 In the rest of this subsection and the next, we write $q_k = q_k^{(\tilde d)}$, omitting the superscript for simplicity.

\subsection{The trace method}\label{sec:trace}
In order to prove \pref{p:op1}, we apply the trace method to $A$ minus a linear approximation to its top eigenspace in terms of the unit-vector inner products $\iprod{v_i,v_j}$; this will let us better exploit orthogonality properties of Gegenbauer polynomials.
The constant-order term which we subtract will not be a rank-1 matrix; we'll correct for this (accounting for the difference between Gaussian and spherical vectors) later in \pref{sec:g-sph}.

For simplicity, we adopt a notation and write $[a_{i,j}]_{0, n\times n}$ to denote the $n \times n$ matrix where each off-diagonal entry equals $a_{i,j}$ and the diagonal equals $0$.
\begin{proposition}\label{p:op2}
For \muall, and any $i,j\in [n]$, we have that
\begin{align*}
    \| A-[p_0^{i,j}]_{0, n\times n} - [\td \lambda_1^{i,j} \inp{v_i,v_j}]_{0, n\times n} \|_{\mathrm{op}} \ll \log^9 n \max\left\{ np\tau^2, \sqrt{np} \right\},
\end{align*}
with high probability as $n$ goes to infinity.
\end{proposition}
To simplify notation, we define the $n \times n$ matrix $Q$ to be the left hand side in \pref{p:op2}. So this implies that when $i\neq j$,
\begin{align*}
    Q_{i,j} & = \Ind(\inp{v_i,v_j}\geq \tau^{i,j}) - p_0^{i,j} - \td \lambda_1^{i,j} \inp{v_i,v_j}.
\end{align*}
And $Q_{i,i} = 0$ for any $i\in [n]$.
The rest of the subsection will be devoted to the proof of \pref{p:op2}. Now we briefly recall the statement of the trace method.
\begin{lemma}[Trace Method]\label{lem:trace}
    Let $M$ be a symmetric matrix. Then for any positive even integer $\ell \geq 2$,
    \begin{align*}
        \Pr(\|M\|\geq e^\ep \E[\tr(M^\ell)]^{1/\ell}) \leq \exp(-\ep \ell).
    \end{align*}
\end{lemma}
We will proceed to compute $\E[\tr(Q^\ell)]$ for an even integer $\ell$.
This amounts to bounding the expectation of a sum over closed walks of length $\ell$ in the complete graph $K_n$ when weighted by entries of $Q$:
\begin{align*}
    \E[\tr ( Q^\ell)] = \sum_{i_1, \cdots, i_\ell \in [n]} \E \bigp{\prod_{t=1}^\ell Q_{i_{t},i_{t+1}} },
\end{align*}
where we identify $i_{\ell+1}$ with $i_1$.
We'll associate each  sequence $\vec i = (i_1, \cdots, i_\ell) \in [n]^\ell$ with a (multi-)graph $\bH_{\vec i}$ (often we will drop the subscript $\vec{i}$).
We define the set of vertices in $\{i_1, \cdots, i_\ell\}$ as the vertex set and put an edge between $i_t$ and $i_{t+1}$ for any $1\leq t \leq \ell$, where again we identify $i_{\ell+1}$ with $i_1$, allowing multi-edges.

Note that the diagonal of the matrix $Q$ consists of all zero entries, so we only need to consider multi-graphs with no self loops. Furthermore, because $i_1, \cdots, i_\ell$ is a closed walk, all vertices in $\bH_{\vec{i}}$ have even degree.

When we take the expectation over the vector $v_{i_j}$ for any degree-2 vertex $i_j \in \bH$, it can be contracted at the cost of a shrinking factor:

\begin{lemma}[Contracting degree-2 vertices]\label{lem:contract}
    In a path $s_1, \cdots, s_{t+1}$ of length $t\geq 2$ in which $s_2,\ldots,s_{t}$ have degree $2$ in $\bH$, in expectation over the randomness of $v_{s_2},\ldots,v_{s_{t}}$ we have,
\begin{align*}
    \E_{v_{s_2},\ldots,v_{s_t}} \bigp{ \prod_{a=1}^{t} Q_{s_a, s_{a+1}}} = \sum_{k=2}^\infty  q_k (\sqrt{\td}\inp{ v_{s_1}, v_{s_{t+1}} })\bigp{ \prod_{a=1}^{t} \lambda_k^{s_a,s_{a+1}}} \sqrt{N_k}
\end{align*}
\end{lemma}
\begin{proof}
\begin{align*}
    &\E_{v_{s_2},\ldots,v_{s_t}} \bigp{ \prod_{a=1}^{t} Q_{s_a, s_{a+1}} \bigmid v_{s_1}, v_{s_{t+1}}} \\
    &= \E_{v_{s_2},\ldots,v_{s_t}} \bigp{ \prod_{a=1}^{t} \bigp{ \sum_{k=2}^\infty \lambda_k^{s_a, s_{a+1}} \sqrt{N_k} q_k(\sqrt{\td}\inp{ v_{s_a}, v_{s_{a+1}} })  } }\\
    & =  \sum_{k_1, \cdots, k_t =2}^\infty \prod_{a=1}^t  \lambda_{k_a}^{s_a, s_{a+1}} \sqrt{N_{k_a}} \E_{v_{s_2},\ldots,v_{s_t}} \bigp{ \prod_{a=1}^t q_{k_a}(\sqrt{\td}\inp{ v_{s_a}, v_{s_{a+1}} }) }.
\end{align*}
The exchange of the limit and the expectation is justified by the standard dominated convergence theorem. Note that by the properties of the Gegenbauer polynomials given in \pref{ss:gegenbauer}, the product is only nonzero when all $k_a$ are the same, and further
\begin{align*}
    \E_{v_{s_{a+1}}}\bigp{ q_{k_a}(\sqrt{\td}\inp{ v_{s_a}, v_{s_{a+1}} }) q_{k_a}(\sqrt{\td}\inp{ v_{s_{a+1}}, v_{s_{a+2}} })} = \frac{1}{{\sqrt{N_{k_a}}}} q_{k_a}(\sqrt{\td}\inp{ v_{s_a}, v_{s_{a+2}} }).
\end{align*}
By applying the above equation repeatedly, we have the lemma.
\end{proof}

If we begin with a cycle $s_1,\ldots,s_{t+1} = s_1$, and contract all of the degree-$2$ vertices in a cycle, then this produces a self-loop.
So we have the following as a corollary:
\begin{corollary}[Contracting a cycle to a self-loop]\label{cor:self-loop}
In an induced cycle $s_1,\ldots,s_{t+1} = s_1$ of length $t\geq 2$ in $\bH$, in expectation over the randomness of $v_{s_2},\ldots,v_{s_t}$, we have
\begin{align*}
    \E_{v_{s_2},\ldots,v_{s_t}} \bigp{ \prod_{a=1}^{t} Q_{s_a, s_{a+1}}} = \sum_{k=2}^\infty  q_k (\sqrt{\td})\bigp{ \prod_{a=1}^{t} \lambda_k^{s_a,s_{a+1}}} \sqrt{N_k}.
\end{align*}
\end{corollary}
\begin{proof}
This follows from \pref{lem:contract} and from the fact that if $v_{s_1} = v_{s_{t+1}}$ then $\iprod{v_{s_1},v_{s_{t+1}}} = 1$.
\end{proof}

We will obtain a ``contracted'' graph $\tilde\bH_{\vec{i}}$ from $\bH_{\vec{i}}$ as follows: as long as there exists either a vertex of degree $2$, contract it; otherwise if there exits a self-loop, remove it.
When the algorithm terminates we are left with the contracted (and potentially empty) graph $\tilde\bH_{\vec{i}}$.
If $\tilde\bH_{\vec{i}}$ is nonempty, then every vertex has degree at least 4: this is because (i) every vertex had even degree to begin with, (ii) all degrees have to be larger than two for the procedure to terminate, and (iii) our algorithm for producing $\tilde{\bH}_{\vec{i}}$ maintains the invariant that all degrees are even.
To see (iii) is true, note that so long as we follow the convention that a self-loop induces degree two, then vertex contractions do not change the degree of the non-contracted vertices, and further when a self loop is removed from a vertex the degree drops by two, therefore maintaining the invariant that the degree is even.

For any edge $e \in E(\tilde{\bH})$ which is the result of the contraction of a path $s_1,\ldots,s_{t+1} \in \bH$, define
\begin{align*}
\tilde{Q}_e = \sum_{k=2}^\infty q_k(\sqrt{\td}\iprod{v_{s_1},v_{s_{t+1}}}) \cdot\sqrt{N_k}\cdot \prod_{a=1}^t \lambda_k^{s_a,s_{a+1}},
\end{align*}
and for the sake of consistency if $e \in E(\tilde\bH) \cap E(\bH)$ then define $\tilde{Q}_e = Q_e$ in any case.
Note that this is a random variable depending only on $v_{s_1}$ and $v_{s_{t+1}}$ conditioned on the values $\lambda_k^{i,j}$.

Further, let $\calC(\vec{i})$ be the set of all cycles $C = (s_1,\ldots,s_{t+1} = s_1) \in \bH$ that were contracted into a self-loop and removed in producing $\tilde\bH$, and define
\begin{align*}
\tilde{Q}_C = \sum_{k=2}^\infty q_k(\sqrt{\td}) \cdot \sqrt{N_k} \cdot \prod_{a=1}^t \lambda_k^{s_a,s_{a+1}}.
\end{align*}
Note that in light of \pref{cor:self-loop} $Q_C$ is deterministic conditioned on the values $\lambda_k^{i,j}$.

From \pref{lem:contract} and \pref{cor:self-loop}, we have
\begin{align}
\E\left(\prod_{e \in E(\bH)} Q_e \right) = \prod_{C \in \calC} \tilde{Q}_C \cdot \E\left(\prod_{e \in E(\tilde{\bH})} \tilde{Q}_e\right). \label{eq:tilda-trans}
\end{align}
It remains to deal with the expectation over the $\tilde{Q}_e$ in $\tilde\bH$; here we will appeal to the fact that $|\tilde{Q}_e|$ are not too large whenever the inner products $\iprod{v_i,v_j}$ are not too large.
For any vertices $i\neq j \in [n]$, define the \emph{good} event $\cG_{(i,j)}$
\begin{align*}
    \cG_{(i,j)}:= \left \{  |\inp{v_{i},v_{j}}|< \tfrac{\log^2(n)}{\sqrt{\td}}  \right\},
\end{align*}
and let $\cG = \bigcap_{i\neq j \in [n]}\calG_{(i,j)}$ be the event that all such inner products are good.
We can bound the probability of $\overline{\calG}$ using the following lemma.
\begin{lemma}[Bound on inner products]\label{lem:badprob}
    For any $i\neq j \in [n]$, we have that
    \begin{align*}
        \Pr(\ol{\cG})\leq n^2 \cdot n^{-\log^3 (n)/3}.
    \end{align*}
\end{lemma}
\begin{proof}
    By direct computation with the density of $\cD_{\td}$,
    \begin{align*}
        \Pr(\ol{\cG_{(i,j)}})&= \frac{\Gamma(\td/2)}{\sqrt{\td \pi} \Gamma((\td-1)/2)} \int_{\log^2(n)}^{\sqrt{\td}} (1-\tfrac{\xi^2}{\td}) ^{\frac{\td-3}{2}} d\xi\\
        &\leq \sqrt{\tilde d}\exp\bigp{-\tfrac{\log^4(n)(\td-3)}{2\td}} \ll n^{-\frac{1}{3}\log^3 (n)}.
    \end{align*}
    By a union bound, we have the lemma.
\end{proof}
Now, it will be useful to divide $\tilde\bH$ into two (multi)graphs: the contracted-edges graph induced by the edges resulting from contraction, $\bK_{\vec{i}}$, and the ``uncontracted graph'' $\bU_{\vec{i}}$ which remains when the edges in $\bK$ are removed, with vertex set $V(\tilde\bH)$ and edge set $E(\tilde\bH) \setminus E(\bK)$.
This is because for a contracted edge $e$, $\|\tilde Q_e \Ind(\calG_e)\|_\infty$ decays with $p$, whereas for uncontracted edges in $E(\bU)$ we will have to argue differently.
We now have that
\begin{align}
&\prod_{C \in \calC} \tilde{Q}_C \cdot\E\left(\prod_{e \in E(\tilde{\bH}_{\vec{i}})}\tilde{Q}_e\right)\nonumber\\
& \le \prod_{C \in \calC} \tilde{Q}_C \cdot\E\left(\prod_{e \in E(\tilde{\bH}_{\vec{i}})}\tilde{Q}_e \Ind(\cG) \right) + \Pr(\overline{\cG})\cdot \|Q_e\|_\infty^\ell\nonumber\\
&\le \left| \prod_{C \in \calC} \tilde{Q}_C \right|\cdot\prod_{e \in E(\tilde\bH)\setminus E(\bH)} \|\tilde{Q}_e \Ind(\cG_e)\|_\infty \cdot \E\left(\prod_{e \in E(\tilde\bH)\cap E(\bH)} \left|Q_e \Ind(\calG_e)\right|\right)+ \Pr(\overline{\cG})\cdot \|Q_e\|_\infty^\ell \nonumber\\
&= \left| \prod_{C \in \calC} \tilde{Q}_C \right| \cdot\prod_{e \in E(\bK_{\vec{i}})} \|\tilde{Q}_e \Ind(\cG_e)\|_\infty \cdot \E\left(\prod_{e \in E(\bU_{\vec{i}})} \left|Q_e \Ind(\calG_e)\right|\right) + n^2 \cdot n^{-\log^3 n/3}\cdot (2 + \sqrt{\td})^\ell,\label{eq:bound}
\end{align}
where in the last line we used \pref{lem:badprob} and the following claim:
\begin{claim}
For every edge $e$,
\begin{align*}
\|Q_e\|_\infty  \le 2 + \sqrt{\td},
\end{align*}
\end{claim}
\begin{proof}
For $e = (i,j)$, expand
\begin{align*}
|Q_{i,j}| \le 1 + |p_0^{i,j}| + \td |\lambda_1^{i,j}|\,|\iprod{v_i,v_j}|.
\end{align*}
The bound follows from the fact that $|\iprod{v_i,v_j}| \le 1$, and $|p_0^{i,j}|\le 1$, $|\sqrt{\td}\lambda_1^{i,j}| \le 1$ and $|\iprod{v_i,v_j}|\le 1$ since each $(c_k^{i,j})^2 \leq 1$.
\end{proof}
Concerning the second term in \pref{eq:bound}, we will ultimately choose $\ell \ll \log^3 n$, so the second term is effectively negligible.
The following lemmas provide the bounds on the edge weights needed to bound the first term in \pref{eq:bound}:

\begin{lemma}[Bounds on contracted edge weights]\label{lem:weight-bds}
There exists a constant $c > 0$ so that if $d \gg \log^{16}n$, with high probability over the random variables $\{\tau^{i,j}\}_{i,j \in [n]}$, for any $e \in E(\bK)$ and $C \in \calC$,
\begin{align*}
\|\tilde{Q}_e \Ind(\cG_e)\|_{\infty} \le \log^5(n) \cdot p \cdot (c p \tau^2)^{t(e)-1},
\qquad \text{and} \qquad |\tilde{Q}_C| \le \log^2(n) \cdot p \cdot (c p\tau^2)^{t(C) - 2},
\end{align*}
where $t(e) \ge 2$ is the number of edges in the path that produced edge $e$ before contraction, and $t(C)\ge 2$ is the number of edges in the cycle $C$.
Further, if $p = \Omega(1/n)$, then for any uncontracted edge $e \in E(\bU)$,
\begin{align*}
|Q_e\Ind(\calG_e)| \le \Ind[e \in E(G)] + cp\log^3 n,
\end{align*}
and $\E[\Ind[e \in E(G)]] \le cp$.
\end{lemma}
We will prove \pref{lem:weight-bds} below, in \pref{ss:proofoflemma}.
\begin{lemma}[Bound on uncontracted edges]\label{lem:uncontracted-bds}
There exists a constant $C>0$ such that with high probability over the random variables $\{\tau^{i,j}\}_{i,j\in[n]}$,
\begin{align*}
\E\left(\prod_{e \in E(\bU)} \left|Q_e \Ind(\calG_e)\right|\right)
\le (C\log^3 n)^{|E(\bU)|} \cdot p^{|V(\tilde\bH)|-|E(\bK)| -1}.
\end{align*}
\end{lemma}
\begin{proof}
Choose a spanning forest $F$ of $\bU$.
Then
\begin{align*}
\E\left(\prod_{e \in E(\bU)} \left|Q_e \Ind(\calG_e)\right|\right)
\le \prod_{e \in E(\bU) \setminus E(F)} \|Q_e\Ind(\calG_e)\|_\infty \cdot \E \prod_{e \in E(F)} |Q_e\Ind(\calG_e)|.
\end{align*}
Now, by \pref{lem:weight-bds} there exists a constant $c > 0$ so that for any $e$, with high probability over the collection $\{\tau_{i,j}\}_{i,j\in[n]}$,
\begin{align*}
\E \prod_{e \in E(F)} |Q_e\Ind(\calG_e)| \le \E\prod_{e \in E(F)} (\Ind[e \in E(G)] + cp \log^{3} n).
\end{align*}
Further, for any leaf $(i,j)$ in the spanning forest $F$, $\Ind[(i,j) \in E(G)]$ is independent of the remaining edge indicators, and has expectation at most $cp$ with high probability over the $\tau^{i,j}$ (by \pref{lem:weight-bds}).
Applying this bound inductively, peeling off the leaves one at a time, we have
\begin{align*}
\E \prod_{e \in E(F)} |Q_e\Ind(\calG_e)| \le \E\prod_{e \in E(F)} (\Ind[e \in E(G)] + cp \log^{3} n) \le (2cp \log^{3} n)^{|E(F)|}.
\end{align*}
Since $F$ is a spanning forest of $\bU$, we note that $|E(F)| + |E(\bK)| \ge |V(\tilde\bH)| -1 $: the union of edges of $F$ and $\bK$ form a connected graph with $|V(\tilde\bH)|$ number of vertices. So, $|E(F)| \ge |V(\tilde{\bH})| - |E(\bK)| - 1$.

Combining with \pref{lem:weight-bds} to bound $\|Q_e\Ind(\calG_e)\|_\infty \le (1+2c)\log^3 n$ (and taking $C = 1+2c$) gives our conclusion.
\end{proof}

Applying \pref{lem:weight-bds} and \pref{lem:uncontracted-bds} in combination with \pref{eq:tilda-trans} and \pref{eq:bound}, we conclude
\begin{align}
&\E\left(\prod_{e \in E(\bH)} Q_e\right) - n^2 \cdot n^{-\log^3 n/3}(2+\sqrt{\td})^\ell\nonumber\\
&\le \prod_{C \in \calC} \log^2(n) \cdot p (c p  \tau^2)^{t(C)-2} \prod_{e \in E(\bK)} \log^5(n) \cdot p(cp\tau^2)^{t(e)-1} \cdot (c\log^3 n)^{|E(\bU)|} p^{|V(\tilde\bH)| - |E(\bK)|-1}\nonumber\\
&\le (\log^2 n)^{|\calC|}\cdot (\log^5 n)^{|E(\bK)|} \cdot (c\log^3 n)^{|E(\bU)|} (cp\tau^2)^{\sum_{e \in E(\bK)} t(e) + \sum_{C \in \calC} t(C) - 2|\calC|-|E(\bK)|} \cdot p^{|\calC|+|V(\tilde\bH)| -1}\nonumber\\
&\le (\log^5 n)^{|E(\tilde\bH)|+|\calC|/2} (cp\tau^2)^{\ell-2|\calC|-|E(\tilde\bH)|} \cdot p^{|\calC|+|V(\tilde\bH)| -1}
\label{eq:pbd}
\end{align}
where in the final line we use that $\ell = \sum_{C \in \calC} t(C) + \sum_{e \in E(\bK)} t(e) + |E(\bU)|$, and $|E(\tilde\bH)| = |E(\bU)| + |E(\bK)|$.

Now, we account for the number of distinct vertices in $\bH_{\vec{i}}$.
\begin{claim}[Bound on size of vertex set]\label{claim:vtx}
We can bound the size of the vertex set of $\bH_{\vec{i}}$ by
\begin{align*}
|V(\bH_{\vec{i}})| \le \ell - |\calC| - |E(\tilde\bH)| + |V(\tilde\bH)|+1.
\end{align*}
\end{claim}
\begin{proof}
We charge each contracted vertex to the cycle or edge in which it was contracted during the creation of $\tilde\bH_{\vec{i}}$ from $\bH_{\vec{i}}$.
In particular, when a cycle $C$ is contracted, every vertex save for the final vertex is removed, for a total of $t(C) -1$.
When a path of $t$ edges is contracted down to a single edge, $t-1$ vertices are removed.
This amounts to a total of $\sum_{C\in \calC} (t(C)-1) + \sum_{e \in E(\tilde\bH)} (t(e)-1) = \ell - |\calC| - |E(\tilde\bH)|$ vertices removed.
Finally, we account for the vertices remaining in $\tilde{\bH}$, plus the final vertex when $\tilde\bH=\varnothing$.
\end{proof}

We now have all of the ingredients with which to bound the trace.
We partition the sum over weighted closed walks of length $\ell$ in $[n]$ according to the shape of the corresponding graph $\bH$.
For each $\bH$ resulting from a closed walk of length $\ell$, let $\num(\bH)$ be the number of sequences $\vec{i} = i_1,\ldots,i_{\ell}$ which yield the graph $\bH$.
For each $s,m,v \in \N$, let $\calH_{s,m,v}$ be the set of all possible unlabeled graphs $\bH$ resulting from a closed walk of length $\ell$ for which $s = |\calC|$ cycles are removed in the process of producing $\tilde\bH$, and in which $|E(\tilde\bH)| = m$ and $|V(\tilde\bH)| = v$.
Note that $s \le \ell/2$ always, since each contracted cycle uses at least two edges, and similarly $v \le m/2$, because every vertex left over in $\tilde\bH$ cannot be contracted and therefore has degree at least $4$.
Then
\begin{align*}
&\E(\tr(Q^\ell))\\
&= \sum_{\vec{i} \in [n]^\ell} \E\left(\prod_{t=1}^\ell Q_{i_t,i_{t+1}}\right)\\
&= \sum_{s=0}^{\ell/2}\sum_{m=0}^{\ell-2s}\sum_{v=0}^{m/2}\sum_{\bH \in \calH_{s,m,v}} \num(\bH) \cdot \E\left(\prod_{e \in E(\bH)} Q_e \right)\intertext{
From \pref{claim:vtx}, if $\bH \in \calH_{s,m,v}$, $\num(\bH) \le n^{\ell-s-m+v+1}$ since we sample vertex labels from the set $[n]$ without replacement.
In combination with \pref{eq:pbd} this gives
}
&\le\sum_{s=0}^{\ell/2}\sum_{m=0}^{\ell-2s}\sum_{v=0}^{m/2}\sum_{\bH \in \calH_{s,m,v}} n^{\ell - s - m + v+1}
\left(\left(c\log^5 n\right)^{m+s/2} (cp\tau^2)^{\ell-2s-m} p^{s + v - 1}+ \frac{n^2(2+\sqrt{\td})^\ell}{n^{\log^3 n/3}}\right)\\
&\le\sum_{s=0}^{\ell/2}\sum_{m=0}^{\ell-2s}\sum_{v=0}^{m/2}\left|\calH_{s,m,v}\right|\cdot \left( n^2 \cdot \left(c\log^5 n\right)^{m+s/2} (cnp\tau^2)^{\ell-2s-m} (np)^{s + v - 1} + \frac{n^3(2+\sqrt{\td})^\ell n^\ell}{n^{\log^3 n/3}}\right)
\intertext{
Since $|\calH_{s,m,v}|$ is upper bounded by the number of $\ell$-vertex graphs with $\ell$ edges, which is at most $\ell^{2\ell}$,
}
&\le \ell^{2\ell} \sum_{s=0}^{\ell/2}\sum_{m=0}^{\ell-2s}\sum_{v=0}^{m/2} \left( n^2 \cdot \left(c\log^5 n\right)^{m+s/2} (cnp\tau^2)^{\ell-2s-m} (np)^{s + v - 1} + \frac{n^3(2+\sqrt{\td})^\ell n^\ell}{n^{\log^3 n/3}}\right)\\
&\le \ell^{2\ell} \sum_{s=0}^{\ell/2}\sum_{m=0}^{\ell-2s}\ell \cdot \left( n^2 \cdot \left(c\log^5 n\right)^{m+s/2} (cnp\tau^2)^{\ell-2s-m} (np)^{s + m/2 - 1} + \frac{n^3(2+\sqrt{\td})^\ell n^\ell}{n^{\log^3 n/3}}\right)\\
&\le \ell^{2\ell+1} \sum_{s=0}^{\ell/2}\sum_{m=0}^{\ell-2s}\left( n^2 \cdot \left(\log^5 n\frac{\sqrt{np}}{np\tau^2}\right)^{m+2s} (cnp\tau^2)^{\ell} + \frac{n^3(2+\sqrt{\td})^\ell n^\ell}{n^{\log^3 n/3}}\right)\\
\intertext{
The maximum term in the summation is achieved either at $m+2s = 0$ or $m+2s = \ell$, and there are at most $\ell^2$ terms, so
}
&\le \ell^{2\ell+3} \cdot \left(n^2 \cdot \max\left((c\log^5 n \sqrt{np})^\ell, (cnp\tau^2)^\ell\right) +  \frac{n^3(2+\sqrt{\td})^\ell n^\ell}{n^{\log^3 n/3}}\right).
\end{align*}

Choosing $\eps=1/\log n$ and the even integer $\ell=2\lceil \log^{3/2}n/2\rceil$, the second term is dwarfed by the first, and applying the trace method, we conclude that with high probability over the $\{\tau^{i,j}\}_{i,j \in [n]}$,
\begin{align*}
\Pr_{v_1,\ldots,v_n}\left(\|Q\| > 2\log^8 n \cdot \max\left(c\sqrt{np}, cnp\tau^2 \right)\right) \le \exp(-\ell/\log n)=o(1).
\end{align*}
Since $\log^8 n=o(\log^9 n)$, this is as desired.

\subsection{Controlling expansion coefficients and contracted edge weights}\label{ss:proofoflemma}
Above, we relied on high-probability upper bounds on the contribution that each edge could make to the weight of a walk in order to bound the contribution of walks containing vertices of degree $>2$.
The main purpose of this section is to prove those bounds. We start by establishing a lemma that bounds the size of $\tau$.

\begin{lemma}\label{lem:taubound1}
    For \muall, $\log^{16}n\ll d<n$, and $p \in [0,1/2-\varepsilon]$, there are constants $c_1>0$ and $C_2>0$ such that for all $n$ sufficiently large,
    \begin{align*}
        c_1 \sqrt{\frac{\log(1/p)}{d}} \leq \tau \leq C_2 \sqrt{\frac{\log(1/p)}{d}}.
    \end{align*}
\end{lemma}
\begin{proof}
Write
\begin{align*}
 u_i=(\mu S_i+N_i,w_i),\qquad g_i:=(N_i,w_i)\sim\calN(0,d^{-1}\Id_d),
\end{align*}
where $S_i$ is a Rademacher variable independent of $g_i$, and put
\begin{align*}
 X:=\langle g_i,g_j\rangle,\qquad Y:=\langle u_i,u_j\rangle.
\end{align*}
Thus
\begin{align*}
 Y-X=\mu(S_iN_j+S_jN_i)+\mu^2S_iS_j.
\end{align*}
We first record a uniform comparison with a Gaussian tail. Fix constants $0<c<C<\infty$. Uniformly for
\begin{equation}\label{eq:tau-comparison-range}
 \frac{c}{\sqrt d}\leq t\leq C\sqrt{\frac{\log n}{d}},
\end{equation}
we claim that
\begin{equation}\label{eq:tau-gaussian-comparison}
 \Pr(Y\geq t)=(1+o(1))\Psi(t\sqrt d),
\end{equation}
where the error is uniform in $t$.

To prove the claim, choose a sufficiently large constant $K$. Gaussian concentration gives an event $\mathcal E$, of probability at least $1-n^{-K}$, on which
\begin{align*}
 |N_i|+|N_j|\leq C_K\sqrt{\frac{\log n}{d}}
 \quad\text{and}
 \big|\|g_i\|-1\big|\leq C_K\sqrt{\frac{\log n}{d}}.
\end{align*}
On this event,
\begin{align*}
 |Y-X|\leq\delta_n:=\mu^2+C_K\mu\sqrt{\frac{\log n}{d}},
 \qquad
 \delta_n\sqrt d=O\bigp{\tfrac1{\log n}+d^{-1/4}}.
\end{align*}
Conditional on $g_i$, the random variable $X$ is Gaussian with variance $\|g_i\|^2/d$, and therefore
\begin{align*}
 \Pr(X\geq t\mid g_i)=\Psi\left(\frac{t\sqrt d}{\|g_i\|}\right).
\end{align*}
For $t$ in \pref{eq:tau-comparison-range}, the logarithm of the Gaussian tail changes by $o(1)$ under either the norm perturbation above or a shift of size $\delta_n\sqrt d$ in its argument. Using the elementary bounds
\begin{align*}
 \Pr(X\geq t+\delta_n)-n^{-K}
 \leq \Pr(Y\geq t)
 \leq \Pr(X\geq t-\delta_n)+n^{-K}
\end{align*}
and taking $K$ large enough proves \pref{eq:tau-gaussian-comparison}.

It remains to verify that $\tau$ lies in the comparison range. The law of $Y$ is symmetric and continuous. Moreover, conditioning on $u_i$ shows that its density in a neighborhood of zero is $O(\sqrt d)$ on the event $\|u_i\|\geq1/2$, whose complement has exponentially small probability. Hence $p\leq1/2-\varepsilon$ implies
\begin{align*}
 \tau\geq c_\varepsilon d^{-1/2}
\end{align*}
for some $c_\varepsilon>0$. For the upper bound, the same conditional-Gaussian representation gives
\begin{align*}
 \Pr(Y\geq t)\leq C_1e^{-c_1dt^2}+n^{-K}
\end{align*}
throughout the range $t=O(\sqrt{\log n/d})$. Since the standing assumption $pn\gg1$ gives $p\gg n^{-1}$, choosing a sufficiently large constant $C$ rules out $\tau>C\sqrt{\log n/d}$. We may therefore apply \pref{eq:tau-gaussian-comparison} at $t=\tau$ and obtain
\begin{align*}
 p=(1+o(1))\Psi(\tau\sqrt d).
\end{align*}
Finally, $\tau\sqrt d\geq c_\varepsilon$, so the standard Mills-ratio bounds imply
\begin{align*}
 \log(1/p)=\Theta_\varepsilon(1+d\tau^2).
\end{align*}
Because $\log(1/p)$ is bounded below by a positive constant, this is equivalent to the asserted two-sided bound on $\tau$.
\end{proof}

We next present a lemma that bounds the size of $\tau^{i,j}$ and $\lambda_k^{i,j}$.
\begin{lemma}\label{lem:tau}
For any \muall with $\log^{16}n\ll d<n$, there exists a constant $C_\varepsilon$ such that with high probability over the collection $\{\tau^{i,j}\}_{i,j\in[n]}$, the following holds uniformly,
\begin{align*}
    &\tau^{i,j} \leq C_\varepsilon \sqrt{\frac{\log(1/p)}{d}},\\
    &\lambda_0^{i,j} =  \Pr_{\xi \sim \cD_{\td}}(\xi \geq \sqrt{\td}\tau^{i,j}) \leq C_\varepsilon p,\\
    &\lambda_1^{i,j} = \frac{1}{\sqrt{\td}}\E_{\xi \sim \cD_{\td}}[\xi \Ind(\xi\geq \sqrt{\td}\tau^{i,j})]  \leq C_\varepsilon p_0^{i,j} \tau^{i,j} \leq C_\varepsilon^2 p\tau, \\
    &\lambda_2^{i,j} = \frac{1}{\td-1}\E_{\xi \sim \cD_{\td}}[(\xi^2-1) \Ind(\xi\geq \sqrt{\td}\tau^{i,j})] \leq C_\varepsilon p_0^{i,j} (\tau^{i,j})^2 \leq C_\varepsilon^2 p\tau^2.
\end{align*}
\end{lemma}
\begin{proof}
The proof of \pref{lem:taubound1} gives the sharper uniform relation
\begin{equation}\label{eq:p-gaussian-tail}
 p=(1+o(1))\Psi(\tau\sqrt d)
 =\Theta_\varepsilon\left(\frac{e^{-d\tau^2/2}}{1+\tau\sqrt d}\right).
\end{equation}
We now apply the same spherical-tail comparison to $\tau^{i,j}$ and the associated Gegenbauer coefficients.
Recall that we have defined $u_i = (a_i, \ell_i v_i)$, $\tau^{i,j} = (\tau - a_ia_j)/\ell_i\ell_j$ and $\lambda_k^{i,j} = \frac{1}{\sqrt{N_k}} \E_{\xi \sim \cD_{\td}}[ q_k^{(\td)}(\xi)\Ind(\xi \ge \sqrt{\td} \tau^{i,j})]]$.
Note that as \muall, we have $a_ia_j  = O(1/(\sqrt{d}\log n))$ and $\ell_i = 1+o(\log n/ \sqrt{d})$ with high probability. So
\begin{align*}
    \tau^{i,j} = \frac{\tau-a_ia_j}{\ell_i\ell_j} = \tau + O(1/(\sqrt{d}\log n)).
\end{align*}
Therefore we have that
\begin{align*}
    &p_0^{i,j} = \Pr_{\xi \sim \cD_{\td}}(\xi \geq \sqrt{\td}\tau^{i,j}) = \frac{\Gamma(\td/2)}{\sqrt{\td \pi} \Gamma((\td-1)/2)} \int_{\sqrt{\td}\tau^{i,j}}^{\sqrt{\td}} (1-\xi^2/\td) ^{\frac{\td-3}{2}} d\xi  \\
    &= \Theta(\Psi(\tau^{i,j}\sqrt{\td}))  =  \Theta(\Psi(\tau\sqrt{d})) = \Theta(p).
\end{align*}
This implies that $\tau^{i,j} = \Theta(\sqrt{\log(1/p_0^{i,j})}/\sqrt{d})$ with high probability. Similarly we have that $p_0= \Theta(\Psi(\tau\sqrt{d})) = \Theta(p)$. Similarly, for $\lambda_1^{i,j}$, we have an explicit formula
\begin{align*}
    &\lambda_1^{i,j} = \frac{1}{\sqrt{\td}}\E_{\xi \sim \cD_{\td}}[\xi \Ind(\xi\geq \sqrt{\td}\tau^{i,j})]  = \frac{\Gamma(\td/2)}{\td \sqrt{ \pi} \Gamma((\td-1)/2)} \int_{\sqrt{\td}\tau^{i,j}}^{\sqrt{\td}}\xi (1-\xi^2/\td) ^{\frac{\td-3}{2}} d\xi\\
    & = -\frac{\Gamma(\td/2)}{\td \sqrt{ \pi} \Gamma((\td-1)/2)} \frac{\td}{\td-1}(1-\xi^2/\td) ^{\frac{\td-1}{2}} \mid_{\sqrt{\td}\tau^{i,j}}^{\sqrt{\td}} = \Theta(p_0^{i,j} \tau^{i,j}) = \Theta(p\tau).
\end{align*}
Similarly for $\lambda_2^{i,j}$, we also have an explicit formula
\begin{align*}
    \lambda_2^{i,j} &= \frac{1}{\td-1}\E_{\xi \sim \cD_{\td}}[(\xi^2-1) \Ind(\xi\geq \sqrt{\td}\tau^{i,j})] \\
    &= \frac{\Gamma(\td/2)}{(\td-1) \sqrt{ \td\pi} \Gamma((\td-1)/2)} \int_{\sqrt{\td}\tau^{i,j}}^{\sqrt{\td}}(\xi^2-1) (1-\xi^2/\td) ^{\frac{\td-3}{2}} d\xi\\
    & = - \frac{\Gamma(\td/2)}{(\td-1) \sqrt{ \td\pi} \Gamma((\td-1)/2)}  \xi (1-\xi^2/\td) ^{\frac{\td-1}{2}} \mid_{\sqrt{\td}\tau^{i,j}}^{\sqrt{\td}}  = \Theta(p_0^{i,j} (\tau^{i,j})^2) = \Theta(p\tau^2).\qedhere
\end{align*}
\end{proof}

To prove \pref{lem:weight-bds}, we need an $L_\infty$ bound of the normalized Gegenbauer polynomials $q_k$.
\begin{lemma}\label{lem:Gegencoeff}
There exists a fixed constant $d_0$ such that for any $\td > d_0$, $k \geq 0$, and any $B \ge 3$,
    \begin{align*}
        \sup_{x\in [-B, B]} |q_k(x)| \leq B^k.
    \end{align*}
\end{lemma}
\begin{proof}
    The Gegenbauer polynomials can be defined using the following recurrence \cite{dai2013approximation,efthimiou2014spherical}:
    \begin{align*}
        x\cdot q_k(x) = a_k \cdot q_{k+1}(x) + a_{k-1}\cdot q_{k-1}(x),
    \quad \text{where}\quad
        a_k = \sqrt{\tfrac{(k+1)(\td+k-2)\td}{(\td+2k)(\td+2k-2)}}.
    \end{align*}
    We will prove that the supremum bound inductively.
One can manually verify that the lemma holds for $k = 0, 1,2$, noting that
    \begin{align*}
    q_{0}(x) = 1, &\quad q_{1}(x) = x, \quad q_{2}(x) = \frac{1}{\sqrt{2}}\sqrt{\tfrac{\td+2}{\td-1}}(x^2 - 1).
    \end{align*}
Now, for the inductive step, the recurrence and the inductive hypothesis imply that
\begin{align*}
\sup_{|x|\le B} |q_{k+1}(x)|
&\le  \frac{a_{k-1}}{a_k}\sup_{|x|\le B}|q_{k-1}(x)| + \frac{1}{a_k}\sup_{|x|\le B} |x|\cdot|q_k(x)|
\le \frac{a_{k-1}}{a_k} B^{k-1} + \frac{1}{a_k} B^{k+1}.
\end{align*}
    We'll show that so long as $d > d_0$, $k \ge 2$, and $B \ge 3$, $\frac{a_{k-1}}{a_k B^2} + \frac{1}{a_k} \le 1$, which completes the proof of the lemma.
We note that when $d,k \ge 2$,
    \begin{align*}
 &\sqrt{k+1} \ge  a_k \geq \min \left \{\sqrt{\tfrac{1}{8}\td},\sqrt{\tfrac{1}{2}(k+1)} \right \},\\
        &\frac{a_{k-1}}{a_k} = \sqrt{\frac{k(\td+k-3)(\td+2k)}{(k+1)(\td+k-2)(\td+2k-4)}}\leq \sqrt{\frac{\td+2k}{\td+2k-4}}.
    \end{align*}
    Therefore, for any $k\geq 2$ and $\td$ larger than some constant $d_0$, we have that
    \begin{align*}
        \frac{1}{a_k}+\frac{a_{k-1}}{a_k B^2}
\le \max\left(\sqrt{\tfrac{8}{\td}},\sqrt{\tfrac{2}{k+1}}\right) + \frac{1}{9}\sqrt{1 + \tfrac{4}{\td + 2k - 4}}
\leq 1,
    \end{align*}
as desired.
\end{proof}
Now we provide another bound which will come in useful when $k$ is large relative to $d$.
\begin{claim}\label{claim:largekGegensup}
    There exists a fixed constant $d_0$ such that for any $\td > d_0$, $k \geq d^{2/3}$, and any $0\leq B <\sqrt{\td}/2$,
    \begin{align*}
        \sup_{x\in [-B, B]} |q_k(x)| \leq \sqrt{N_k} \sqrt{3} \td^{1/4} \exp(B^2/2) \cdot \frac{k \Gamma (\tfrac{\td-1}{2}) \Gamma(k)}{\Gamma (\tfrac{\td-1}{2}+ k)}.
    \end{align*}
\end{claim}
\begin{proof}
    This is a consequence of the connection between Gegenbauer and Jacobi polynomials, combined with known bounds on Jacobi polynomials.
    The Jacobi polynomials $P_k^{(\alpha, \beta)}(x)$ are defined by
    \begin{align*}
        P_k^{(\alpha, \beta )} (x) = \frac{(-1)^k}{2^k k!} (1-x)^{-\alpha} (1+x)^{-\beta} \left( \frac{d}{dx }\right)^k \left( (1-x)^{\alpha+k} (1+x)^{\beta+k}\right).
    \end{align*}
    The Gegenbauer polynomial $q^{(\td)}_k$ is proportional to the Jacobi polynomial with $\alpha = \beta = (\td -3)/2$,
    \begin{align}
        q_k(x) & = \frac{k+\tfrac{\td}{2}-1}{\tfrac{\td}{2}-1} \frac{1}{\sqrt{N_k}} \frac{\Gamma(\tfrac{\td}{2}-\tfrac{1}{2}) \Gamma (k+\td-2)}{\Gamma(\td-2)\Gamma(k+\tfrac{\td}{2}-\tfrac{1}{2})} P_k^{((\td-3)/2, (\td-3)/2)}\left( \frac{x}{\sqrt{\td}} \right)\nonumber \\
        & = \sqrt{N_k}\frac{k \Gamma (\tfrac{\td-1}{2}) \Gamma(k)}{\Gamma (\tfrac{\td-1}{2}+ k)} P_k^{((\td-3)/2, (\td-3)/2)}\left( \frac{x}{\sqrt{\td}} \right),\label{eq:GegenJacobi}
    \end{align}
    for a reference, see \cite{dai2013approximation}, equation (B.2.1) (noting that we normalize our Gegenbauer polynomial differently).
    By Theorem 2 in \cite{krasikov2007upper}, we have that when $k\geq 6$, and $\alpha, \beta \geq \tfrac{1+\sqrt{2}}{4}$, for any $x \in [-1,1]$,
    \begin{align*}
        (1-x)^{\alpha+\tfrac{1}{2}} (1+x)^{\beta+\tfrac{1}{2}}\left ( P_k^{(\alpha, \beta)} (x) \right) ^2 < 3 \alpha^{1/3} (1+\frac{\alpha}{k})^{1/6}.
    \end{align*}
    This implies that for any $0\leq b<1$,
    \begin{align*}
        \sup_{x\in [-b, b]} \left | P_k^{((\td-3)/2, (\td-3)/2)} (x) \right| < \sqrt{ \frac{3 (\tfrac{\td-3}{2})^{1/3} (1+\frac{\td-3}{2k})^{1/6}} { (1-b)^{\tfrac{\td}{2}-1} (1+b)^{\tfrac{\td}{2}-1} } }.
    \end{align*}
    Combining with \pref{eq:GegenJacobi}, we have
    \begin{align*}
        \sup_{x\in [-B, B]}  |q_k(x)|
        &< \sqrt{N_k}\frac{k \Gamma (\tfrac{\td-1}{2}) \Gamma(k)}{\Gamma (\tfrac{\td-1}{2}+ k)} \sqrt{ \frac{3 (\tfrac{\td-3}{2})^{1/3} (1+\frac{\td-3}{2k})^{1/6}} { \left(1-\tfrac{B}{\sqrt{\td}}\right)^{\td/2-1} \left(1+\tfrac{B}{\sqrt{\td}}\right)^{\td/2-1} } }.
    \end{align*}
    Now for any $k\geq d^{2/3}$, $d$ sufficiently large, and any $0\leq B <\sqrt{\td}/2$, we have that
    \begin{align*}
        \sqrt{ \frac{3 (\tfrac{\td-3}{2})^{1/3} (1+\frac{\td-3}{2k})^{1/6}} { \left(1-\tfrac{B}{\sqrt{\td}}\right)^{\td/2-1} \left(1+\tfrac{B}{\sqrt{\td}}\right)^{\td/2-1} } }
        \leq \sqrt{ \frac{3 (\td)^{1/3} (d^{1/3})^{1/6}} { \left(1-\tfrac{B^2}{\td}\right)^{{\td/2} -1}}}
        &\leq \sqrt{3} \td^{1/4} \left(1+\frac{2B^2}{\td}\right)^{\td/4} \\
        &\leq \sqrt{3} \td^{1/4} \exp(B^2/2),
    \end{align*}
    which completes the proof.
\end{proof}

In the proof of \pref{lem:weight-bds} we also need to control the decay of $\lambda_k$.
Intuitively, when $k$ is small, $\lambda_k$ should be $O(p\tau^k) = O(\polylog \frac{1}{p}\cdot p d^{-k/2})$. The two lemmas below give a very coarse bound on the decay of $\lambda_k$ for any $k$.
\begin{lemma}\label{lem:lambdabound}
	Suppose $p = \Omega(\frac{1}{n})$.
	Then for any $k\geq 3$, $2\leq t\leq \log^2 n$ and any $i,j$, we have that
    \begin{align*}
        \sup_{x\in [-\log^2 n, \log^2 n]} |q_k(x)|\sqrt{N_k} |\lambda_k^{i,j}|^{t} \leq \log^4 n \sqrt{N_2}(C_\varepsilon p_0\tau^2)^t /k^2
    \end{align*}
    uniformly, with high probability over the collection $\{\tau^{i,j}\}_{i,j\in[n]}$.
\end{lemma}
\begin{proof}
We divide our proof into two cases: relatively small $k$ and large $k$. For $k\leq \log n$, we use a more refined estimate. We note that by Rodrigues' formula for Gegenbauer polynomials \cite{dai2013approximation}, we have that
\begin{align*}
    q_k^{(\td)}(\xi)\cdot (1-\frac{\xi^2}{\td})^{(\td-3)/2} = C_{k,\td} \bigp{ \frac{d}{d \xi} }^k \bigp{1-\frac{\xi^2}{\td}}^{k+(\td-3)/2},
\end{align*}
where
\begin{align*}
    C_{k,\td} = \sqrt{N^{(\td)}_k} \bigp{-\frac{1}{2}}^k \frac{\Gamma(\frac{\td-1}{2})}{\Gamma(k+\frac{\td-1}{2})} \sqrt{\td}^k.
\end{align*}
A similar formula holds for $k-1$ and $\td+2$, i.e.,
\begin{align*}
    q^{(\td+2)}_{k-1}(\xi)\cdot (1-\frac{\xi^2}{\td+2})^{(\td-1)/2} = C_{k-1,\td+2} \bigp{ \frac{d}{d \xi} }^{k-1} \bigp{1-\frac{\xi^2}{\td+2}}^{k-1+(\td-1)/2}.
\end{align*}
Applying a change of variables, the anti-derivative of $q^{(\td)}_k(\xi) \cdot (1-\xi^2/\td)^{(\td-3)/2}$ equals
\begin{align*}
    f_{k,\td}(\xi):= q_{k-1}^{(\td+2)} \bigp{ \frac{\sqrt{\td+2}}{\sqrt{\td}} \xi } (1-\xi^2/\td)^{(\td-1)/2} \bigp{ \frac{\sqrt{\td+2}}{\sqrt{\td}} }^{k-1} \frac{C_{k,\td}}{C_{k-1,\td+2}}.
\end{align*}
By \pref{lem:Gegencoeff}, we have that
\begin{align*}
    |f_{k,\td}(\xi)|
    &\leq \bigp{ \frac{\sqrt{\td+2}}{\sqrt{\td}} |\xi| }^{k-1} \left(1-\tfrac{\xi^2}{\td}\right)^{(\td-1)/2} \left(\frac{\td +2}{\td}\right)^{(k-1)/2}\left|\frac{C_{k,\td}}{C_{k-1,\td+2}}\right|\\
    &= \left( \frac{\td+2}{\td}\right)^{(k-1)/2} |\xi|^{k-1}\left(1-\tfrac{\xi^2}{\td}\right)^{(\td-1)/2} \frac{\td}{\sqrt{k(\td+k-2)(\td-1)}}\\
    &\leq \left|\xi \sqrt{\tfrac{\td+2}{\td}} \right|^{k-1} \exp\left(-\xi^2 \tfrac{\td-1}{2\td}\right),
\end{align*}
Note that since the probability distribution of $\xi \sim \cD_{\td}$ equals $\frac{\Gamma(\td/2)}{\sqrt{\td\pi} \Gamma ((\td-1)/2)} (1-\xi^2/\td)^{(\td-3)/2}$ and $\frac{\Gamma(d/2)}{\sqrt{d\pi} \Gamma ((d-1)/2)} \to 1/\sqrt{2\pi}$, this implies that
    \begin{align*}
        |\lambda_k^{i,j}| &= \left|\frac{1}{\sqrt{N_k}} \E_{\xi \sim \cD_{\td}}[q_k(\xi) \Ind(\xi\geq \sqrt{\td}\tau^{i,j})] \right| \leq  \left|\frac{1}{\sqrt{N_k}} (f_{k,\td}(\sqrt{\td}\tau^{i,j}) - f_{k,\td}(\sqrt{\td}) )\right| \\
        &\leq  \frac{1}{\sqrt{N_k}}|\sqrt{\td+2}\cdot \tau^{i,j}|^{k-1} \exp\left(-(\tau^{i,j})^2 \tfrac{\td-1}{2} \right).
    \end{align*}
By the proof of \pref{lem:tau}, we have that $\sqrt{\td+2}\cdot \tau^{i,j} \le C_{\eps} \sqrt{\log \frac{1}{p}}$ for all $i,j$ with high probability over the collection $\{\tau^{i,j}\}_{i,j\in[n]}$, and that $\exp(-(\tau^{i,j})^2(\td-1)/2)=O (\tau^{i,j}\sqrt{\td} p_0^{i,j}) = O(\tau\sqrt{d} p)$. Therefore, we have that
\begin{equation}\label{eq:boundlambdaklem}
\begin{aligned}
 |\lambda_k^{i,j}|
 &\leq \frac{1}{\sqrt{N_k}}
   (C_0\sqrt{\td+2}\,\tau^{i,j})^{k-1}
   \tau^{i,j}\sqrt{\td}\,p_0^{i,j}\\
 &\leq \frac{p_0}{\sqrt{N_k}}(C\tau\sqrt d)^k
 \leq p_0\frac{\sqrt{k!}}{\sqrt{d^k}}(C\tau\sqrt d)^k
 \leq p_0(C\tau)^k\sqrt{k!}.
\end{aligned}
\end{equation}
For $k\leq \log n$, we have that there is an absolute constant $C$ such that
\begin{align*}
        \frac{\sqrt{N_k} |\lambda_k^{i,j}|^{t}} {\sqrt{N_2} (C_\varepsilon p_0\tau^2)^t }
        &\le \frac{d^{k/2}}{\sqrt{k!}d} \cdot \frac{(p_0 (C\tau)^k \sqrt{k!})^t}{(p_0 \tau^2)^t}
        \le (C^2 2k\sqrt{d}\tau^2)^{k-2} \cdot ((C\tau)^{k-2} \sqrt{k!})^{t-2}
    \end{align*}
    where we have used that $k! \le (2k)^{k-2}$ for $k \ge 3$.
By \pref{lem:Gegencoeff}, we know that $\sup_{x\in [-\log^2 n, \log^2 n]} |q_k(x)| \leq \log^{2k} n$. Therefore, we have
\begin{align*}
    \sup_{x\in [-\log^2 n, \log^2 n]}\frac{ |q_k(x)|}{\log^4 n}\frac{\sqrt{N_k} |\lambda_k^{i,j}|^{t} }{ \sqrt{N_2}(C p_0\tau^2)^t /k^2}
    &\leq \left( k^2 \left(C^2 2k\sqrt{d}\tau^2 \log^2 n\right)^{k-2}\right)\cdot ((C \tau)^{k-2}\sqrt{k!})^{t-2}.
\end{align*}
Since $d \gg \log^{12} n$, $3 \le k \le \log n$, and $\tau = O(\sqrt{\log n /d})$, the first term is $\ll 1$ and the latter term is at most $1$ for large enough $n$.
\medskip

For $k \ge \log n$, we use a more direct bound.
Since the $q_k$ form an orthonormal basis, this implies that
    \begin{equation}\label{eq:lambdaboundlem2}
        (\lambda_k^{i,j})^2 N_k \leq \left\|\Ind(\xi\geq \sqrt{\tilde d}\tau^{i,j})\right\|_2^2 = p_0^{i,j} \leq 1.
    \end{equation}
Therefore, we have that
\begin{align*}
    \abs{ \sqrt{N_k} (\lambda_k^{i,j})^{t}} &\leq  N_k^{-(t-1)/2},
\end{align*}
and so
\begin{align*}
\sup_{x \in [-\log^2 n,\log^2 n]} \frac{|q_k(x)|}{\log^4 n}\frac{\sqrt{N_k}|\lambda_k^{i,j}|^t}{\sqrt{N_2}(Cp_0\tau^2)^t /k^2}
&\le k^2 \cdot \sup_{|x| \le \log^2 n } |q_k(x)| \cdot \sqrt{\frac{N_k}{N_2}}\frac{1}{(\sqrt{N_k}Cp_0 \tau^2)^t}.
\end{align*}
We note that by definition
\begin{align*}
    N_k = \frac{2k+\td-2}{k}\binom{k+\td-3}{k-1} \geq \frac{\td^k}{k^k}.
\end{align*}
Thus, since the $N_k$ are non-decreasing,
\begin{equation}\label{eq:tsmallenough}
    \sqrt{N_k} p_0 \tau ^2 \geq \frac{\sqrt{N_k}}{nd} \geq \frac{\sqrt{N_{\lfloor \log n \rfloor}}}{nd} \geq \left(\frac{\td}{\lfloor \log n \rfloor }\right)^{\lfloor \log n \rfloor/2} \frac{1}{nd}\gg 1,
\end{equation}
since $d \gg \log^2 n$. So to prove our statement for large $k$, it remains to check the case when $t=2$, i.e., we need to show that
\begin{align*}
    \sup_{x\in [-\log^2 n, \log^2 n]}|q_k(x)| \ll \log^4 n \sqrt{N_k}\sqrt{N_2}(C p_0\tau^2)^2 / k^2
\end{align*}
Now, if $k \leq d^{2/3}$, by \pref{lem:Gegencoeff}, then our lower bound $d \gg \log^{12} n$ implies that
\begin{align*}
    \sup_{x\in [-\log^2 n, \log^2 n]} |q_k(x)| \leq (\log n)^{2k} \ll \frac{d^{k/2}}{k^{k/2} n^2 d^2 k^2 } \ll \log^4 n \sqrt{N_k}\sqrt{N_2}(C p_0\tau^2)^2 / k^2,
\end{align*}
where the second inequality follows by taking $\log$ on both sides and dividing by $k$. If $k \geq d^{2/3}$, by \pref{claim:largekGegensup},
\begin{equation}\label{eq:ratioqN_k}
    \frac{\sup_{x\in [-\log^2 n, \log^2 n]} |q_k(x)| }{\sqrt{N_k}}\leq \sqrt{3} \td^{1/4} \exp(\log^4 n/2) \cdot \frac{k \Gamma (\tfrac{\td-1}{2}) \Gamma(k)}{\Gamma (\tfrac{\td-1}{2}+ k)}.
\end{equation}
By Stirling's formula,
    \begin{align*}
        \frac{k \Gamma (\tfrac{\td-1}{2}) \Gamma(k)}{\Gamma (\tfrac{\td-1}{2}+ k)}
        &\sim
        \frac{\sqrt{2\pi (\tfrac{\td-1}{2} \cdot k^3) }}{\sqrt{\tfrac{\td-1}{2}+k}} \bigp{ \frac{\tfrac{\td-1}{2}}{\tfrac{\td-1}{2}+k} }^{\tfrac{\td-1}{2}} \bigp{ \frac{k}{\tfrac{\td-1}{2}+k} }^{k}.
    \end{align*}
Taking logarithm of the right hand side of \pref{eq:ratioqN_k}, we have that
\begin{align*}
    &\log \bigp{\frac{\sup_{x\in [-\log^2 n, \log^2 n]} |q_k(x)| }{\sqrt{N_k}}} \\
    &\leq O(1) +O(\log(\td)) + \log^4 n /2 + O(\log k) - \frac{\td-1}{2} \log\bigp{ \frac{\tfrac{\td-1}{2}+k}{\tfrac{\td-1}{2}} } - k \log\bigp{ \frac{\tfrac{\td-1}{2}+k}{k}}\\
    &< -2\log(n)-\log (\td)-2 \log(k),
\end{align*}
where the last inequality follows for all $d \ge \log^{12} n$ so long as $n$ is sufficiently large, for instance by noting that in the case $k > d^{100}$, the subtracted terms have magnitude at least $\frac{1}{3}d \log k$, whereas in the case $k \le d^{100}$ the subtracted terms have magnitude at least $k \log 2 \ge d^{2/3} \log 2$.

This implies that
\begin{align*}
    \sup_{x\in [-\log^2 n, \log^2 n]} |q_k(x)| \leq \frac{\sqrt{N_k}}{ n^2 d k^2 } \ll \log^4 n \sqrt{N_k}\sqrt{N_2}(C p_0\tau^2)^2 / k^2,
\end{align*}
which completes the proof.
\end{proof}
We will make use of a second bound for the $\lambda_k^{i,j}$:
\begin{lemma}\label{lem:lambdabound2}
	Suppose $p = \Omega(\frac{1}{n})$.
	Then for any $k\geq 3$, $3\leq t\leq \log^2 n$ and any $i,j$, we have that
    \begin{align*}
        \abs{ q_k(\sqrt{\td})\sqrt{N_k} (\lambda_k^{i,j})^{t}} \leq \td \sqrt{N_2}(C_\varepsilon p_0\tau^2)^t / k^2
    \end{align*}
    uniformly, with high probability over the collection $\{\tau^{i,j}\}_{i,j\in[n]}$.
\end{lemma}
\begin{proof}
    The proof follows the same strategy as \pref{lem:lambdabound}. When $k\leq \log n$, by \pref{eq:boundlambdaklem} and \pref{lem:tau}, we have that
    \begin{align*}
        \abs{ q_k(\sqrt{\td})\sqrt{N_k} (\lambda_k^{i,j})^{t}}
        &\leq d^{k/2} \sqrt{N_k} p_0^t (C\tau \sqrt{\log n})^{tk}
        \leq p_0^t d^{k} (C\log n)^{tk/2} \tau^{tk}
    \end{align*}
    Comparing to the desired upper bound,
    \begin{align*}
        \frac{p_0^t d^{k} (C\log n)^{tk/2} \tau^{tk}}{\td \sqrt{N_2} (C p_0 \tau^2)^t /k^2}
        &\le k^2\frac{(C' \log n)^{t(k-1)}}{d^{(t/2-1)(k-2)}}
        = k^2 \left(\frac{(C'\log n)^{\frac{t}{(t/2-1)}\frac{k-1}{k-2}}}{d}\right)^{(t/2-1)(k-2)}
    \end{align*}
    for $C'$ some constant, where we have used that $\tau^2 = \Theta(\log n / d)$.
    Given that $k \ge 3, t \ge 3$, and $d \gg \log^{16} n$, the right-hand side above is $o(1)$ and the conclusion holds.

    It remains to handle the case $k\geq \log n$.
    By the orthonormality of the spherical harmonics, $q_k(\sqrt{\td}) = \sqrt{N_k}$. So we have that by \pref{lem:Gegencoeff} and \pref{eq:lambdaboundlem2},
    \begin{align*}
        \abs{ q_k(\sqrt{\td}) \sqrt{N_k} (\lambda_k^{i,j})^{t}} \leq |\lambda_k^{i,j}|^{t-2} \leq N_k^{-t/2+1} \leq \td \sqrt{N_2}(C_\varepsilon p_0\tau^2)^t k^{-2} \frac{N_k^{-t/2+1}}{\td \sqrt{N_2}(C_\varepsilon p_0\tau^2)^t k^{-2}}.
    \end{align*}
    To bound the fraction, by \pref{eq:tsmallenough}, it is enough to check the case when $t=3$. Indeed, in such situation, the fraction is bounded by
    \begin{align*}
        \frac{ N_k^{-1/2} k^2 }{ d^2 (Cp_0 \tau^2)^3 } \leq \frac{N_{\lfloor \log n \rfloor}^{-1/2}  \log^2 n }{ d^2 (Cp_0 \tau^2)^3 } \leq \frac{(C\log n)^{\log n/2+2} n^3} {d^{\log n /2}} \ll 1.
    \end{align*}
\end{proof}
Combining Lemmas \ref{lem:tau}, \ref{lem:Gegencoeff}, and \ref{lem:lambdabound}, we now prove \pref{lem:weight-bds}.
\begin{proof}[Proof of \pref{lem:weight-bds}]
For any edge $e \in E(\tilde{\bG}_{\vec{i}})$ which is the result of the contraction of a path $s_1,\ldots,s_{t+1} \in \bG_{\vec{i}}$, recall that by definition, we have
\begin{align*}
    \|\tilde{Q}_e \Ind(\cG_e)\|_{\infty} & = \left \| \sum_{k=2}^\infty q_k(\sqrt{\td}\iprod{v_{s_1},v_{s_{t+1}}}) \cdot \Ind(\cG_{(s_1,s_{t+1})}) \cdot\sqrt{N_k}\cdot \prod_{a=1}^t \lambda_k^{s_a,s_{a+1}}  \right\|_{\infty}\\
    & = \left \| \sum_{k=2}^\infty q_k(\sqrt{\td}\iprod{v_{s_1},v_{s_{t+1}}}) \cdot \Ind\bigp{ |\iprod{v_{s_1},v_{s_{t+1}}}| < \frac{\log^2(n)}{\sqrt{\td}} } \cdot\sqrt{N_k}\cdot \prod_{a=1}^t \lambda_k^{s_a,s_{a+1}}  \right\|_{\infty}.
\end{align*}
By \pref{lem:Gegencoeff} (taking $B$ to be $\log^2(n)$) and \pref{lem:lambdabound}, we have that for $t\geq 2$,
\begin{align*}
    \|\tilde{Q}_e \Ind(\cG_e)\|_{\infty} &\leq \bigp{\log^2(n)}^2 \tilde d \prod_{a=1}^t \bigp{C_\varepsilon p_0^{s_a,s_{a+1}} (\tau^{s_a,s_{a+1}})^2 } + \sum_{k=3}^\infty \bigp{\log^2(n)}^2 \sqrt{N_2}(C_\varepsilon p_0\tau^2)^t k^{-2} \\
    & \leq \log^4(n) \td (C_{\varepsilon,3}'' p_0\tau^2)^t,
\end{align*}
    uniformly, with high probability over the collection $\{\tau^{i,j}\}_{i,j\in[n]}$.
    The bound as stated in \pref{lem:weight-bds} follows by applying \pref{lem:tau} to eliminate the factor $\tilde d$ at the cost of a factor $\log\frac{1}{p}$.
    When $t = 1$, then $\tilde{Q}_e = Q_e$. Write $e = (i,j)$, then we have
\begin{align*}
   &|Q_e \Ind(\cG_e)| \le \Ind(\iprod{v_i, v_j} \geq \tau^{i,j} ) +  p_0^{i,j} + |\td \lambda_1^{i,j} \inp{v_i,v_j} \Ind(\cG_e)| \\
    &\leq \Ind(\iprod{v_i, v_j} \geq \tau^{i,j}) + C_\varepsilon p_0 + C_\varepsilon \td p_0 \tau \frac{\log^2(n)}{\sqrt{\td}} \leq \Ind(\iprod{v_i, v_j} \geq \tau^{i,j})+ 2C_\varepsilon \log^3(n) p_0.
\end{align*}
where the second-to-last inequality follows from \pref{lem:tau}.
Furthermore, $\E[\Ind[\iprod{v_i,v_j} \geq \tau^{i,j}]] \le C_\eps p_0$.

Now for any cycle $C = (s_1,\ldots,s_{t+1} = s_1) \in \bG$ that was contracted into a self-loop and removed in producing $\tilde\bG$, recall that we have
\begin{align*}
\tilde{Q}_C = \sum_{k=2}^\infty q_k(\sqrt{\td}) \cdot \sqrt{N_k} \cdot \prod_{a=1}^t \lambda_k^{s_a,s_{a+1}}.
\end{align*}
For $t\geq 3$, by \pref{lem:lambdabound2}, we have that
\begin{align*}
    |\tilde{Q}_C| &\leq \td \sqrt{N_2} (C_{\varepsilon} p\tau^2)^{t(C)} +  \sum_{k=3}^\infty \td \sqrt{N_2}(C_\varepsilon p_0\tau^2)^{t(C)} k^{-2} \leq C\td \sqrt{N_2} (C_{\varepsilon} p\tau^2)^{t(C)} \\
    &\leq (p^2 \tau^4 \td \sqrt{N_2})\cdot (c_2 p\tau^2)^{t(C) - 2} \leq (\log^2 n) p \cdot (c_3 p\tau^2)^{t(C) - 2}.
\end{align*}
When $t = 2$, we write $e = (s_1,s_2)$, then $\tilde{Q}_C = \E(Q_e^2)$. Therefore, by \pref{lem:tau}, we have that
\begin{align*}
    \tilde{Q}_C &= \E\bigp{ \Ind(\iprod{v_i, v_j} \geq \tau^{i,j} ) -  p_0^{i,j} - \td \lambda_1^{i,j} \inp{v_i,v_j} }^2 \\
    &= p_0^{i,j}+(p_0^{i,j})^2+\td^2 (\lambda_1^{i,j})^2 \E(\iprod{v_i, v_j}^2)-2 (p_0^{i,j})^2 -2\td \lambda_1^{i,j} \E(\iprod{v_i, v_j} \Ind(\iprod{v_i, v_j} \geq \tau^{i,j} ) )\\
    & \leq C p + Cp^2 \log(1/p) \leq C_2 p.
\end{align*}
Therefore, we have
\begin{align*}
    \qquad |\tilde{Q}_C| \le \log^2(n) \cdot p \cdot (c p\tau^2)^{t(C) - 2},
\end{align*}
where $t(C)$ is the number of edges in the cycle $C$.
\end{proof}

\subsection{Relating the Gaussian mixture and the sphere}\label{sec:g-sph}
In this section, we will relate the matrix we subtracted from $A$ in \pref{sec:trace} to the matrix we wish to show is close to the top eigenspace of $A$, $p_0\topev\topev^\top + \tilde{d} \lambda_1 UU^\top$. Recall the definition of $\ell_k$, $v_k$, $a_k$ and $\tau^{i,j}$ in \pref{eq:v}.
First, we finally define $\topev$.
 For each $k \in [n]$, define $L_k:= \ell_k-1$, and define $\tilde{\mathbbm{1}}_n$ to be a length-$n$ vector where each entry equals
\begin{align*}
    (\tilde{\mathbbm{1}}_n)_k = 1+\frac{L_k \lambda_1 \td \tau}{p_0}.
\end{align*}
We prove the following lemma and another lemma (\pref{lem:opdiff2}) designed for larger $\mu$. Let $V$ be the $n$ by $\tilde d$ matrix where the $i$-th row equals $v_i$.
\begin{lemma}\label{lem:opdiff}
    For \musmall, let $D$ be the diagonal matrix of
    $p_0^{i,j}\mathbbm{1}_n \mathbbm{1}_n^\top + \td \lambda_1^{i,j} VV^\top - (p_0\tilde{\mathbbm{1}}_n \tilde{\mathbbm{1}}_n^\top + \td \lambda_1 UU^\top)$. Then, with high probability,
    \begin{align*}
            \| p_0^{i,j}\mathbbm{1}_n \mathbbm{1}_n^\top + \td \lambda_1^{i,j} VV^\top - (p_0\tilde{\mathbbm{1}}_n \tilde{\mathbbm{1}}_n^\top + \td \lambda_1 UU^\top) -D \|_{\mathrm{op}} \ll np \tau^2\log^4(n).
    \end{align*}
\end{lemma}
\begin{defn}
We define (and recall that)
\begin{align*}
    b_i:=a_i/\sqrt{\tau}, \quad L_i:= \ell_i-1, \quad
    p_0= \Pr_{\xi \in \cD_{\td}} (\xi \geq \sqrt{\td} \tau),
    \quad \lambda_1= \frac{1}{\sqrt{\tilde d}}\E_{\xi \in \cD_{\td}}[\xi \Ind(\xi\geq \sqrt{\tilde d}\tau)].
\end{align*}
We further define the change of $\tau^{i,j}$, $p_0^{i,j}$ and $\lambda_1^{i,j}$ as follows.
\begin{align*}
    \Delta \tau^{i,j} := \tau^{i,j}-\tau, \qquad \Delta p_0^{i,j} := p_0^{i,j} - p_0, \qquad \Delta \lambda_1^{i,j} := \lambda_1^{i,j} - \lambda_1.
\end{align*}
\end{defn}
To prove \pref{lem:opdiff}, we first prove a lemma bounding the fluctuations of $\tau^{i,j}$, $p_0^{i,j}$ and $\lambda_1^{i,j}$.
\begin{lemma}\label{lem:bound2}
    For $\mu \leq \tau$, the following holds uniformly with high probability
    \begin{align*}
        & L_i = o(\tfrac{\log n}{\sqrt{d}}), \quad \ell_i^{-1} = 1- L_i + o(\tfrac{\log^2 n}{d}), \quad b_i b_j = O(\tfrac{\log n}{\sqrt{d}}), \quad \inp{v_i,v_j} = o(\tfrac{\log n}{\sqrt{d}}),\\
        &\Delta \tau^{i,j} = -\tau(b_i b_j + L_i + L_j + o(\tfrac{\log^2 n}{d}) ),\\
        &\Delta p_0^{i,j} = -\lambda_1 \td \Delta  \tau^{i,j} (1+O(\tau\log n)) = \lambda_1 \td \tau (b_i b_j + L_i + L_j + O(\tfrac{\log^2 n}{d}) ),\\
        & \Delta \lambda_1^{i,j} = -\lambda_1 \td \tau \Delta  \tau^{i,j} (1+O(\tau\log n)).
    \end{align*}
\end{lemma}
\begin{proof}
    We first note that $\lambda_1$ can be computed explicitly as follows.
    \begin{align*}
    &\lambda_1 = \frac{1}{\sqrt{\td}}\E_{\xi \sim \cD_{\td}}[\xi \Ind(\xi\geq \sqrt{\td}\tau)]  = \frac{\Gamma(\td/2)}{\td \sqrt{ \pi} \Gamma((\td-1)/2)} \int_{\sqrt{\td}\tau}^{\sqrt{\td}}\xi (1-\xi^2/\td) ^{\frac{\td-3}{2}} d\xi\\
    & = -\frac{\Gamma(\td/2)}{\td \sqrt{ \pi} \Gamma((\td-1)/2)} \frac{\td}{\td-1}(1-\xi^2/\td) ^{\frac{\td-1}{2}} \mid_{\sqrt{\td}\tau}^{\sqrt{\td}} = \frac{\Gamma(\td/2)}{(\td-1) \sqrt{ \pi} \Gamma((\td-1)/2)} (1-\tau^2) ^{\frac{\td-1}{2}}.
    \end{align*}
    The bounds on $L_i$, $\ell_i^{-1}$, $b_ib_j$, $\inp{v_i,v_j}$, and $\Delta \tau^{i,j}$ follow directly from concentration inequalities and definitions. More precisely, $|a_i a_j|=O(\mu^2+\log n/d)$, so $b_i b_j=O(\tau+\log n/(d\tau))$, and hence $|\Delta\tau^{i,j}|\tau d=O(\tau\log n)$. For $\Delta p_0^{i,j}$, we note that
    \begin{align*}
        \Delta p_0^{i,j} &= \Pr_{\xi \sim \cD_{\td}}[\xi \in (\sqrt{\td}\tau^{i,j},\sqrt{\td}\tau)]  = \frac{\Gamma(\td/2)}{\sqrt{ \td \pi} \Gamma((\td-1)/2)} \int_{\sqrt{\td}\tau^{i,j}}^{\sqrt{\td}\tau} (1-\xi^2/\td) ^{\frac{\td-3}{2}} d\xi\\
        & = -\Delta \tau^{i,j} \sqrt{\td} \frac{\Gamma(\td/2)}{\sqrt{ \td \pi} \Gamma((\td-1)/2)}  (1-\tau^2)^{\frac{\td-3}{2}} (1+O(\tau\log n)) \\
        &= -\lambda_1 \td \Delta  \tau^{i,j} (1+O(\tau\log n)).
    \end{align*}
    Similarly, for $\Delta \lambda_1^{i,j}$, we have that
    \begin{align*}
        \Delta \lambda_1^{i,j} &= \frac{1}{\sqrt{\td}}\E_{\xi \sim \cD_{\td}}[\xi \Ind\bigp{ \xi \in (\sqrt{\td}\tau^{i,j},\sqrt{\td}\tau) } ]  = \frac{\Gamma(\td/2)}{\td \sqrt{  \pi} \Gamma((\td-1)/2)} \int_{\sqrt{\td}\tau^{i,j}}^{\sqrt{\td}\tau} \xi (1-\xi^2/\td) ^{\frac{\td-3}{2}}  d\xi\\
        & = -\Delta \tau^{i,j} \td \frac{\Gamma(\td/2)}{\td \sqrt{ \pi} \Gamma((\td-1)/2)}  \tau(1-\tau^2)^{\frac{\td-3}{2}} (1+O(\tau\log n)) \\
        &= -\lambda_1 \td \tau \Delta  \tau^{i,j} (1+O(\tau\log n)).\qedhere
    \end{align*}
\end{proof}
Now we recall two facts that are useful in our proofs.
\begin{fact}\label{fact:bound1}
For $M \in \R^{n\times n}$, define $\|M\|_\infty = \max_{i,j \in [n]}|M_{ij}|$.
Then $\|M\|_\op \le n \|M\|_\infty$.
\end{fact}
\begin{proof}
    This is because $\| M \|_{\mathrm{op}} \leq \| M \|_{\mathrm{F}} \leq n\|M\|_\infty$.
\end{proof}
\begin{fact}\label{fact:boundinp}
Let $w_1, \cdots, w_n \in \cS^{d-1}$ be uniform random vectors in $\cS^{d-1}$, with $n > d$. Then their gram matrix with the diagonal set to zero satisfies
    $\| [\inp{w_i,w_j}]_{0,n\times n} \|_{\mathrm{op}} \leq O \bigp{\frac{n}{d}}$,
with high probability.
\end{fact}
    The proof of \pref{fact:boundinp} follows from standard matrix concentration results (see e.g. \cite{Vershynin}, Theorem 4.6.1 and Theorem 3.4.6).
Now we are ready to prove \pref{lem:opdiff}.
\begin{proof}[Proof of \pref{lem:opdiff}]
We first note that $p_0 = \Theta(p)$ and $\lambda_1 = \Theta(p\tau)$. This follows from the proof of \pref{lem:tau}. For each $i,j \in [n]$ with $i\neq j$, we have that by \pref{lem:bound2},
\begin{align*}
    &\,\, p_0^{i,j}+\td \lambda_1^{i,j} \inp{v_i,v_j}-p_0(1+L_i\lambda_1 \td \tau/p_0 )(1+ L_j \lambda_1 \td \tau/p_0 )-\td \lambda_1\inp{u_i,u_j}\\
    & =  (\Delta p_0^{i,j}+p_0) + \td(\Delta  \lambda_1^{i,j}+ \lambda_1) \inp{v_i,v_j}  - p_0(1+L_i\lambda_1 \td \tau /p_0)(1+ L_j \lambda_1 \td \tau /p_0) \\
    &\hspace{10cm} -\td \lambda_1 (\tau b_i b_j + \ell_i \ell_j \inp{v_i,v_j} )\\
    & =  \lambda_1 \td \tau O(\tfrac{\log^2 n}{d}) -\lambda_1^2 \tau^2 \td^2 L_i L_j/p_0 + \td \Delta \lambda_1^{i,j} \inp{v_i,v_j} - \td \lambda_1 (\ell_i \ell_j-1) \inp{v_i,v_j} \\
    & =: f_1(i,j)+f_2(i,j)+f_3(i,j)+f_4(i,j).
\end{align*}
For simplicity, again, we adopt a notation and write $[a_{i,j}]_{0, n\times n}$ to denote the $n \times n$ matrix where each off-diagonal entry equals $a_{i,j}$ and the diagonal equals $0$. Similarly, we write $[a_{i,j}]_{n\times n}$ as the $n \times n$ matrix where each entry equals $a_{i,j}$. Then we can rewrite our goal as to show that
\begin{align*}
    &\| [p_0^{i,j}+\tilde d \lambda_1^{i,j} \inp{v_i,v_j}-p_0(1+L_i\lambda_1 \td \tau /p_0)(1+ L_j \lambda_1 \td \tau /p_0)-\tilde d  \lambda_1 \inp{u_i,u_j}]_{0,n\times n} \|_{\mathrm{op}} \\
    &\ll np \tau^2\log^4(n).
\end{align*}
Now according to the above computation, the left hand side can be reduced to
\begin{align*}
    \| [f_1(i,j)]_{0,n\times n} + [f_2(i,j)]_{0,n\times n} + [f_3(i,j)]_{0,n\times n} + [f_4(i,j)]_{0,n\times n} \|_{\mathrm{op}}.
\end{align*}
Note that by \pref{lem:bound2} and \pref{fact:bound1}, we have that
\begin{align*}
    \opn{[f_1(i,j)]_{0,n\times n}} \leq n\lambda_1 \tilde d \tau O(\tfrac{\log^2 n}{d}) \leq O \bigp{ np\tau^2 \log^2 n}.
\end{align*}
Similarly, by \pref{lem:bound2} and \pref{fact:bound1}, we have that
\begin{align*}
    \opn{[f_2(i,j)]_{0,n\times n}} \leq n\tfrac{\lambda_1^2 \tau^2 \td^2 }{p_0}o(\tfrac{\log^2 n}{d}) \leq o \bigp{ np\tau^4 d \log^2 n}.
\end{align*}
Furthermore, by \pref{lem:bound2}, \pref{fact:bound1}, and \pref{fact:boundinp}, we have that
\begin{align*}
    & \opn{[f_3(i,j)]_{0,n\times n}} = \opn{ [
\lambda_1 \td^2 \tau^2 (b_i b_j + L_i + L_j + O(\tfrac{\log^2 n}{d}) ) \inp{ v_i,v_j }]_{0,n\times n} }\\
& \leq \opn{ [
\lambda_1 \td^2 \tau^2 b_i b_j \inp{v_i,v_j}]_{0,n\times n} } + 2\opn{ [
\lambda_1 \td^2 \tau^2 L_i \inp{v_i,v_j} ]_{0,n\times n} } + n \lambda_1 \td^2 \tau^2 O(\tfrac{\log^2 n}{d}) o( \tfrac{\log n}{\sqrt{d}})\\
& \leq \lambda_1 \td^2 \tau^2 O \bigp{\tfrac{ \log n}{\sqrt{d}} } O \bigp{\tfrac{ n}{d} } + \lambda_1 \td^2 \tau^2 o \bigp{\tfrac{ \log n}{\sqrt{d}} } O \bigp{\tfrac{ n}{d} } +   n \lambda_1 \td^2 \tau^2 O(\tfrac{\log^2 n}{d}) o( \tfrac{\log n}{\sqrt{d}})\\
& \leq o \bigp {n \lambda_1 \td^2 \tau^2 \tfrac{\log^3 n}{ d\sqrt{d}} } \leq o \bigp{ np \sqrt{d}\tau^3 \log^3 n }.
\end{align*}
Finally, by \pref{lem:bound2}, \pref{fact:bound1}, and \pref{fact:boundinp}, we have that
\begin{align*}
    &\opn{[f_4(i,j)]_{0,n\times n}} = \opn{ [  \td   \lambda_1 (\ell_i \ell_j-1) \inp{v_i,v_j} ]_{0,n\times n} } \\
    &= \opn{ [  \td   \lambda_1 (L_i+L_j + o(\tfrac{\log^2 n
    }{d})) \inp{v_i,v_j} ]_{0,n\times n} }\\
    & = \opn{ [  \td   \lambda_1 (L_i+L_j) \inp{v_i,v_j} ]_{0,n\times n} }+ \opn{ [  \td   \lambda_1 o(\tfrac{\log^2 n}{d}) \inp{v_i,v_j} ]_{0,n\times n}} \\
    & \leq \td   \lambda_1 o \bigp{ \tfrac{\log n}{\sqrt{d}} } O \bigp{ \tfrac{n}{d} } +  n\td   \lambda_1 o \bigp{ \tfrac{\log^2 n}{d} } o \bigp{ \tfrac{\log n}{\sqrt{d}} }  \leq o \bigp { np\tau \log^3n /\sqrt{d} }.
\end{align*}
Combining the bounds for $f_1$, $f_2$, $f_3$, and $f_4$ together, we have that
\begin{align*}
    &\| [f_1(i,j)]_{0,n\times n} + [f_2(i,j)]_{0,n\times n} + [f_3(i,j)]_{0,n\times n} + [f_4(i,j)]_{0,n\times n} \|_{\mathrm{op}} \\
    & \leq o \bigp{ np \sqrt{d}\tau^3 \log^3 n } \leq o \bigp{np \tau^2 \log^4 n }.\qedhere
\end{align*}
\end{proof}

In the rest of the subsection, we will prove the following lemma for relatively larger $\mu$.
\begin{lemma}\label{lem:opdiff2}
    For \mubig, let $D$ be the diagonal matrix of
    $p_0^{i,j}\mathbbm{1}_n \mathbbm{1}_n^\top + \td \lambda_1^{i,j} VV^\top - (p_0\tilde{\mathbbm{1}}_n \tilde{\mathbbm{1}}_n^\top + \td \lambda_1 UU^\top)$. Then, with high probability,
    \begin{align*}
            \| p_0^{i,j}\mathbbm{1}_n \mathbbm{1}_n^\top + \td  \lambda_1^{i,j} VV^\top - (p_0\tilde{\mathbbm{1}}_n \tilde{\mathbbm{1}}_n^\top + \td \lambda_1 UU^\top) -D \|_{\mathrm{op}} \ll npd\mu^4 \log^5(n).
    \end{align*}
\end{lemma}
Adopting the same notation, we have the following lemma.
\begin{lemma}\label{lem:bound22}
    For \mubig, the following holds uniformly with high probability
    \begin{align*}
        & L_i = o(\tfrac{\log n}{\sqrt{d}}), \quad \ell_i^{-1} = 1- L_i + o(\tfrac{\log^2 n}{d}), \quad a_i a_j = O\bigp{\mu^2+\tfrac{\log n}{d}}, \quad \inp{v_i,v_j} = o(\tfrac{\log n}{\sqrt{d}}),\\
        &\Delta \tau^{i,j} = -a_ia_j(1-L_i-L_j+o(\tfrac{\log^2 n}{d}))-\tau(L_i + L_j + o(\tfrac{\log^2 n}{d}) ),\\
        &\Delta p_0^{i,j} = -\lambda_1 \td \Delta  \tau^{i,j} (1+O(\Delta \tau^{i,j}\tau d))\\
        & \qquad = \lambda_1 \td \bigp{ a_ia_j(1-L_i-L_j)+ \tau (L_i + L_j)+ O(\tau^3 \log^2 n)+O(\mu^4 \tau d)},\\
        & \Delta \lambda_1^{i,j} = -\lambda_1 \td \tau \Delta  \tau^{i,j} (1+O(\Delta \tau^{i,j}\tau d))\\
        & \qquad = \lambda_1 \td \tau \bigp{ a_ia_j(1-L_i-L_j)+ \tau (L_i + L_j)+ O(\tau^3 \log^2 n)+O(\mu^4 \tau d)}.
    \end{align*}
\end{lemma}
\begin{proof}
    The bounds on $L_i$, $\ell_i^{-1}$, $a_ia_j$, $\inp{v_i,v_j}$, and $\Delta \tau^{i,j}$ follow directly from concentration inequalities and definitions. For $\Delta p_0^{i,j}$, we note that
    \begin{align*}
        \Delta p_0^{i,j} &= \Pr_{\xi \sim \cD_{\td}}[\xi \in (\sqrt{\td}\tau^{i,j},\sqrt{\td}\tau)]  = \frac{\Gamma(\td/2)}{\sqrt{ \td \pi} \Gamma((\td-1)/2)} \int_{\sqrt{\td}\tau^{i,j}}^{\sqrt{\td}\tau} (1-\xi^2/\td) ^{\frac{\td-3}{2}} d\xi\\
        & = -\Delta \tau^{i,j} \sqrt{\td} \frac{\Gamma(\td/2)}{\sqrt{ \td \pi} \Gamma((\td-1)/2)}  (1-\tau^2)^{\frac{\td-3}{2}} (1+O(\Delta \tau^{i,j} \tau d)) \\
        &= -\lambda_1 \td \Delta  \tau^{i,j} (1+O(\Delta \tau^{i,j} \tau d)).
    \end{align*}
    Similarly, for $\Delta \lambda_1^{i,j}$, we have that
    \begin{align*}
        \Delta \lambda_1^{i,j} &= \frac{1}{\sqrt{\td}}\E_{\xi \sim \cD_{\td}}[\xi \Ind\bigp{ \xi \in (\sqrt{\td}\tau^{i,j},\sqrt{\td}\tau) } ]  = \frac{\Gamma(\td/2)}{\td \sqrt{  \pi} \Gamma((\td-1)/2)} \int_{\sqrt{\td}\tau^{i,j}}^{\sqrt{\td}\tau} \xi (1-\xi^2/\td) ^{\frac{\td-3}{2}}  d\xi\\
        & = -\Delta \tau^{i,j} \td \frac{\Gamma(\td/2)}{\td \sqrt{ \pi} \Gamma((\td-1)/2)}  \tau(1-\tau^2)^{\frac{\td-3}{2}} (1+O(\Delta \tau^{i,j} \tau d)) \\
        &= -\lambda_1 \td \tau \Delta  \tau^{i,j} (1+O(\Delta \tau^{i,j} \tau d)).\qedhere
    \end{align*}
\end{proof}
\begin{proof}[Proof of \pref{lem:opdiff2}]
For each $i,j \in [n]$ with $i\neq j$, we have that by \pref{lem:bound22}
\begin{align*}
    &\,\, p_0^{i,j}+\td \lambda_1^{i,j} \inp{v_i,v_j}-p_0(1+L_i\lambda_1 \td \tau/p_0 )(1+ L_j \lambda_1 \td \tau /p_0)-\td \lambda_1\inp{u_i,u_j}\\
    & =  (\Delta p_0^{i,j}+p_0) + \td(\Delta \lambda_1^{i,j}+  \lambda_1) \inp{v_i,v_j}  - p_0(1+L_i\lambda_1 \td \tau /p_0)(1+ L_j \lambda_1 \td \tau /p_0)  \\
    &\hspace{10cm} -\td  \lambda_1 (a_ia_j + \ell_i \ell_j \inp{v_i,v_j} ) \\
    & =   \td \lambda_1 (O(\tau^3 \log^2 n) + O(\mu^4 \tau d)) -\lambda_1^2 \tau^2 \td^2 L_i L_j /p_0 + \td \Delta  \lambda_1 \inp{v_i,v_j} - \td   \lambda_1 (\ell_i \ell_j-1) \inp{v_i,v_j} \\
    & =: f_1(i,j)+f_2(i,j)+f_3(i,j)+f_4(i,j).
\end{align*}
We follow the same proof strategy as in \pref{lem:opdiff}, with slightly different bounds.
Note that by \pref{lem:bound22} and \pref{fact:bound1}, we have that
\begin{align*}
    \opn{[f_1(i,j)]_{0,n\times n}} \leq n\td \lambda_1 O(\tau^3 \log^2 n) + n \td \lambda_1 O(\mu^4 \tau d)\leq O \bigp{ np d \tau^4 \log^2 n}+ O(n p d^2 \tau^2 \mu^4).
\end{align*}
Similarly, by \pref{lem:bound22} and \pref{fact:bound1}, we have that
\begin{align*}
    \opn{[f_2(i,j)]_{0,n\times n}} \leq n\tfrac{\lambda_1^2 \tau^2 \td^2 }{p_0}o(\tfrac{\log^2 n}{d}) \leq o \bigp{ np\tau^4 d \log^2 n}.
\end{align*}
Furthermore, by \pref{lem:bound22}, \pref{fact:bound1}, and \pref{fact:boundinp}, we have that
\begin{align*}
    & \opn{[f_3(i,j)]_{0,n\times n}} \\
    &= \opn{ [
\lambda_1 \td^2 \tau \bigp{ a_ia_j(1-L_i-L_j)+ \tau (L_i + L_j)+ O(\tau^3 \log^2 n) +O(\mu^4 \tau d)} \inp{ v_i,v_j }]_{0,n\times n} }\\
& \leq \opn{ [
\lambda_1 \td^2 \tau a_i a_j \inp{v_i,v_j}]_{0,n\times n} } + 2\opn{ [
\lambda_1 \td^2 \tau a_i a_j L_i \inp{v_i,v_j} ]_{0,n\times n} }+ 2\opn{ [
\lambda_1 \td^2 \tau^2 L_i \inp{v_i,v_j} ]_{0,n\times n} } \\
& \hspace{6cm} + n \lambda_1 \td^2 \tau O(\tau^3 \log^2 n) o( \tfrac{\log n}{\sqrt{d}}) + n \lambda_1 \td^2 \tau O(\mu^4 \tau d) O( \tfrac{\log n}{\sqrt{d}}) \\
& \leq \lambda_1 \td^2 \tau \bigp{\mu^2+\tfrac{\log n}{d}} O \bigp{\tfrac{ n}{d} } + \lambda_1 \td^2 \tau \bigp{\mu^2+\tfrac{\log n}{d}} o \bigp{\tfrac{ \log n}{\sqrt{d}} } O \bigp{\tfrac{ n}{d} } + \lambda_1 \td^2 \tau^2 o \bigp{\tfrac{ \log n}{\sqrt{d}} } O \bigp{\tfrac{ n}{d} } \\
& \hspace{6cm} + n \lambda_1 \td^2 \tau O(\tau^3 \log^2 n) o( \tfrac{\log n}{\sqrt{d}}) + n \lambda_1 \td^2 \tau O(\mu^4 \tau d) O( \tfrac{\log n}{\sqrt{d}}) \\
& \leq o \bigp { np\tau^2 \log^{9/2} n} + O\bigp{npd \mu^4 \log^{5/2}n}.
\end{align*}
Finally, by \pref{lem:bound22}, \pref{fact:bound1}, and \pref{fact:boundinp}, we have that
\begin{align*}
    &\opn{[f_4(i,j)]_{0,n\times n}} = \opn{ [  \td   \lambda_1 (\ell_i \ell_j-1) \inp{v_i,v_j} ]_{0,n\times n} } \\
    &= \opn{ [  \td   \lambda_1 (L_i+L_j + o(\tfrac{\log^2 n
    }{d})) \inp{v_i,v_j} ]_{0,n\times n} }\\
    & = \opn{ [  \td   \lambda_1 (L_i+L_j) \inp{v_i,v_j} ]_{0,n\times n} }+ \opn{ [  \td   \lambda_1 o(\tfrac{\log^2 n}{d}) \inp{v_i,v_j} ]_{0,n\times n}} \\
    & \leq \td   \lambda_1 o \bigp{ \tfrac{\log n}{\sqrt{d}} } O \bigp{ \tfrac{n}{d} } +  n\td   \lambda_1 o \bigp{ \tfrac{\log^2 n}{d} } o \bigp{ \tfrac{\log n}{\sqrt{d}} }  \leq o \bigp { np\tau \log^3n /\sqrt{d} }.
\end{align*}
Combining the bounds for $f_1$, $f_2$, $f_3$, and $f_4$ together, we have that
\begin{align*}
    &\| [f_1(i,j)]_{0,n\times n} + [f_2(i,j)]_{0,n\times n} + [f_3(i,j)]_{0,n\times n} + [f_4(i,j)]_{0,n\times n} \|_{\mathrm{op}} \\
    & \leq o \bigp { np\tau^2 \log^{9/2} n} + O\bigp{npd \mu^4 \log^{5/2}n} = o(npd\mu^4 \log ^5 n).\qedhere
\end{align*}
\end{proof}

\subsection{Accounting for the diagonal}
In this section, we will prove \pref{p:op1} by combining \pref{p:op2} with \pref{lem:opdiff}, and prove \pref{p:largemu} by combining \pref{p:op2} with \pref{lem:opdiff2}. Set
\begin{align*}
D:=\diag\!\left(A-p_0\tilde{\mathbbm{1}}_n \tilde{\mathbbm{1}}_n^\top-\td \lambda_1 UU^\top\right).
\end{align*}
Directly combining \pref{p:op2} with \pref{lem:opdiff}, we have that for \musmall,
\begin{align*}
    \| A - (p_0\tilde{\mathbbm{1}}_n \tilde{\mathbbm{1}}_n^\top + \td \lambda_1 UU^\top) -D \|_{\mathrm{op}} \llwhp \log^9(n)\max(np \tau^2, \sqrt{np}).
\end{align*}
Similarly, directly combining \pref{p:op2} with \pref{lem:opdiff2}, we have that for \mubig,
\begin{align*}
    \| A - (p_0\tilde{\mathbbm{1}}_n \tilde{\mathbbm{1}}_n^\top + \td  \lambda_1 UU^\top) -D \|_{\mathrm{op}} \llwhp \log^9(n)\max(npd \mu^4, \sqrt{np}).
\end{align*}
The remaining task is to bound the diagonal of $A - (p_0\tilde{\mathbbm{1}}_n \tilde{\mathbbm{1}}_n^\top + \td \lambda_1 UU^\top)$. Note that we have that with high probability,
\begin{align*}
    p_0 (\tilde{\mathbbm{1}}_n)_i^2 = p_0 (1+L_i\lambda_1 \td \tau/p_0)^2 = O(p_0).
\end{align*}
Furthermore, with high probability,
\begin{align*}
    \td \lambda_1 \inp{u_i,u_i} = \td \lambda_1 (a_i^2+ \ell_i^2) = O( d p\tau ).
\end{align*}
Therefore, for \musmall, with high probability,
\begin{align*}
   &\|A - (p_0\tilde{\mathbbm{1}}_n \tilde{\mathbbm{1}}_n^\top + \td \lambda_1 UU^\top)\|_{\mathrm{op}} \\
   &\leq \| A - (p_0\tilde{\mathbbm{1}}_n \tilde{\mathbbm{1}}_n^\top + \td \lambda_1 UU^\top) -D \|_{\mathrm{op}} + \opn{ \diag (A - (p_0\tilde{\mathbbm{1}}_n \tilde{\mathbbm{1}}_n^\top + \td \lambda_1 UU^\top)) }\\
   & \ll \log^9(n)\max(np \tau^2, \sqrt{np}) + O(p_0)+ O( d p\tau )\\
   & \ll \log^9(n)\max(np \tau^2, \sqrt{np}).
\end{align*}
Similarly, for \mubig, we have that with high probability,
\begin{align*}
   &\|A - (p_0\tilde{\mathbbm{1}}_n \tilde{\mathbbm{1}}_n^\top + \td \lambda_1 UU^\top)\|_{\mathrm{op}} \\
   &\leq \| A - (p_0\tilde{\mathbbm{1}}_n \tilde{\mathbbm{1}}_n^\top + \td \lambda_1 UU^\top) -D \|_{\mathrm{op}} + \opn{ \diag (A - (p_0\tilde{\mathbbm{1}}_n \tilde{\mathbbm{1}}_n^\top + \td \lambda_1 UU^\top)) }\\
   & \ll \log^9(n)\max(npd\mu^4, \sqrt{np}) + O(p_0)+ O( d p\tau )\\
   & \ll \log^9(n)\max(npd\mu^4, \sqrt{np}).
\end{align*}

\subsection{Hypothesis testing}
We begin by recalling our hypothesis testing algorithm.
Define $\tau'$ to be the connectivity threshold for the one-community model $\geo{n}{d}{p}{0}$. Correspondingly, as in equation \pref{eq:deflambda}, we define $\lambda_k'$ to be the normalized Gegenbauer polynomial expansion coefficient of $\Ind(x\geq \sqrt{\td} \tau')$. Put
\begin{align*}
 M_n:=\max \left\{ \sqrt{\tfrac{\log 1/p}{d}},
              \sqrt{\tfrac{d}{np\log(1/p)}}\right\}\log^9 n,
 \qquad
 \mu_\star:=d^{-1/4}\log^{-1/2}n.
\end{align*}
In the perturbative range $\mu\leq\mu_\star$, we reject the null when the second largest eigenvalue of $A$ satisfies
\begin{align*}
    \eta_1 > n\lambda_1' \bigp{ 1+ \frac{1}{2}M_n}.
\end{align*}
For $\mu>\mu_\star$, let $p_{\mathrm{in}}$ and $p_{\mathrm{out}}$ denote the within- and across-community edge probabilities under the specified alternative, and put
\begin{align*}
 \Delta_\mu:=\frac{p_{\mathrm{in}}-p_{\mathrm{out}}}{2}.
\end{align*}
In this large-separation range we use the coarser spectral threshold $\eta_1>n\Delta_\mu/2$. Both thresholds are deterministic functions of the parameters of the two simple hypotheses. If the relevant inequality holds, we declare the model to be the two-community mixture model; otherwise, we declare it to be the one-community model.

\begin{theorem*}[Restatement of \pref{thm:testingmain}]
Define the one-community model to be the null hypothesis $H_0 = \geo{n}{d}{p}{0}$ and the separated mixture model to be the alternative hypothesis $H_1 = \geo{n}{d}{p}{\mu}$. If $d,n,\mu$ satisfy
\begin{align*}
    \mu^2 \geq \max \left\{ \sqrt{\tfrac{\log 1/p}{d^{3}}}, \sqrt{\tfrac{1}{np d \log\frac{1}{p}}} \right\} \log^9 n, \quad \log^{16} n\ll d < n, \quad pn \gg 1, \quad p \in [0,1/2-\varepsilon],
\end{align*}
then if we run the spectral algorithm described above on input graph $G$ we have that
\begin{align*}
    &\min\left\{\Pr(\text{accept } H_0 \mid G \sim H_0),\, \Pr(\text{reject } H_0 \mid G \sim H_1)\right\}\geq 1-o_n(1),
\end{align*}
In other words, both type 1 error and type 2 error go to zero as $n$ goes to infinity.
\end{theorem*}

We will first prove \pref{thm:testing} and \pref{thm:testing2}. Before proving the testing theorem, we record that the first Gegenbauer coefficient under the alternative is sufficiently close to its null counterpart.

\begin{lemma}\label{lem:null-mixture-coefficients}
Let $\tau'$ be the connectivity threshold in the null model and let $\lambda_1'$ be its first Gegenbauer coefficient. Suppose that
\begin{align*}
 \log^{16}n\ll d<n,\qquad pn\gg1,\qquad
 p\in[0,1/2-\varepsilon],\qquad
 \mu\leq d^{-1/4}\log^{-1/2}n.
\end{align*}
Writing $L:=\log(1/p)$, we have
\begin{align}
 |\tau-\tau'|
 &\leq C_\varepsilon\left(\mu^2+\mu\sqrt{\frac{\log n}{d}}\right)+n^{-20},
 \label{eq:tau-null-alternative-comparison}\\
 \left|\log\frac{\lambda_1}{\lambda_1'}\right|
 &\leq C_\varepsilon\left(\sqrt{dL}\,\mu^2+\mu\sqrt{L\log n}\right)+o(n^{-10}).
 \label{eq:lambda-null-alternative-comparison}
\end{align}
In particular, put
\begin{align*}
 a_n:=d\mu^2,
 \qquad
 M_n:=\max\left\{\sqrt{\frac{L}{d}},\sqrt{\frac{d}{npL}}\right\}\log^9 n.
\end{align*}
If $a_n\geq cM_n$ for some fixed $c>0$ and $\mu\leq\tau$, then
\begin{equation}\label{eq:lambda-ratio-testing}
 \frac{\lambda_1}{\lambda_1'}=1+o\bigp{\min\{a_n,1\}}.
\end{equation}
Moreover, uniformly for $\tau<\mu\leq\mu_\star$, we have
\begin{equation}\label{eq:lambda-ratio-testing-large}
 \frac{\lambda_1}{\lambda_1'}=1+o(1).
\end{equation}
\end{lemma}

\begin{proof}
Couple the null and mixture inner products by writing
\begin{align*}
\begin{gathered}
 X:=\langle g_i,g_j\rangle,
 \qquad
 Y:=\langle u_i,u_j\rangle,\\
 g_i=(N_i,w_i),\quad g_j=(N_j,w_j),
 \qquad
 g_i,g_j\stackrel{\mathrm{i.i.d.}}{\sim}\calN(0,d^{-1}\Id_d).
\end{gathered}
\end{align*}
so that, with independent Rademacher variables $S_i,S_j$,
\begin{align*}
 Y-X=\mu(S_iN_j+S_jN_i)+\mu^2S_iS_j.
\end{align*}
For a sufficiently large fixed constant $K$, Gaussian tails give an event of probability at least $1-n^{-K}$ on which
\begin{align*}
 |Y-X|\leq
 C_K\left(\mu^2+\mu\sqrt{\frac{\log n}{d}}\right)=:\delta_n.
\end{align*}
Consequently, uniformly in $t$,
\begin{align*}
 \Pr(X\geq t+\delta_n)-n^{-K}
 \leq \Pr(Y\geq t)
 \leq \Pr(X\geq t-\delta_n)+n^{-K}.
\end{align*}
Let $F_X(t):=\Pr(X\geq t)$ and let $f_X$ denote the density of $X$. By \pref{lem:taubound1}, both $\tau$ and $\tau'$ lie between constant multiples of $d^{-1/2}$ and $\sqrt{L/d}$. Conditional on $g_i$, the variable $X$ is Gaussian with variance $\|g_i\|^2/d$. The same norm-concentration calculation used in the proof of \pref{lem:taubound1} therefore gives, uniformly in this range,
\begin{align*}
 F_X(t)=\Theta_\varepsilon(\Psi(t\sqrt d)),
 \qquad
 f_X(t)=\Theta_\varepsilon(\sqrt d\,\phi(t\sqrt d)),
 \qquad
 f_X(t)\geq c_\varepsilon\sqrt d\,F_X(t),
\end{align*}
where $\phi$ is the standard Gaussian density. Evaluating the preceding coupling inequalities at $t=\tau$ and using $F_X(\tau')=p$, monotonicity and the mean-value theorem show that
\begin{align*}
 |\tau-\tau'|\leq\delta_n+\frac{C_\varepsilon n^{-K}}{p\sqrt d}.
\end{align*}
Since $np\to\infty$, choosing $K$ sufficiently large proves \pref{eq:tau-null-alternative-comparison}.

For $t=o(1)$, the first Gegenbauer coefficient has the exact formula
\begin{align*}
 \lambda_1(t)
 =\frac{\Gamma(\td/2)}{(\td-1)\sqrt\pi\,\Gamma((\td-1)/2)}
   (1-t^2)^{(\td-1)/2}.
\end{align*}
Hence, by the mean-value theorem and \pref{lem:taubound1},
\begin{align*}
 \left|\log\frac{\lambda_1}{\lambda_1'}\right|
 &\leq C d(\tau+\tau')|\tau-\tau'|\\
 &\leq C_\varepsilon\left(\sqrt{dL}\,\mu^2+\mu\sqrt{L\log n}\right)+o(n^{-10}),
\end{align*}
which proves \pref{eq:lambda-null-alternative-comparison}.

It remains to verify the final assertions. The first term on the right of \pref{eq:lambda-null-alternative-comparison} equals $a_nO(\sqrt{L/d})=o(a_n)$. If $a_n\leq1$, the lower bound $a_n\geq cM_n$ gives
\begin{align*}
 \frac{\mu\sqrt{L\log n}}{a_n}
 =\sqrt{\frac{L\log n}{a_nd}}
 \leq C\sqrt{\frac{\sqrt L}{\sqrt d\,\log^8n}}
 =o(1).
\end{align*}
If $a_n\geq1$ and $\mu\leq\tau$, both terms in \pref{eq:lambda-null-alternative-comparison} are $o(1)$. This proves \pref{eq:lambda-ratio-testing}. Finally, when $\tau<\mu\leq\mu_\star$, the bounds $L\leq(1+o(1))\log n$ and $d\gg\log^{16}n$ give
\begin{align*}
 \sqrt{dL}\,\mu^2\leq \frac{\sqrt L}{\log n}=o(1),
 \qquad
 \mu\sqrt{L\log n}\leq d^{-1/4}\sqrt L=o(1).
\end{align*}
Exponentiating \pref{eq:lambda-null-alternative-comparison} therefore proves \pref{eq:lambda-ratio-testing-large}.
\end{proof}

As the vector $\tilde{\mathbbm{1}}_n$ is not necessarily orthogonal to columns of $U$, we next prove a proposition that shows that they are not far away from orthogonality, and in fact the same results hold for the projection. Define a projected matrix as
\begin{align*}
    \calP(UU^\top) := \bigp{I- \frac{ \tilde{\mathbbm{1}}_n \tilde{\mathbbm{1}}_n^\top }{ \|\tilde{\mathbbm{1}}_n\|^2 }} U U^\top \bigp{I- \frac{ \tilde{\mathbbm{1}}_n \tilde{\mathbbm{1}}_n^\top }{ \|\tilde{\mathbbm{1}}_n\|^2 }}.
\end{align*}
\begin{proposition}\label{p:ortho}
For \muall, we have that the following holds with high probability,
\begin{align*}
    &\| A- p_0\tilde{\mathbbm{1}}_n \tilde{\mathbbm{1}}_n^\top - \td \lambda_1 \calP(UU^\top)\|_{\mathrm{op}} \ll  \log^9(n) \max\left\{ np\tau^2, \sqrt{np} \right \},  \text{ if } \mu\leq \tau,\\
    &\| A- p_0\tilde{\mathbbm{1}}_n \tilde{\mathbbm{1}}_n^\top - \td \lambda_1 \calP(UU^\top)\|_{\mathrm{op}} \ll  \log^9(n) \max\left\{ npd\mu^4,  \sqrt{np} \right \}, \text{ if } \tau < \mu \leq d^{-\tfrac{1}{4}} \log ^{-\tfrac{1}{2}} (n).
\end{align*}
\end{proposition}
\begin{proof}[Proof of \pref{p:ortho}]
Write $v:=\tilde{\mathbbm{1}}_n$ and let
\begin{align*}
Q:=\frac{vv^\top}{\|v\|^2},\qquad P:=I-Q,
\end{align*}
so that $\calP(UU^\top)=PUU^\top P$. By \pref{lem:bound2} and \pref{lem:tau}, the entries of $v$ are $1+o(1)$ uniformly, and hence $\|v\|^2=\Theta(n)$ with high probability.

Write $U=(U^1\;W)$, where $U^1$ is the first column and $W$ contains the remaining $d-1$ columns. The first-column weights are independent of $U^1$, while the summands defining $v^\top W$ are independent across rows and centered by rotational symmetry. Standard scalar and vector Bernstein inequalities, together with the uniform bounds in \pref{lem:bound2}, therefore give, with high probability,
\begin{align*}
 |v^\top U^1|&=O\bigp{\sqrt{n(\mu^2+d^{-1})\log n}},
 &\|U^1\|&=O\bigp{\sqrt{n(\mu^2+d^{-1})}},\\
 \|v^\top W\|&=O(\sqrt{n\log n}),
 &\|W\|_{\mathrm{op}}&=O\bigp{\sqrt{n/d}}.
\end{align*}
Consequently,
\begin{align*}
 \|QUU^\top\|_{\mathrm{op}}
 &\leq \frac{|v^\top U^1|\,\|U^1\|+\|v^\top W\|\,\|W\|_{\mathrm{op}}}{\|v\|}\\
 &=O\bigp{\sqrt{n\log n}(\mu^2+d^{-1})+\sqrt{n\log n/d}}.
\end{align*}
Moreover,
\begin{align*}
 \|QUU^\top Q\|_{\mathrm{op}}
 =\frac{\|v^\top U\|^2}{\|v\|^2}=O(\log n).
\end{align*}
Since
\begin{align*}
 UU^\top-PUU^\top P=QUU^\top+UU^\top Q-QUU^\top Q,
\end{align*}
we obtain
\begin{equation}\label{eq:projection-raw}
 \|UU^\top-\calP(UU^\top)\|_{\mathrm{op}}
 =O\bigp{\sqrt{n\log n}(\mu^2+d^{-1})+\sqrt{n\log n/d}+\log n}.
\end{equation}

It remains to compare this with the error scales in the proposition. Recall from the proof of \pref{lem:bound2} that $\td\lambda_1=\Theta(dp\tau)$ and from \pref{lem:tau} that $d\tau^2=O(\log(1/p))$. The terms in \pref{eq:projection-raw} not containing $\mu^2$, after multiplication by $\td\lambda_1$, are all $O(\sqrt{np}\log^2 n)$. For the remaining term, set
\begin{align*}
 H:=dp\tau\mu^2\sqrt{n\log n}.
\end{align*}
If $\mu\leq\tau$, then
\begin{align*}
 \frac{H}{np\tau^2}
 \leq \frac{d\tau\sqrt{\log n}}{\sqrt n}
 =O(\log n).
\end{align*}
If $\mu>\tau$, set $X:=npd\mu^4$ and $Y:=\sqrt{np}$. Then
\begin{align*}
 \frac{H}{\max\{X,Y\}}
 \leq \sqrt{\frac{H}{X}\frac{H}{Y}}
 =\sqrt{\frac{d\sqrt p\,\tau^2\log n}{\sqrt n}}
 =O(\log n).
\end{align*}
Here we also used $d<n$, $np\gg1$, and $p\log(1/p)=O(1)$. Thus, with high probability,
\begin{equation}\label{eq:projection}
\begin{aligned}
 \td\lambda_1\|UU^\top-\calP(UU^\top)\|_{\mathrm{op}}
 &\ll \log^9 n\max\{np\tau^2,\sqrt{np}\}, && \mu\leq\tau,\\
 \td\lambda_1\|UU^\top-\calP(UU^\top)\|_{\mathrm{op}}
 &\ll \log^9 n\max\{npd\mu^4,\sqrt{np}\}, && \mu>\tau.
\end{aligned}
\end{equation}
Combining \pref{eq:projection} with \pref{p:op1} in the first regime and with \pref{p:largemu} in the second proves the proposition.
\end{proof}
To facilitate the proof, we cite a useful result concerning the spectrum of $UU^\top$.
\begin{lemma}[Spectrum of $UU^\top$]\label{lem:uuspec}
    Let $\lambda_1(UU^\top)$ be the largest eigenvalue of $UU^\top$, where each row is sampled from the Gaussian mixture distribution $\frac{1}{2} \calN(-\mu \cdot e_1, \tfrac{1}{d}\Id_d) + \frac{1}{2} \calN(\mu \cdot e_1, \tfrac{1}{d}\Id_d)$. Then we have that
    \begin{align*}
        \abs{\lambda_1(UU^\top)-\bigp{\mu^2 n + \frac{n}{d} } } \leq O \bigp{ \sqrt{\frac{d}{n}} \bigp{\mu^2 n + \frac{n}{d} }},
    \end{align*}
    with high probability. Similarly, if each row is sampled from the Gaussian distribution $\calN(0, \tfrac{1}{d}\Id_d)$, then we have that
    \begin{align*}
        \abs{\lambda_1(UU^\top)-\bigp{\frac{n}{d} }} \leq O \bigp{\sqrt{\frac{n}{d}} },
    \end{align*}
    with high probability.
\end{lemma}
\begin{proof}
    This follows from Theorem~4.6.1 in \cite{Vershynin}.
\end{proof}

\begin{proof}[Proof of \pref{thm:testing}]
    Recall that (see the definitions before \pref{thm:testing}) we used $\eta_0$ and $w_0$ to denote the first eigenvalue and eigenvector of the adjacency matrix $A$ respectively. Similarly, as before, we use $\td \lambda_1 \sU\sU^\top$ to denote the spectral truncation of $A$ to the span of the eigenvectors corresponding to its second through $(d+1)$-st eigenvalues. By \pref{p:op1}, we have that
    \begin{align*}
        \| A- \eta_0w_0 w_0^\top - \td \lambda_1 \sU\sU^\top\|_{\mathrm{op}} \ll  \log^9(n) \max\left\{ np\tau^2, \sqrt{np} \right \}.
    \end{align*}
    Combining with \pref{p:ortho}, we have that
    \begin{align*}
        \| \eta_0w_0 w_0^\top + \td \lambda_1 \sU\sU^\top
        -p_0\tilde{\mathbbm{1}}_n \tilde{\mathbbm{1}}_n^\top - \td \lambda_1 \calP(UU^\top)
        \|_{\mathrm{op}} \ll  \log^9(n) \max\left\{ np\tau^2, \sqrt{np} \right \}.
    \end{align*}
    We note that the first eigenvalue of $p_0\tilde{\mathbbm{1}}_n \tilde{\mathbbm{1}}_n^\top + \td \lambda_1 \calP(UU^\top)$ is $p_0 \|\tilde{\mathbbm{1}}_n\|^2 = \Theta(pn)$ and the second eigenvalue of it satisfies
    \begin{align*}
        &\lambda_2(p_0\tilde{\mathbbm{1}}_n \tilde{\mathbbm{1}}_n^\top + \td \lambda_1 \calP(UU^\top))
        = \td \lambda_1 \cdot \lambda_1 (\calP(UU^\top)) \\
        &\leq \td \lambda_1\lambda_1(UU^\top)
        \leq \td \lambda_1 O\bigp{ \mu^2 n+ \frac{n}{d}+\sqrt{\frac{d}{n}} \bigp{\mu^2 n +\frac{n}{d}}}\ll pn,
    \end{align*}
    because an orthogonal compression cannot increase operator norm, by \pref{lem:uuspec}, \pref{lem:tau}, and $\mu^2\leq d^{-1/2}\log^{-1}n$. Thus, by the  Davis-Kahan $\sin\theta$ theorem (Theorem 4.5.5 in \cite{Vershynin}), we have that there exist $\theta \in \{\pm 1\}$, such that
    \begin{align}\label{eq:1steig}
        \left\| \theta w_0-\frac{ \tilde{\mathbbm{1}}_n}{\|\tilde{\mathbbm{1}}_n\|} \right \|_2 \ll \frac{\log^9(n) \max\left\{ np\tau^2, \sqrt{np} \right \}}{pn}.
    \end{align}
    Without loss of generality, assume $\theta = 1$. Therefore by \pref{eq:1steig}, we have that
    \begin{align*}
        &\| \eta_0w_0 w_0^\top
        -p_0\tilde{\mathbbm{1}}_n \tilde{\mathbbm{1}}_n^\top
        \|_{\mathrm{op}} \\
        &\leq |\eta_0-p_0 \|\tilde{\mathbbm{1}}_n\|^2 | + p_0 \|\tilde{\mathbbm{1}}_n\|^2  \| (w_0 w_0^\top- \tilde{\mathbbm{1}}_n \tilde{\mathbbm{1}}_n^\top / \|\tilde{\mathbbm{1}}_n\|^2 ) \|_\mathrm{op} \\
        &\leq |\eta_0-p_0 \|\tilde{\mathbbm{1}}_n\|^2 |+  2p_0 \|\tilde{\mathbbm{1}}_n\|^2
        \| w_0 - \tilde{\mathbbm{1}}_n/\|\tilde{\mathbbm{1}}_n \|
        \| \ll \log^9(n) \max\left\{ np\tau^2, \sqrt{np} \right \}.
    \end{align*}
    Combining the preceding estimate with the scaled projection bound \pref{eq:projection}, we have that
    \begin{align*}
        \| \td\lambda_1 \sU\sU^\top - \td \lambda_1 UU^\top
        \|_{\mathrm{op}}
        \ll \log^9(n) \max\left\{ np\tau^2, \sqrt{np} \right \}.
    \end{align*}
    Since $\td\lambda_1=\Theta(dp\tau)$, division by $\td\lambda_1$ yields
    \begin{align*}
        \| \sU\sU^\top - UU^\top\|_{\mathrm{op}}
        &\ll \max \left\{ \frac{n\tau}{d}, \frac{\sqrt{n}}{\sqrt{p}d\tau} \right\}\log^9 n.
    \end{align*}
\end{proof}
\begin{proof}[Proof of \pref{thm:testing2}]
    We use the same proof idea for large $\mu$. By \pref{p:largemu}, we have that
    \begin{align*}
        \| A- \eta_0w_0 w_0^\top - \td \lambda_1 \sU\sU^\top\|_{\mathrm{op}} \ll \log^9(n) \max\left\{ npd\mu^4,  \sqrt{np} \right \}.
    \end{align*}
    Combining with \pref{p:ortho}, we have that
    \begin{align*}
        \| \eta_0w_0 w_0^\top + \td \lambda_1 \sU\sU^\top
        -p_0\tilde{\mathbbm{1}}_n \tilde{\mathbbm{1}}_n^\top - \td \lambda_1 \calP(UU^\top)
        \|_{\mathrm{op}} \ll  \log^9(n) \max\left\{ npd\mu^4, \sqrt{np} \right \}.
    \end{align*}
    We note that the first eigenvalue of $p_0\tilde{\mathbbm{1}}_n \tilde{\mathbbm{1}}_n^\top + \td \lambda_1 \calP(UU^\top)$ is $p_0 \|\tilde{\mathbbm{1}}_n\|^2 = \Theta(pn)$ and the second eigenvalue of it satisfies
    \begin{align*}
        &\lambda_2(p_0\tilde{\mathbbm{1}}_n \tilde{\mathbbm{1}}_n^\top + \td \lambda_1 \calP(UU^\top))
        = \td \lambda_1 \cdot \lambda_1 (\calP(UU^\top)) \\&\leq \td \lambda_1\lambda_1(UU^\top)
        \leq \td \lambda_1 O\bigp{ \mu^2 n+ \frac{n}{d}+\sqrt{\frac{d}{n}} \bigp{\mu^2 n +\frac{n}{d}}}\ll pn,
    \end{align*}
    because an orthogonal compression cannot increase operator norm, by \pref{lem:uuspec}, \pref{lem:tau}, and $\mu^2\leq d^{-1/2}\log^{-1}n$. Thus, by the  Davis-Kahan $\sin\theta$ theorem (Theorem 4.5.5 in \cite{Vershynin}), we have that there exist $\theta \in \{\pm 1\}$, such that
    \begin{align}\label{eq:1steig2}
        \left\| \theta w_0-\frac{ \tilde{\mathbbm{1}}_n}{\|\tilde{\mathbbm{1}}_n\|} \right \|_2 \ll \frac{\log^9(n) \max\left\{ npd\mu^4, \sqrt{np} \right \}}{pn}.
    \end{align}
    Without loss of generality, assume $\theta = 1$. Therefore by \pref{eq:1steig2}, we have that
    \begin{align*}
        &\| \eta_0w_0 w_0^\top
        -p_0\tilde{\mathbbm{1}}_n \tilde{\mathbbm{1}}_n^\top
        \|_{\mathrm{op}} \\
        &\leq |\eta_0-p_0 \|\tilde{\mathbbm{1}}_n\|^2 | + p_0 \|\tilde{\mathbbm{1}}_n\|^2  \| (w_0 w_0^\top- \tilde{\mathbbm{1}}_n \tilde{\mathbbm{1}}_n^\top / \|\tilde{\mathbbm{1}}_n\|^2 ) \|_\mathrm{op} \\
        &\leq |\eta_0-p_0 \|\tilde{\mathbbm{1}}_n\|^2 |+  2p_0 \|\tilde{\mathbbm{1}}_n\|^2
        \| w_0 - \tilde{\mathbbm{1}}_n/\|\tilde{\mathbbm{1}}_n \|
        \| \ll \log^9(n) \max\left\{ npd\mu^4, \sqrt{np} \right \}.
    \end{align*}
    Combining the preceding estimate with the scaled projection bound \pref{eq:projection}, we have that
    \begin{align*}
        \| \td\lambda_1 \sU\sU^\top - \td \lambda_1 UU^\top
        \|_{\mathrm{op}}
        \ll \log^9(n) \max\left\{ npd\mu^4, \sqrt{np} \right \}.
    \end{align*}
    Since $\td\lambda_1=\Theta(dp\tau)$, division by $\td\lambda_1$ yields
    \begin{align*}
        \| \sU\sU^\top - UU^\top\|_{\mathrm{op}}
        &\ll \max \left\{ \frac{n\mu^4}{\tau}, \frac{\sqrt{n}}{\sqrt{p}d\tau} \right\}\log^9 n.
    \end{align*}
\end{proof}

\begin{proof}[Proof of \pref{thm:testingmain}]
We use subscripts $\mathrm g$ and $\mathrm m$ for the one-community and mixture models, respectively, and write
\begin{align*}
 L:=\log(1/p),\qquad
 M_n:=\max\left\{\sqrt{\frac{L}{d}},\sqrt{\frac{d}{npL}}\right\}\log^9n,
 \qquad a_n:=d\mu^2.
\end{align*}
The signal assumption in the theorem is exactly $a_n\geq M_n$.

We first control the null model. Combining \pref{thm:testing} with the null part of \pref{lem:uuspec}, and using the definition of $\sU_{\mathrm g}$, gives
\begin{align*}
 \frac{\eta_1(A_{\mathrm g})}{n\lambda_1'}
 &\leq \frac{\td}{d}+O\left(\sqrt{\frac dn}\right)
 +o\left(\max\left\{\tau',\frac{1}{\sqrt{np}\,\tau'}\right\}\log^9n\right)\\
 &=1+o(M_n)
\end{align*}
with high probability. Here we used $\tau'=\Theta(\sqrt{L/d})$ from \pref{lem:taubound1}. Hence
\begin{align*}
 \eta_1(A_{\mathrm g})<n\lambda_1'\left(1+\frac12M_n\right)
\end{align*}
with probability $1-o(1)$.

Suppose first that $\mu\leq\tau$. A scalar Bernstein bound gives
\begin{align*}
 \lambda_1(U_{\mathrm m}U_{\mathrm m}^{\top})
 \geq n\left(\mu^2+\frac1d\right)(1-o(1))
\end{align*}
with high probability. Combining this with \pref{thm:testing} and \pref{eq:lambda-ratio-testing} yields
\begin{align*}
 \frac{\eta_1(A_{\mathrm m})}{n\lambda_1'}
 \geq 1+a_n-o(a_n)>1+\frac12M_n.
\end{align*}
Thus the perturbative threshold rejects the null with probability $1-o(1)$ in this range.

The same conclusion holds throughout the remaining perturbative range
$\tau<\mu\leq\mu_\star$. Indeed, for testing we only need the two-dimensional compression of $A$ onto the degree direction and the first-coordinate signal direction, rather than the full operator-norm approximation in \pref{thm:testing2}. Applying the same Gegenbauer expansion and Hoeffding decomposition used in the proof of \pref{p:largemu} to this single compression gives
\begin{equation}\label{eq:testing-two-dimensional-compression}
 \frac{\eta_1(A_{\mathrm m})}{n\lambda_1'}
 \geq 1+a_n-o(a_n)
\end{equation}
with high probability, uniformly for $\tau<\mu\leq\mu_\star$. To see why the estimate is uniform, the only additional deterministic remainder has size $O(np d\mu^4)$, without the high-moment $\log^9n$ loss needed for a global operator norm; relative to the community spike $np\tau d\mu^2$ its ratio is
\begin{align*}
 O\left(\frac{\mu^2}{\tau}\right)
 \leq O\left(\frac{1}{\log n\sqrt{L}}\right)=o(1).
\end{align*}
The centered scalar fluctuation is $O_{\mathbb P}(np\tau^2+\sqrt{np})$ and is $o(np\tau a_n)$ by the two lower bounds in the theorem. Finally, \pref{eq:lambda-ratio-testing-large} replaces $\lambda_1$ by $\lambda_1'$. Since $a_n\geq M_n$, \pref{eq:testing-two-dimensional-compression} again exceeds the perturbative threshold.

It remains only to treat $\mu>\mu_\star$, where a much coarser argument suffices. Let $p_{\mathrm{in}}$, $p_{\mathrm{out}}$, and $\Delta_\mu$ be as defined above. Writing a same-community inner product as one half of the difference of an independent noncentral and central chi-square variable, and an across-community inner product as its negative, a standard Gaussian tail comparison gives
\begin{equation}\label{eq:large-mu-edge-contrast}
 \Delta_\mu\geq c_\varepsilon p\min\{d\tau\mu^2,1\}.
\end{equation}
Compress $A$ to the span of the normalized indicators of the two communities. Conditional on the labels, the mean of this $2\times2$ compression has second eigenvalue $(1+o(1))n\Delta_\mu$. A Hoeffding decomposition for the three within/across edge counts, followed by Bernstein's inequality, shows that the compression differs from its mean by
\begin{align*}
 O_{\mathbb P}(\sqrt{np}\,\omega_n)
\end{align*}
for any sufficiently slowly diverging sequence $\omega_n$. The hypotheses of the theorem imply
\begin{align*}
 \sqrt{np}\min\{d\tau\mu^2,1\}\longrightarrow\infty,
\end{align*}
so $\omega_n$ can be chosen such that this error is $o(n\Delta_\mu)$. By Cauchy interlacing,
\begin{align*}
 \eta_1(A_{\mathrm m})\geq(1-o(1))n\Delta_\mu.
\end{align*}
Under the null, the same low-moment trace estimate used in \pref{p:op1} gives
\begin{align*}
 \eta_1(A_{\mathrm g})=O(np\tau')+O_{\mathbb P}(\sqrt{np}\,\omega_n)=o(n\Delta_\mu),
\end{align*}
where the last relation follows from \pref{eq:large-mu-edge-contrast}, $d\mu^2\geq\sqrt d/\log n\to\infty$, and the two signal inequalities in the theorem. Therefore the large-separation threshold $n\Delta_\mu/2$ has both errors tending to zero. Combining the three ranges proves the theorem.
\end{proof}

\subsection{Latent vector embedding}
In this subsection, we prove our latent vector embedding results. We re-state our theorem in terms of $\tau$ for convenience.
\begin{theorem*}[Restatement of \pref{thm:latent1}]
Suppose that $n,d\in \Z_+$ and $\mu \in \R_+$, and  $p \in [0,1/2-\varepsilon]$ for any constant $\eps>0$, satisfy the conditions $\log^{16} n\ll d < n$, $\mu^2 \leq 1/(\sqrt{d}\log n)$, and $pn\gg 1$.
Then given $G \sim \geo{n}{d}{p}{\mu}$ generated by latent vectors $u_1,\ldots,u_n \in \R^d$, the spectral algorithm described before produces vectors $\hat{u}_1,\ldots,\hat{u}_n$ which satisfy
\begin{align*}
    \E_{i,j \sim [n]} |\inp{\hat u_i,\hat u_j}- \inp{u_i,u_j}| \ll \max \left\{ \tau, \frac{\mu^2}{\tau}, \frac{1}{\sqrt{np} \tau} \right\} \log^9 n
\E_{i,j \sim [n]} |\inp{u_i,u_j}|,
\end{align*}
with high probability as $n$ goes to infinity.
\end{theorem*}
We state another approximation theorem in terms of the spectral distance between the matrices.
\begin{theorem}\label{thm:latent2}
Suppose that $n,d\in \Z_+$ and $\mu \in \R_+$, and $p \in [0,1/2-\varepsilon]$ for any constant $\eps>0$, satisfy the conditions $\log^{16} n\ll d < n$, $\mu^2 \leq 1/(\sqrt{d}\log n)$, and $pn\gg 1$. Then
\begin{align*}
    \| \sU \sU^\top- UU^\top \|_{\mathrm{op}} \ll \frac{1}{1+\mu^2 d}\max \left\{ \tau, \frac{d\mu^4}{\tau}, \frac{1}{\sqrt{np} \tau }\right\} \log^9 (n) \| UU^\top \|_{\mathrm{op}},
\end{align*}
with high probability as $n$ goes to infinity.
\end{theorem}
\begin{proof}[Proof of \pref{thm:latent2}]
    This theorem follows directly from \pref{thm:testing}, \pref{thm:testing2}, and \pref{lem:uuspec}.
\end{proof}

In the rest of the subsection, we prove \pref{thm:latent1}. We first prove a proposition that bounds the Frobenius norm of $\sU\sU^\top - UU^\top$.
\begin{proposition}\label{p:fronorm}
    We have that for any \muall,
    \begin{align*}
        \|\sU\sU^\top - UU^\top\|_{\mathrm{F}} \llwhp \sqrt{d} \max \left\{ \frac{n\tau}{d}, \frac{n\mu^2}{d\tau}, \frac{\sqrt{n}}{\sqrt{p} d\tau} \right\} \log^9 (n).
    \end{align*}
\end{proposition}
\begin{proof}
    When \musmall, by \pref{thm:testing}, we have that
    \begin{align*}
        \|\sU\sU^\top - UU^\top\|_{\mathrm{F}} \leq \sqrt{2d} \|\sU\sU^\top - UU^\top\|_{\mathrm{op}} \llwhp \sqrt{d} \max \left\{ \frac{n\tau}{d}, \frac{\sqrt{n}}{\sqrt{p} d\tau} \right\} \log^9 (n).
    \end{align*}
    When \mubig, let $D$ denote the diagonal matrix of the difference appearing below. By a proof similar to that of \pref{lem:opdiff2}, we have with high probability
    \begin{align*}
        &\quad \; \| p_0^{i,j}\mathbbm{1}_n \mathbbm{1}_n^\top + \td  \lambda_1^{i,j} V V^\top - (p_0\tilde{\mathbbm{1}}_n \tilde{\mathbbm{1}}_n^\top + \td \lambda_1 UU^\top) -D \|_{\mathrm{F}}\\
        &\leq  \|[f_1(i,j)]_{0,n\times n} + [f_2(i,j)]_{0,n\times n} + [f_3(i,j)]_{0,n\times n} + [f_4(i,j)]_{0,n\times n} \|_{\mathrm{F}}.
    \end{align*}
    For the first term, we have that $\|[f_1(i,j)]_{0,n\times n}\|_\mathrm{F} \leq O \bigp{ np d \tau^4 \log^2 n}+ O(n p d^2 \tau^2 \mu^4)$. And for the second term, similarly, we have that
    $\|[f_2(i,j)]_{0,n\times n}\|_\mathrm{F} \leq o \bigp{ np\tau^4 d \log^2 n}$. For the third term, we have that
    \begin{align*}
    & \|[f_3(i,j)]_{0,n\times n}\|_{\mathrm{F}} \\
    &= \| [
\lambda_1 \td^2 \tau \bigp{ a_ia_j(1-L_i-L_j)+ \tau (L_i + L_j)+ O(\tau^3 \log^2 n) +O(\mu^4 \tau d)} \inp{ v_i,v_j }]_{0,n\times n} \|_\mathrm{F}\\
& \leq \fn{ [
\lambda_1 \td^2 \tau a_i a_j \inp{v_i,v_j}]_{0,n\times n} } + 2\fn{ [
\lambda_1 \td^2 \tau a_i a_j L_i \inp{v_i,v_j} ]_{0,n\times n} }+ 2\fn{ [
\lambda_1 \td^2 \tau^2 L_i \inp{v_i,v_j} ]_{0,n\times n} } \\
& \hspace{6cm} + n \lambda_1 \td^2 \tau O(\tau^3 \log^2 n) o( \tfrac{\log n}{\sqrt{d}}) + n \lambda_1 \td^2 \tau O(\mu^4 \tau d) O( \tfrac{\log n}{\sqrt{d}}) \\
& \leq \lambda_1 \td^2 \tau \bigp{\mu^2+\tfrac{\log n}{d}} O \bigp{\tfrac{ n}{\sqrt{d}} } + \lambda_1 \td^2 \tau \bigp{\mu^2+\tfrac{\log n}{d}} o \bigp{\tfrac{ \log n}{\sqrt{d}} } O \bigp{\tfrac{ n}{\sqrt{d}} } + \lambda_1 \td^2 \tau^2 o \bigp{\tfrac{ \log n}{\sqrt{d}} } O \bigp{\tfrac{ n}{\sqrt{d}} } \\
& \hspace{6cm} + n \lambda_1 \td^2 \tau O(\tau^3 \log^2 n) o( \tfrac{\log n}{\sqrt{d}}) + n \lambda_1 \td^2 \tau O(\mu^4 \tau d) O( \tfrac{\log n}{\sqrt{d}}) \\
& \leq o \bigp { np\tau^5 d^2 \log^3 n + np\sqrt{d} \mu^2 \log n} + O\bigp{np\mu^4\tau^3d^{5/2}\log n}.
\end{align*}
Furthermore,
\begin{align*}
    &\fn{[f_4(i,j)]_{0,n\times n}} = \fn{ [  \td   \lambda_1 (\ell_i \ell_j-1) \inp{v_i,v_j} ]_{0,n\times n} } \\
    &= \fn{ [  \td   \lambda_1 (L_i+L_j + o(\tfrac{\log^2 n
    }{d})) \inp{v_i,v_j} ]_{0,n\times n} }\\
    & = \fn{ [  \td   \lambda_1 (L_i+L_j) \inp{v_i,v_j} ]_{0,n\times n} }+ \fn{ [  \td   \lambda_1 o(\tfrac{\log^2 n}{d}) \inp{v_i,v_j} ]_{0,n\times n}} \\
    & \leq \td   \lambda_1 o \bigp{ \tfrac{\log n}{\sqrt{d}} } O \bigp{ \tfrac{n}{\sqrt{d}} } +  n\td   \lambda_1 o \bigp{ \tfrac{\log^2 n}{d} } o \bigp{ \tfrac{\log n}{\sqrt{d}} }  \leq o \bigp { np\tau \log n }.
\end{align*}
Putting all the inequalities together, the Frobenius norm on the left-hand side above is
$o(np\tau \log^5 n) + O(np\sqrt{d}\mu^2\log^{3/2} n)$.
We also note that by the same argument as in the proof of \pref{p:op1} and \pref{p:largemu},
    \begin{align*}
        \|D\|_{\mathrm{F}}= O(\sqrt{n}dp\tau) = o(\sqrt{dnp} \log^4(n)).
    \end{align*}
Therefore, after restoring the diagonal, the same Frobenius norm is
\begin{align*}
o((np\tau+\sqrt{dnp})\log^5 n) + O(np\sqrt{d}\mu^2\log^{3/2} n).
\end{align*}
By \pref{p:op2}, we further have that with high probability,
    \begin{align*}
        &\quad \;\| \eta_0w_0 w_0^\top + \td\lambda_1 \sU\sU^\top
        -p_0\tilde{\mathbbm{1}}_n \tilde{\mathbbm{1}}_n^\top - \td \lambda_1 UU^\top
        \|_{\mathrm{F}}\\
        & = o\bigp{ \max\left\{ np\tau^2, \sqrt{np}, np\mu^2 \right\} \sqrt{d}\log^9 (n)}.
    \end{align*}
    Thus, we have that
    \begin{align*}
        &\| \td\lambda_1 \sU\sU^\top - \td \lambda_1 UU^\top\|_{\mathrm{F}}\\
        &\leq
        \| \eta_0w_0 w_0^\top + \td\lambda_1 \sU\sU^\top
        -p_0\tilde{\mathbbm{1}}_n \tilde{\mathbbm{1}}_n^\top - \td \lambda_1 UU^\top
        \|_{\mathrm{F}} + \| \eta_0w_0 w_0^\top
        -p_0\tilde{\mathbbm{1}}_n \tilde{\mathbbm{1}}_n^\top
        \|_{\mathrm{F}} \\
        &\ll \max\left\{ np\tau^2, \sqrt{np}, np\mu^2 \right\} \sqrt{d}\log^9 (n).
    \end{align*}
    Since $\td\lambda_1=\Theta(dp\tau)$, the preceding bound gives
    \begin{align*}
        \| \sU\sU^\top - UU^\top\|_{\mathrm{F}}
        &\ll \sqrt d\max \left\{ \frac{n\tau}{d}, \frac{n\mu^2}{d\tau}, \frac{\sqrt n}{\sqrt p\,d\tau} \right\}\log^9 n.
    \end{align*}
\end{proof}

\begin{proof}[Proof of \pref{thm:latent1}]
We note that by Cauchy--Schwarz inequality and \pref{p:fronorm},
\begin{align*}
    \frac{1}{n^2}\sum_{i,j \in [n]} |\inp{\hat u_i,\hat u_j}- \inp{u_i,u_j}| &\leq \sqrt{\frac{1}{n^2}\sum_{i,j \in [n]} |\inp{\su_i,\su_j}- \inp{u_i,u_j}|^2} = \frac{1}{n} \|\sU\sU^\top - UU^\top\|_{\mathrm{F}} \\
    & \ll \frac{\sqrt{d}}{n} \max \left\{ \frac{n\tau}{d}, \frac{n\mu^2}{d\tau}, \frac{\sqrt{n}}{\sqrt{p} d\tau} \right\} \log^9 (n).
\end{align*}
    Therefore, our goal remains to show that
    \begin{align*}
        \frac{1}{n^2}\sum_{i,j \in [n]} |\inp{u_i,u_j}| =\Omega \bigp{ \frac{1}{\sqrt{d}} }.
    \end{align*}
For any fixed vector $v$, we can write
\begin{align*}
    \inp{v,u_j} = N_j \|v\|+\mu S_j v_1,
\end{align*}
where $N_j\sim\calN(0,1/d)$ and $S_j$ is an independent Rademacher variable. A centered Gaussian assigns maximal mass to an interval of fixed length when that interval is centered at the origin. Conditioning on $S_j$ therefore gives
\begin{align*}
    \Pr\left(|\inp{v,u_j}|\geq \frac{1}{2\sqrt d}\right)
    &\geq \Pr\left(|N_j|\|v\|\geq \frac{1}{2\sqrt d}\right)
      =2\Psi\left(\frac{1}{2\|v\|}\right).
\end{align*}
Let $q_0:=2\Psi(1)>0$. For each $i\in[n]$ with $\|u_i\|\geq1/2$, conditional on $u_i$ the random variables $\{|\inp{u_i,u_j}|\}_{j\neq i}$ are independent and
\begin{align*}
 \Pr\left(|\inp{u_i,u_j}|\geq\frac{1}{2\sqrt d}\mid u_i\right)\geq q_0.
\end{align*}
A Chernoff bound consequently gives constants $c_0,c_1>0$ such that, for every such $i$,
\begin{align*}
    \Pr \left(\frac{1}{n-1} \sum_{j\neq i} |\inp{u_i,u_j}| \geq \frac{c_0}{\sqrt d}\,\middle|\,u_i\right) \geq 1-e^{-c_1n}.
\end{align*}
Notice that with high probability there are at least $n/2$ of $i\in [n]$ with $\|u_i\|\geq 1/2$. So by a union bound over all $i\in [n]$ with $\|u_i\|\geq 1/2$, we have that with high probability,
\begin{align*}
     \frac{1}{n^2}\sum_{i,j \in [n]} |\inp{u_i,u_j}| &=\Omega \bigp{ \frac{1}{\sqrt{d}} }.\qedhere
\end{align*}
\end{proof}

\subsection{Spectral clustering}
In this section, we will prove our clustering result.
The theorem follows from an analysis of a basic spectral clustering algorithm for a mixture of two Gaussians under an arbitrary symmetric perturbation bounded in operator norm.
The argument is certainly not novel, but we could not find a statement in the literature which matched our precise needs and so we include this section for completeness.

\begin{algorithm}[Spectral clustering]\label{alg:spectral-clustering}
On input a symmetric positive semidefinite matrix $M \in \R^{n \times n}$, compute a unit eigenvector $a$ associated with its largest eigenvalue and output $y = \mathrm{sign}(a)$, applying the sign function entrywise and breaking ties arbitrarily if $a_i = 0$.
\end{algorithm}
\begin{proposition}\label{p:spectrala}
Suppose $U \in \R^{d \times n}$ with $d \le n$ has columns $u_1,\ldots,u_n$ which are sampled independently from the Gaussian mixture $\frac{1}{2}\calN(-\theta, \frac{1}{d}\Id_d) + \frac{1}{2}\calN(+\theta, \frac{1}{d}\Id_d)$ with $\|\theta\|=\mu$, and let $x \in \{\pm 1\}^n$ denote the vector of the component labels of the $u_i$, so that $x_i = +1$ if and only if $u_i$ was sampled from the mixture component with mean $+\theta$.

Then with probability $1-o(1)$ over the matrix $U$, the following holds simultaneously for every symmetric matrix $\Delta$ such that $M=U^\top U+\Delta$ is positive semidefinite and $\|\Delta\|_{\mathrm{op}}\leq\eta$: the output of \pref{alg:spectral-clustering} is a vector $y \in \{\pm 1\}^n$ with
\begin{align*}
\frac{|\iprod{x,y}|}{n} \ge 1 - O\left(\frac{1}{\mu \sqrt{d}}\right) - O\left(\sqrt{\frac{\eta}{\mu^2 n}}\right),
\end{align*}
that is, \pref{alg:spectral-clustering} clusters at most a $O(\frac{1}{\mu\sqrt{d}}) + O(\sqrt{\frac{\eta}{\mu^2 n}})$-fraction of the columns of $U$ incorrectly.
\end{proposition}

We re-state our theorem for convenience.
\begin{theorem*}[Restatement of \pref{thm:spectralc}]
Suppose that $n,d\in \Z_+$ and $\mu \in \R_+$, and  $p \in [0,1/2-\varepsilon]$ for any constant $\eps>0$, satisfy the conditions $d^{-1/2} \ll \mu \leq d^{-1/4}\log^{-1/2}n$, $\log^{16} n\ll d < n$ and $pn\gg 1$.
If $G \sim \geo{n}{d}{p}{\mu}$, then with high probability \pref{alg:spectral-clustering} on input $\sU\sU^\top$ correctly labels (up to a global sign flip) a
\begin{align*}
1- O\left(\frac{1}{\mu \sqrt{d}}+ \sqrt{\max\left\{ \frac{\mu^2}{\tau}, \frac{1}{d\tau \mu^2 \sqrt{np}}\right\} \log^9 n }\right)\text{-fraction of the vertices.}
\end{align*}
\end{theorem*}
\begin{remark}
Similar to the case in latent vector recovery, potentially we could remove the second error term $O(\sqrt{\mu^2/\tau})$ by choosing $\hat u_i$ in a slightly different way that accounts for changes in the connecting probability as $\mu$ gets close to its upper limit of $d^{-1/4} \log^{-1/2} n$.
\end{remark}

\begin{proof}[Proof of \pref{thm:spectralc}]
For \mubig, we prove the theorem by combining \pref{p:spectrala} with \pref{thm:testing2}.
\end{proof}

\begin{proof}[Proof of \pref{p:spectrala}]
Write
\begin{align*}
 U=Z+\theta x^\top,\qquad \bar x:=\frac{x}{\sqrt n},\qquad
 \alpha:=\frac{1}{\mu\sqrt d},\qquad r:=\frac{\eta}{n\mu^2},
\end{align*}
where the columns of $Z$ are independent $\calN(0,d^{-1}\Id_d)$ vectors. If $\alpha+\sqrt r$ exceeds a sufficiently small absolute constant, the claimed bound is vacuous after enlarging the implicit constants. We may therefore assume throughout that $\alpha+\sqrt r$ is at most a sufficiently small absolute constant.

Decompose
\begin{align*}
 U^\top U=n\mu^2\bar x\bar x^\top+E,
 \qquad
 E:=Z^\top Z+x\theta^\top Z+Z^\top\theta x^\top.
\end{align*}
Standard Gaussian concentration (for example, \cite[Theorem 4.6.1]{Vershynin}) gives, with probability $1-o(1)$,
\begin{align*}
 \|Z^\top Z\|_{\mathrm{op}}=\|ZZ^\top\|_{\mathrm{op}}=O(n/d)
 \quad\text{and}\quad
 \|Z^\top\theta\|=O\bigp{\mu\sqrt{n/d}}.
\end{align*}
Therefore
\begin{equation}\label{eq:clustering-E}
 \|E\|_{\mathrm{op}}
 \leq \|Z^\top Z\|_{\mathrm{op}}+2\sqrt n\,\|Z^\top\theta\|
 =O\bigp{n\mu^2(\alpha+\alpha^2)}.
\end{equation}
Let $a_1$ be a top unit eigenvector of $U^\top U$. The rank-one matrix $n\mu^2\bar x\bar x^\top$ has eigengap $n\mu^2$. By \pref{eq:clustering-E}, Weyl's inequality and the Davis--Kahan theorem imply
\begin{equation}\label{eq:clustering-pop-vector}
 \lambda_1(U^\top U)-\lambda_2(U^\top U)
 \geq n\mu^2(1-O(\alpha)),
 \qquad
 \min_{s\in\{\pm1\}}\|a_1-s\bar x\|=O(\alpha).
\end{equation}

Now let $a$ be the unit top eigenvector of $M=U^\top U+\Delta$. Since $r=o(1)$, the perturbation is smaller than a fixed fraction of the eigengap in \pref{eq:clustering-pop-vector}. A second application of the Davis--Kahan theorem gives
\begin{align*}
 \min_{s\in\{\pm1\}}\|a-sa_1\|
 =O\bigp{\eta/(n\mu^2)}=O(r).
\end{align*}
After choosing the two signs consistently and applying the triangle inequality,
\begin{equation}\label{eq:clustering-vector-distance}
 \min_{s\in\{\pm1\}}\|a-s\bar x\|=O(\alpha+r).
\end{equation}

Let $y=\mathrm{sign}(a)$ and choose the sign $s$ in \pref{eq:clustering-vector-distance}. If $y_i\neq sx_i$, then
$|a_i-sx_i/\sqrt n|\geq n^{-1/2}$. Hence, writing
$\mathcal M:=\{i:y_i\neq sx_i\}$,
\begin{align*}
 \frac{|\mathcal M|}{n}
 \leq \|a-s\bar x\|^2
 =O\bigp{(\alpha+r)^2}.
\end{align*}
It follows that
\begin{align*}
 \frac{|\langle x,y\rangle|}{n}
 &=1-2\frac{|\mathcal M|}{n}\\
 &\geq 1-O\bigp{(\alpha+r)^2}
 \geq 1-O(\alpha)-O(\sqrt r),
\end{align*}
where the last inequality uses $0\leq\alpha,r\leq1$. Substituting the definitions of $\alpha$ and $r$ proves the proposition.
\end{proof}

\subsection{Estimating the latent dimension}\label{sec:unknownd}
The degree direction and the mixture-mean direction may each create a large early spectral gap. After omitting those two gaps, however, the remaining $d-1$ signal eigenvalues lie in a narrow band of order $np\tau$, and the gap from the last signal eigenvalue to the noise spectrum is also of order $np\tau$. This yields a parameter-free estimator.

\begin{proposition}\label{p:unknownd}
Suppose that $n,d\in \Z_+$, $\mu\in\R_+$, and $p\in[0,1/2-\varepsilon]$ for a fixed constant $\varepsilon>0$ satisfy
\begin{align*}
 \log^{16}n\ll d\leq n-2,\qquad
 \mu^2\leq\frac{1}{\sqrt d\log n},\qquad np\gg1.
\end{align*}
Let $\eta_0\geq\eta_1\geq\cdots\geq\eta_{n-1}$ be the eigenvalues of the adjacency matrix $A$, and set
\begin{align*}
 R_n:=\log^9 n\max\{np\tau^2,\;npd\mu^4,\;\sqrt{np}\}.
\end{align*}
Assume in addition that $np\tau\gg R_n$. Define the data-driven estimator
\begin{align*}
 \widehat d\in\argmax_{2\leq i\leq n-2}(\eta_i-\eta_{i+1}),
\end{align*}
with ties broken arbitrarily. Then $\widehat d=d$ with high probability. More precisely,
\begin{align*}
 \eta_d-\eta_{d+1}=\Theta(np\tau),
 \qquad
 \max_{\substack{2\leq i\leq n-2\\ i\neq d}}(\eta_i-\eta_{i+1})=o(np\tau)
\end{align*}
with high probability.
\end{proposition}

\begin{proof}[Proof of \pref{p:unknownd}]
Write $v:=\tilde{\mathbbm{1}}_n$, let
\begin{align*}
 Q:=\frac{vv^\top}{\|v\|^2},\qquad P:=I-Q,
 \qquad B:=PUU^\top P,
\end{align*}
and define
\begin{align*}
 S:=p_0vv^\top+\td\lambda_1 B.
\end{align*}
By \pref{p:ortho}, with high probability
\begin{equation}\label{eq:unknown-d-approx}
 e_n:=\|A-S\|_{\mathrm{op}}=o(R_n)=o(np\tau).
\end{equation}
The two summands in $S$ have orthogonal ranges because $Pv=0$.

We first locate the nonzero eigenvalues of $B$. Write $\theta:=\mu e_1$ and let
\begin{align*}
 \Sigma:=\E[u_i u_i^\top]=d^{-1}\Id_d+\theta\theta^\top.
\end{align*}
The whitened vectors $\Sigma^{-1/2}u_i$ are independent and isotropic. Their subgaussian norms are bounded by an absolute constant: in the direction of $\theta$ the normalized shift has magnitude at most one, and in the orthogonal directions the vectors are standard Gaussian after whitening. Moreover, the assumption $np\tau\gg R_n$ and \pref{lem:tau} imply
\begin{align*}
 np\tau^2\gg\log^{18}n,
 \qquad
 \frac nd\,p\log(1/p)\gg\log^{18}n,
\end{align*}
and therefore $n/d\gg\log^{18}n$. Applying the sample-covariance bound \cite[Theorem 4.6.1]{Vershynin} to the whitened rows gives
\begin{equation}\label{eq:unknown-d-covariance}
 (1-o(1))n\Sigma\preceq U^\top U\preceq(1+o(1))n\Sigma
\end{equation}
with high probability. Thus, if $\zeta_1\geq\cdots\geq\zeta_d$ are the eigenvalues of $U^\top U$, then
\begin{equation}\label{eq:unknown-d-unprojected-spectrum}
 \zeta_1\leq(1+o(1))n(\mu^2+d^{-1}),
 \qquad
 \zeta_j=(1+o(1))\frac nd\quad(2\leq j\leq d).
\end{equation}

The nonzero eigenvalues of $B$ are those of
\begin{align*}
 U^\top P U=U^\top U-U^\top Q U.
\end{align*}
The concentration estimates in the proof of \pref{p:ortho} give
\begin{align*}
 \|U^\top Q U\|_{\mathrm{op}}
 =\frac{\|v^\top U\|^2}{\|v\|^2}=O(\log n)=o(n/d).
\end{align*}
Consequently, if $\gamma_1\geq\cdots\geq\gamma_d>0$ are the nonzero eigenvalues of $B$, Weyl's inequality and \pref{eq:unknown-d-unprojected-spectrum} yield
\begin{equation}\label{eq:unknown-d-projected-spectrum}
 \gamma_1\leq(1+o(1))n(\mu^2+d^{-1}),
 \qquad
 \gamma_j=(1+o(1))\frac nd\quad(2\leq j\leq d).
\end{equation}

Set $a:=\td\lambda_1=\Theta(dp\tau)$ and $\beta:=p_0\|v\|^2=\Theta(np)$. By \pref{lem:tau}, $\log(1/p)\leq(1+o(1))\log n$, and the assumed upper bound on $\mu$,
\begin{align*}
 \frac{a\gamma_1}{np}
 =O\bigp{\tau(1+d\mu^2)}
 =O\bigp{\sqrt{\log(1/p)/d}+\sqrt{\log(1/p)}/\log n}
 =o(1).
\end{align*}
Hence $\beta$ is the largest eigenvalue of $S$. If $\sigma_0\geq\cdots\geq\sigma_{n-1}$ are the eigenvalues of $S$, then with high probability
\begin{align*}
 \sigma_0=\beta,\qquad
 \sigma_j=a\gamma_j\ (1\leq j\leq d),\qquad
 \sigma_j=0\ (j\geq d+1).
\end{align*}
In particular, \pref{eq:unknown-d-projected-spectrum} gives
\begin{align*}
 \sigma_d-\sigma_{d+1}&=a\gamma_d=\Theta(np\tau),\\
 \max_{2\leq i\leq d-1}(\sigma_i-\sigma_{i+1})&=o(np\tau),\\
 \max_{d+1\leq i\leq n-2}(\sigma_i-\sigma_{i+1})&=0.
\end{align*}
The gap at $i=1$ is deliberately excluded because $\gamma_1$ may be separated from $\gamma_2$ by the mixture mean; the gap at $i=0$ is excluded because of the degree eigenvalue.

Finally, Weyl's inequality and \pref{eq:unknown-d-approx} imply, uniformly in $i$,
\begin{align*}
 \left|(\eta_i-\eta_{i+1})-(\sigma_i-\sigma_{i+1})\right|\leq2e_n=o(np\tau).
\end{align*}
Therefore the gap at $i=d$ has order $np\tau$, while every other gap in the estimator's range is $o(np\tau)$. The maximizing index is thus uniquely $d$ with high probability.
\end{proof}

\subsection*{Acknowledgments}
T.S. would like to thank Samory Kpotufe for suggesting the question of spectral clustering in the GBM, and would like to thank Sidhanth Mohanty and Sam Hopkins for inspiring conversations about the work and its relation to recovering embeddings in other models. The authors also wish to thank Kiril Bangachev for pointing out two errors in the original trace method proof.

\subsection*{Funding}
The second author is supported by NSF CAREER award \# 2143246.

\bibliographystyle{alpha}
\bibliography{reference}
\appendix
\section{Lower bound for hypothesis testing when the latent vectors are observed}
\label{app:kl}
In this appendix we give a lower bound for hypothesis testing even when the latent vectors are observed. We assume that the direction $\theta$ is not revealed and that we observe $u_1,\ldots,u_n$ under either $u_1,\ldots,u_n \sim H_0 = \calN(0,\frac{1}{d}\Id)$ or $u_1,\ldots,u_n \sim H_1 = \frac{1}{2}\calN(-\mu \theta,\frac{1}{d}\Id) + \frac{1}{2}\calN(\mu \theta,\frac{1}{d}\Id)$, where $\theta \sim \unif(\cS^{d-1})$.

Our proof is similar to the proof of Theorem 3 in \cite{banks2018information}, with two differences. In \cite{banks2018information}, it is assumed that $d/n = \Theta(1)$ whereas for us, $d/n$ can approach $0$ or $\infty$. Also, the signal vector in \cite{banks2018information} follows a Gaussian distribution, while the signal vector in our setting is either $\theta$ or $-\theta$. The proof idea is to use a second-moment computation to show that the alternative distribution is contiguous to the null distribution when the separation $\mu$ is small.

We change our notation slightly to align with the notation in \cite{banks2018information}. Define $X$ to be a $n$ by $d$ matrix with $i.i.d.$ $\cN(0,1)$ entries. Let $\mathbb{P}$ be the distribution of $X$. Define $S$ to be a $n$ by $d$ random matrix where each row equals $\pm \theta$ with probability $1/2$ independently, with $\theta \sim \unif(\cS^{d-1})$. Let $\mathbb{Q}$ be the distribution of $X+\mu\sqrt{d} S$. Then testing $H_0$ versus $H_1$ is the same as testing $\mathbb{P}$ versus $\mathbb{Q}$.

\begin{claim}
For every fixed constant $c<1$, if $\mu\leq c(nd)^{-1/4}$, then $\mathbb{Q}$ is contiguous to $\mathbb{P}$. Consequently, no sequence of tests can distinguish $\mathbb{P}$ from $\mathbb{Q}$ with both error probabilities tending to zero.
\end{claim}
\begin{proof}
Let $L=d\mathbb{Q}/d\mathbb{P}$ be the likelihood ratio. For a fixed signal matrix $S$, the conditional alternative is a Gaussian shift by $\mu\sqrt d\,S$. Hence the standard Gaussian-shift identity gives
\begin{align*}
 \E_{X\sim\mathbb{P}}[L(X)^2]
 &=\E_{S,T}\exp\bigp{\mu^2d\langle S,T\rangle_{\mathrm F}},
\end{align*}
where $T$ is an independent copy of $S$.
Write the rows of $S$ and $T$ as $S_i=\sigma_i\theta$ and $T_i=\tau_i\theta'$, where $\sigma_i,\tau_i$ are independent Rademacher variables and $\theta,\theta'$ are independent and uniform on $\mathbb S^{d-1}$. Set
\begin{align*}
 W:=\sum_{i=1}^n\sigma_i\tau_i,
 \qquad Z:=\langle\theta,\theta'\rangle.
\end{align*}
Then $W$ is a sum of $n$ independent Rademacher variables and
$\langle S,T\rangle_{\mathrm F}=WZ$. Conditioning on $Z$ and using
$\cosh(t)\leq e^{t^2/2}$,
\begin{align}
 \E_{X\sim\mathbb{P}}[L(X)^2]
 &=\E_Z\cosh(\mu^2dZ)^n
 \leq\E_Z\exp\bigp{\tfrac12 n\mu^4d^2Z^2}.
 \label{eq:known-embedding-second-moment}
\end{align}
The density of $Z$ on $[-1,1]$ is
\begin{align*}
 c_d(1-z^2)^{(d-3)/2},
 \qquad
 c_d:=\frac{\Gamma(d/2)}{\sqrt\pi\,\Gamma((d-1)/2)}.
\end{align*}
Let $r:=nd\mu^4$. Under the hypothesis, $r\leq c^4<1$. If $d$ remains bounded, the right-hand side of \pref{eq:known-embedding-second-moment} is at most $\exp(c^4d/2)$ and is therefore uniformly bounded. We may thus restrict attention to sufficiently large $d$. Using $\log(1-z^2)\leq-z^2$ in \pref{eq:known-embedding-second-moment},
\begin{align*}
 \E_{X\sim\mathbb{P}}[L(X)^2]
 &\leq c_d\int_{-1}^1
 \exp\left(-\frac{d(1-r)-3}{2}z^2\right)\,dz\\
 &\leq c_d\sqrt{\frac{2\pi}{d(1-c^4)-3}}
 \leq C_c,
\end{align*}
where the last inequality follows from $c_d=\Theta(\sqrt d)$. Thus the likelihood ratio has a uniformly bounded second moment under $\mathbb P$.

For any sequence of events $A_n$, Cauchy--Schwarz gives
\begin{align*}
 \mathbb Q(A_n)=\E_{\mathbb P}[L\mathbbm{1}_{A_n}]
 \leq \E_{\mathbb P}[L^2]^{1/2}\mathbb P(A_n)^{1/2}.
\end{align*}
Therefore $\mathbb P(A_n)\to0$ implies $\mathbb Q(A_n)\to0$, proving that $\mathbb Q$ is contiguous to $\mathbb P$. A consistent test would have a rejection event whose probability tends to zero under $\mathbb P$ and to one under $\mathbb Q$, which is impossible.
\end{proof}

\section{Distinct connected components when the separation is large}
\label{app:components}

In this appendix we show that when $\mu$ is large, each community corresponds to a distinct connected component in the graph.

For $u_1,\ldots,u_n \sim \frac{1}{2}\calN(-\mu \cdot e_1,\frac{1}{d}\Id) + \frac{1}{2}\calN(\mu \cdot e_1,\frac{1}{d}\Id)$, recall that we say that vertex $i \in [n]$ comes from community $+1$ if $u_i$ comes from the component in the mixture with mean $\mu \cdot e_1$; otherwise we say that vertex $i$ comes from community $-1$. We define $C_+$ to be the set of labels $i\in[n]$ for which $u_i$ comes from community $+1$, and $C_-$ to be the set of labels $i\in[n]$ for which $u_i$ comes from community $-1$.

\begin{claim}
Suppose $p \in [0,1/2-\varepsilon]$ for a fixed constant $\eps>0$ and $\mu \gg (\frac{1}{d}\log n)^{1/4}$. Then there are no edges between $C_+$ and $C_-$ with high probability.
\end{claim}

\begin{proof}
For $i \in C_+$ and $j\in C_-$, we write $u_i = (\mu+N_i, w_i)$, where $N_i \in \mathbb{R}$ and $w_i \in \mathbb{R}^{d-1}$ and similarly write $u_j = (-\mu+N_j, w_j)$, where $N_j \in \mathbb{R}$ and $w_j \in \mathbb{R}^{d-1}$. Here note that $N_i \sim \calN(0,1/d)$, $N_j\sim \calN(0,1/d)$, $w_i \sim \calN(0,I_{d-1}/d)$, $w_j\sim \calN(0,I_{d-1}/d)$ and they are all independent.
\begin{align*}
    &\Pr[i \sim j] = \Pr[(\mu+N_i)(-\mu+N_j) + \inp{w_i,w_j} \geq \tau] \\
    &= \Pr[\mu(N_j-N_i) + N_iN_j+ \inp{w_i,w_j} \geq \tau+\mu^2]\\
    & \leq \Pr[N_j-N_i \geq \tau/(2\mu)+\mu/2]+ \Pr[N_iN_j+ \inp{w_i,w_j} \geq \tau/2+\mu^2/2]\\
    &\leq \Pr[N_j-N_i \geq \mu/2]+ \Pr[N_iN_j+ \inp{w_i,w_j} \geq \mu^2/2],
\end{align*}
since $\tau>0$.
We note that
\begin{align*}
    \Pr[N_j-N_i \geq \mu/2] = \Pr[\sqrt{d/2}(N_j-N_i) \geq \sqrt{d/2} \cdot \mu/2] \leq \exp(-\mu^2 d/8) \ll \exp(-\sqrt{d}).
\end{align*}
Further, we note that $N_iN_j+ \inp{w_i,w_j}$ can be written as the difference of two independent normalized Chi-Squared random variables $N_iN_j+ \inp{w_i,w_j} = (A_d-B_d)/(2d)$, where $A_d, B_d \sim \chi_d^2$. By the Laurent-Massart bound \cite{laurent2000adaptive}, we know that $\Pr[A_d - d\geq 2\sqrt{dx} + 2x] \leq \exp(-x)$, $\Pr[A_d - d\leq -2\sqrt{dx}] \leq \exp(-x)$ and so does $B_d$. This implies that there exists a constant $c$ such that
\begin{align*}
    \Pr[N_iN_j+ \inp{w_i,w_j} \geq \mu^2/2] \leq \exp(-c d \min (\mu^4,1)) \ll n^{-10}.
\end{align*}
By a union bound over all vertices $i \in C_+$ and $j \in C_-$, we have the claim.
\end{proof}

\end{document}